\documentclass[acmsmall]{acmart}
\usepackage[normalem]{ulem}
\useunder{\uline}{\ul}{}
\usepackage{lscape}
\usepackage{longtable}
\usepackage{adjustbox}
\usepackage{amsmath}
\usepackage{graphicx}
\usepackage{mathtools}
\usepackage[title]{appendix}

\AtBeginDocument{%
  \providecommand\BibTeX{{%
    \normalfont B\kern-0.5em{\scshape i\kern-0.25em b}\kern-0.8em\TeX}}}

\setcopyright{acmcopyright}
\copyrightyear{2020}
\acmYear{2020}
\acmDOI{---------}

\begin{document}

\title{Evolution of Semantic Similarity - A Survey}

\author{Dhivya Chandrasekaran}
\email{dchandra@lakeheadu.ca}
\orcid{1234-5678-9012}
\author{Vijay Mago}
\email{vmago@lakeheadu.ca}
\affiliation{%
  \institution{Lakehead University}
  \streetaddress{955 Oliver Road}
  \city{Thunderbay}
  \state{Ontario}
  \postcode{P7B 5E1}
}

\renewcommand{\shortauthors}{D Chandrasekaran and V Mago}

\begin{abstract}
Estimating the semantic similarity between text data is one of the challenging and open research problems in the field of Natural Language Processing (NLP). The versatility of natural language makes it difficult to define rule-based methods for determining semantic similarity measures. In order to address this issue, various semantic similarity methods have been proposed over the years. This survey article traces the evolution of such methods beginning from traditional NLP techniques like kernel-based methods to the most recent research work on transformer-based models, categorizing them based on their underlying principles as knowledge-based, corpus-based, deep neural network-based methods, and hybrid methods. Discussing the strengths and weaknesses of each method, this survey provides a comprehensive view of existing systems in place, for new researchers to experiment and develop innovative ideas to address the issue of semantic similarity.
\end{abstract}

\ccsdesc[500]{General and reference~Surveys and overviews}
\ccsdesc[500]{Information systems~Ontologies}
\ccsdesc[500]{Theory of computation~Unsupervised learning and clustering}
\ccsdesc[500]{Computing methodologies~Lexical semantics}

\keywords{semantic similarity, linguistics, supervised and unsupervised methods, knowledge-based methods, word embeddings, corpus-based methods}

\maketitle
\section{Introduction}
With the exponential increase in text data generated over time, Natural Language Processing (NLP) has gained significant attention from Artificial Intelligence (AI) experts. Measuring the semantic similarity between various text components like words, sentences, or documents plays a significant role in a wide range of NLP tasks like information retrieval \cite{kim2017bridging}, text summarization \cite{mohamed2019srl}, text classification \cite{kim2014convolutional}, essay evaluation \cite{janda2019syntactic}, machine translation \cite{zou2013bilingual}, question answering \cite{bordes2014question, LOPEZGAZPIO2017186}, among others. In the early days, two text snippets were considered similar if they contain the same words/characters. The techniques like Bag of Words (BoW), Term Frequency - Inverse Document Frequency (TF-IDF) were used to represent text, as real value vectors to aid calculation of semantic similarity. However, these techniques did not attribute to the fact that words have different meanings and different words can be used to represent a similar concept. For example, consider two sentences \textit{``John and David studied Maths and Science.''} and \textit{``John studied Maths and David studied Science.''} Though these two sentences have exactly the same words they do not convey the same meaning. Similarly, the sentences \textit{``Mary is allergic to dairy products.''} and \textit{``Mary is lactose intolerant.''} convey the same meaning; however, they do not have the same set of words. These methods captured the lexical feature of the text and were simple to implement, however, they ignored the semantic and syntactic properties of text. To address these drawbacks of the lexical measures various semantic similarity techniques were proposed over the past three decades.
\\
Semantic Textual Similarity (STS) is defined as the measure of semantic equivalence between two blocks of text. Semantic similarity methods usually give a ranking or percentage of similarity between texts, rather than a binary decision as similar or not similar. Semantic similarity is often used synonymously with semantic relatedness. However, semantic relatedness not only accounts for the semantic similarity between texts but also considers a broader perspective analyzing the shared semantic properties of two words. For example, the words \textit{`coffee'} and \textit{`mug'} may be related to one another closely, but they are not considered semantically similar whereas the words \textit{`coffee'} and \textit{`tea'} are semantically similar. Thus, semantic similarity may be considered, as one of the aspects of semantic relatedness. The semantic relationship including similarity is measured in terms of semantic distance, which is inversely proportional to the relationship \cite{HadjTaieb2019}. 
\begin{figure}[h]
    \centering
    \includegraphics[scale=0.5]{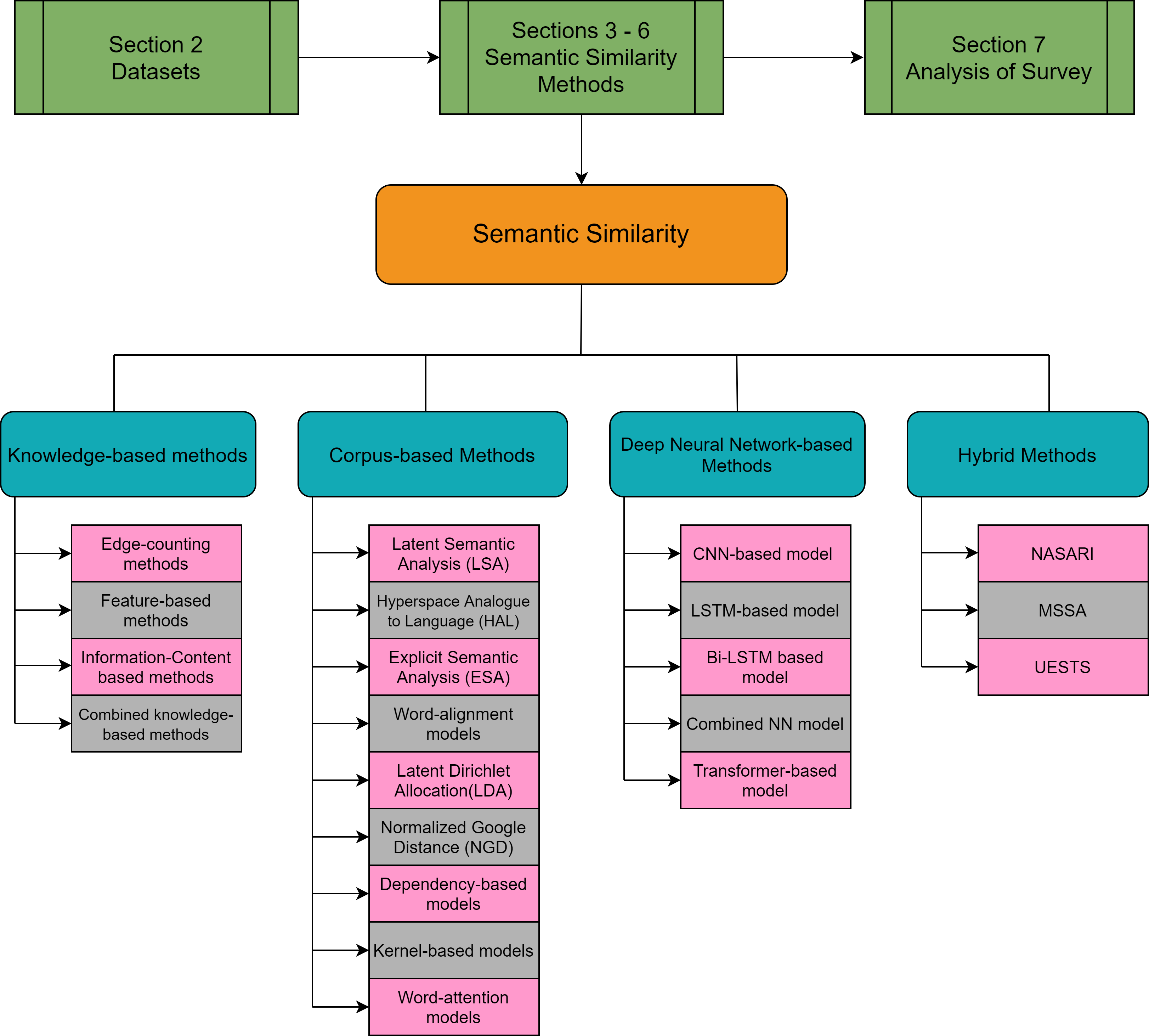}
    \caption{Survey Architecture}
    \label{fig:sa}
\end{figure}
\subsection{Motivation behind the survey}
Most of the survey articles published recently related to semantic similarity, provide in-depth knowledge of one particular semantic similarity technique or a single application of semantic similarity. Lastra-Díaz et al. survey various knowledge-based methods \cite{LASTRADIAZ2019645} and IC-based methods \cite{lastra2015new}, Camacho-Colladas et al. \cite{camacho2018word} discuss various vector representation methods of words, Taieb et al. \cite{HadjTaieb2019}, on the other hand, describe various semantic relatedness methods and Berna Altınel et al. \cite{ALTINEL20181129} summarise various semantic similarity methods used for text classification. The motivation behind this survey is to provide a comprehensive account of the various semantic similarity techniques including the most recent advancements using deep neural network-based methods. 
\\
This survey traces the evolution of Semantic Similarity Techniques over the past decades, distinguishing them based on the underlying methods used in them. Figure \ref{fig:sa} shows the structure of the survey. A detailed account of the widely used datasets available for semantic similarity is provided in Section 2. Sections 3 to 6 provide a detailed description of semantic similarity methods broadly classified as 1) Knowledge-based methods, 2) Corpus-based methods, 3) Deep neural network-based methods, and 4) Hybrid methods. Section 7 analyzes the various aspects and inference of the survey conducted. This survey provides a deep and wide knowledge of existing techniques for new researchers who venture to explore one of the most challenging NLP tasks, Semantic Textual Similarity.

\section{Datasets}
In this section, we discuss some of the popular datasets used to evaluate the performance of semantic similarity algorithms. The datasets may include word pairs or sentence pairs with associated standard similarity values. The performance of various semantic similarity algorithms is measured by the correlation of the achieved results with that of the standard measures available in these datasets. Table \ref{tab:Dataset list} lists some of the popular datasets used to evaluate the performance of semantic similarity algorithms. The below subsection describes the attributes of the dataset and the methodology used to construct them.
\begin{longtable}[c]{|l|l|l|l|l|}
\hline
\textbf{Dataset Name} & \textbf{Word/Sentence pairs} & \textbf{Similarity score range} & \textbf{Year} & \textbf{Reference} \\ \hline
\endfirsthead
\multicolumn{5}{c}%
{{\bfseries Table \thetable\ continued from previous page}} \\
\hline
\textbf{Dataset Name} & \textbf{Word/Sentence pairs} & \textbf{Similarity score range} & \textbf{Year} & \textbf{Reference} \\ \hline
\endhead
R\&G       & 65    & 0-4  & 1965 & \cite{rubenstein1965contextual} \\ \hline
M\&C       & 30    & 0-4  & 1991 & \cite{miller1991contextual} \\ \hline
WS353      & 353   & 0-10 & 2002 & \cite{finkelstein2001placing}\\ \hline
LiSent     & 65    & 0-4  & 2007 &  \cite{li2006sentence}\\ \hline
SRS        & 30    & 0-4  & 2007 &  \cite{pedersen2007measures}\\ \hline
WS353-Sim  & 203   & 0-10 & 2009 & \cite{agirre2009study} \\ \hline
STS2012 & 5250   & 0-5  & 2012 &  \cite{agirre2012semeval} \\ \hline
STS2013 & 2250   & 0-5  & 2013 &  \cite{agirre2013sem} \\ \hline
WP300      & 300   & 0-1  & 2013 &  \cite{li2013computing} \\ \hline
STS2014 & 3750   & 0-5  & 2014 &  \cite{agirre2014semeval} \\ \hline
SL7576     & 7576  & 1-5  & 2014 &  \cite{silberer2014learning}\\ \hline
SimLex-999 & 999   & 0-10 & 2014 &  \cite{hill2015simlex}\\ \hline
SICK       & 10000 & 1-5  & 2014 &  \cite{marellisick}\\ \hline
STS2015 & 3000   & 0-5  & 2015 &  \cite{agirre2015semeval} \\ \hline
SimVerb    & 3500  & 0-10 & 2016 &  \cite{gerz2016simverb} \\ \hline
STS2016 & 1186   & 0-5  & 2016 &  \cite{agirre2016semeval} \\ \hline
WiC        & 5428  & NA   & 2019 &  \cite{pilehvar2019wic}\\ \hline
\caption{Popular benchmark datasets for Semantic similarity}
\label{tab:Dataset list}\\
\end{longtable}
\subsection{Semantic similarity datasets} The following is a list of widely used semantic similarity datasets arranged chronologically.
\begin{itemize}
    \item \textbf{Rubenstein and Goodenough   (R\&G)} \cite{rubenstein1965contextual}: This dataset was created as a result of an experiment conducted among 51 undergraduate students (native English speakers) in two different sessions. The subjects were provided with 65 selected English noun pairs and requested to assign a similarity score for each pair over a scale of 0 to 4, where 0 represents that the words are completely dissimilar and 4 represents that they are highly similar. This dataset is the first and most widely used dataset in semantic similarity tasks \cite{7572993}.
    \item \textbf{Miller and Charles (M\&C)} \cite{miller1991contextual}: Miller and Charles repeated the experiment performed by Rubenstein and Goodenough in 1991 with a subset of 30 word pairs from the original 65 word pairs. 38 human subjects ranked the word pairs on a scale from 0 to 4, 4 being the "most similar."
    \item\textbf{WS353} \cite{finkelstein2001placing}: WS353 contains 353 word pairs with an associated score ranging from 0 to 10. 0 represents the least similarity and 10 represents the highest similarity. The experiment was conducted with a group of 16 human subjects. This dataset measures semantic relatedness rather than semantic similarity. Subsequently, the next dataset was proposed.
    \item\textbf{WS353-Sim} \cite{agirre2009study}: This dataset is a subset of WS353 containing 203 word pairs from the original 353 word pairs that are more suitable for semantic similarity algorithms specifically.
    \item\textbf{LiSent} \cite{li2006sentence}: 65 sentence pairs were built using the dictionary definition of 65 word pairs used in the R\&G dataset. 32 native English speakers volunteered to provide a similarity range from 0 to 4, 4 being the highest. The mean of the scores given by all the volunteers was taken as the final score.
    \item\textbf{SRS} \cite{pedersen2007measures}: Pedersen et al. \cite{pedersen2007measures} attempted to build a domain specific semantic similarity dataset for the biomedical domain. Initially 120 pairs were selected by a physician distributed with 30 pairs over 4 similarity values. These term pairs were then ranked by 13 medical coders on a scale of 1-10. 30 word pairs from the 120 pairs were selected to increase reliability and these word pairs were annotated by 3 physicians and 9 (out of the 13) medical coders to form the final dataset.
    \item\textbf{SimLex-999} \cite{hill2015simlex}: 999 word pairs were selected from the UFS Dataset \cite{nelson2004university} of which 900 were similar and 99 were related but not similar. 500 native English speakers, recruited via Amazon Mechanical Turk were asked to rank the similarity between the word pairs over a scale of 0 to 6, 6 being the most similar. The dataset contains 666 noun pairs, 222 verb pairs, and 111 adjective pairs.
   \item\textbf{Sentences Involving Compositional Knowledge (SICK) dataset} \cite{marellisick}: The SICK dataset consists of 10,000 sentence pairs, derived from two existing datasets the ImageFlickr 8 and MSR-Video descriptions dataset. Each sentence pair is associated with a relatedness score and a text entailment relation. The relatedness score ranges from 1 to 5, and the three entailment relations are "NEUTRAL, ENTAILMENT and CONTRADICTION." The annotation was done using crowd-sourcing techniques. 
   \item\textbf{STS datasets} \cite{agirre2012semeval,agirre2013sem,agirre2014semeval,agirre2015semeval,agirre2016semeval,cer2017semeval}: The STS datasets were built by combining sentence pairs from different sources by the organizers of the SemEVAL shared task. The dataset was annotated using Amazon Mechanical Turk and further verified by the organizers themselves. Table \ref{tab:ststable} shows the various sources from which the STS dataset was built.
        \begin{longtable}[c]{|l|l|r|l|}
        \hline
        \textbf{Year} & \textbf{Dataset} & \textbf{Pairs} & \textbf{Source}        \\ \hline
        \endfirsthead
        \multicolumn{4}{c}%
        {{\bfseries Table \thetable\ continued from previous page}} \\
        \hline
        \textbf{Year} & \textbf{Dataset} & \textbf{Pairs} & \textbf{Source}        \\ \hline
        \endhead
        \hline
        \endfoot
        \endlastfoot
        2012          & MSRPar           & 1500           & newswire               \\
        2012          & MSRvid           & 1500           & videos                 \\
        2012          & OnWN             & 750            & glosses                \\
        2012          & SMTNews          & 750            & WMT eval.              \\
        2012          & SMTeuroparl      & 750            & WMT eval.              \\ \hline
        2013          & HDL              & 750            & newswire               \\
        2013          & FNWN             & 189            & glosses                \\
        2013          & OnWN             & 561            & glosses                \\
        2013          & SMT              & 750            & MT eval.               \\ \hline
        2014          & HDL              & 750            & newswire headlines     \\
        2014          & OnWN             & 750            & glosses                \\
        2014          & Deft-forum       & 450            & forum posts            \\
        2014          & Deft-news        & 300            & news summary           \\
        2014          & Images           & 750            & image descriptions     \\
        2014          & Tweet-news       & 750            & tweet-news pairs       \\ \hline
        2015          & HDL              & 750            & newswire headlines     \\
        2015          & Images           & 750            & image descriptions     \\
        2015          & Ans.-student     & 750            & student answers        \\
        2015          & Ans.-forum       & 375            & Q \& A forum answers   \\
        2015          & Belief           & 375            & committed belief       \\ \hline
        2016          & HDL              & 249            & newswire headlines     \\
        2016          & Plagiarism       & 230            & short-answers plag.    \\
        2016          & post-editing     & 244            & MT postedits           \\
        2016          & Ans.-Ans         & 254            & Q \& A forum answers   \\
        2016          & Quest.-Quest.    & 209            & Q \& A forum questions \\ \hline
        2017          & Trail            & 23             & Mixed STS 2016         \\ \hline
        \caption{STS English language training dataset (2012-2017) \cite{cer2017semeval}}
        \label{tab:ststable}\\
        \end{longtable}
\end{itemize}

\section{Knowledge-based Semantic-Similarity Methods}
Knowledge-based semantic similarity methods calculate semantic similarity between two terms based on the information derived from one or more underlying knowledge sources like ontologies/lexical databases, thesauri, dictionaries, etc. The underlying knowledge-base offers these methods a structured representation of terms or concepts connected by semantic relations, further offering an ambiguity free semantic measure, as the actual meaning of the terms, is taken into consideration \cite{SANCHEZ20127718}. In this section, we discuss four lexical databases widely employed in knowledge-based semantic similarity methods and further discuss in brief, different methodologies adopted by some of the knowledge-based semantic similarity methods. 
\subsection{Lexical Databases} 
\begin{itemize}
    \item WordNet \cite{miller1995wordnet} is a widely used lexical database for knowledge-based semantic similarity methods that accounts for more than 100,000 English concepts \cite{SANCHEZ20127718}. WordNet can be visualized as a graph, where the nodes represent the meaning of the words (concepts), and the edges define the relationship between the words \cite{7572993}. WordNet's structure is primarily based on synonyms, where each word has different \textit{synsets} attributed to their different meanings. The similarity between two words depends on the path distance between them \cite{8630924}.
    \item  Wiktionary\footnote{https://en.wiktionary.org} is an open-source lexical database that encompasses approximately 6.2 million words from 4,000 different languages. Each entry has an article page associated with it, and it accounts for a different sense of each entry. Wiktionary does not have a well-established taxonomic lexical relationship within the entries, unlike WordNet, which makes it difficult to be used in semantic similarity algorithms \cite{PILEHVAR201595}.
    \item With the advent of Wikipedia\footnote{http://www.wikipedia.org}, most techniques for semantic similarity exploit the abundant text data freely available to train the models \cite{mihalcea2007wikify}. Wikipedia has the text data organized as Articles. Each article has a title (concept), neighbors, description, and categories. It is used as both structured taxonomic data and/or as a corpus for training corpus-based methods \cite{QU20181002}. The complex category structure of Wikipedia is used as a graph to determine the Information Content of concepts, which in turn aids in calculating the semantic similarity \cite{JIANG2017248}.
    \item BabelNet \cite{navigli2012babelnet} is a lexical resource that combines WordNet with data available on Wikipedia for each \textit{synset}. It is the largest multilingual semantic ontology available with nearly over 13 million \textit{synsets} and 380 million semantic relations in 271 languages. It includes over four million \textit{synsets} with at least one associated Wikipedia page for the English language \cite{CAMACHOCOLLADOS201636}.
\end{itemize}
\subsection{Types of Knowledge-based semantic similarity methods}
Based on the underlying principle of how the semantic similarity between words is assessed, knowledge-based semantic similarity methods can be further categorized as edge-counting methods, feature-based methods, and information content-based methods. 
\subsubsection{\textbf{Edge-counting methods:}}The most straight forward edge counting method is to consider the underlying ontology as a graph connecting words taxonomically and count the edges between two terms to measure the similarity between them. The greater the distance between the terms the less similar they are. This measure called $path$ was proposed by Rada et al. \cite{rada1989development} where the similarity is inversely proportional to the shortest path length between two terms. In this edge-counting method, the fact that the words deeper down the hierarchy have a more specific meaning, and that, they may be more similar to each other even though they have the same distance as two words that represent a more generic concept was not taken into consideration. Wu and Palmer \cite{wu1994verbs} proposed $wup$ measure, where the depth of the words in the ontology was considered an important attribute. The $wup$ measure counts the number of edges between each term and their Least Common Subsumer (LCS). LCS is the common ancestor shared by both terms in the given ontology. Consider, two terms denoted as $t_1 ,t_2$, their LCS denoted as $t_{lcs}$, and the shortest path length between them denoted as $min\_len(t_1,t_2)$,
\\$path$ is measured as, 
\begin{equation}
    sim_{path}(t_1,t_2) = \frac{1}{1+min\_len(t_1,t_2)}
\end{equation} 
and $wup$ is measured as, 
\begin{equation}
    sim_{wup}(t_1,t_2) = \frac{2depth(t_{lcs})}{depth(t_1)+depth(t_2)}
\end{equation} 

Li et al. \cite{li2003approach} proposed a measure that takes into account both the minimum path distance and depth. $li$ is measured as,
\begin{equation}
    sim_{li} = e^{-\alpha min\_len(t_1,t_2)} . \frac{e^{\beta depth(t_{lcs})} - e^{-\beta depth(t_{lcs})}}{e^{\beta depth(t_{lcs})} + e^{-\beta depth(t_{lcs})}}
\end{equation}

However, the edge-counting methods ignore the fact that the edges in the ontologies need not be of equal length. To overcome this shortcoming of simple edge-counting methods feature-based semantic similarity methods were proposed. 
\subsubsection{\textbf{Feature-based methods:}}The feature-based methods calculate similarity as a function of properties of the words, like gloss, neighboring concepts, etc. \cite{SANCHEZ20127718}. Gloss is defined as the meaning of a word in a dictionary; a collection of glosses is called a glossary. There are various semantic similarity methods proposed based on the gloss of words. Gloss-based semantic similarity measures exploit the knowledge that words with similar meanings have more common words in their gloss. The semantic similarity is measured as the extent of overlap between the gloss of the words in consideration. The Lesk measure \cite{banerjee2003extended}, assigns a value of relatedness between two words based on the overlap of words in their gloss and the glosses of the concepts they are related to in an ontology like WordNet \cite{LASTRADIAZ2019645}. Jiang et al. \cite{jiang2015feature} proposed a feature-based method where semantic similarity is measured using the glosses of concepts present in Wikipedia. Most feature-based methods take into account common and non-common features between two words/terms. The common features contribute to the increase of the similarity value and the non-common features decrease the similarity value. The major limitation of feature-based methods is its dependency on ontologies with semantic features, and most ontologies rarely incorporate any semantic features other than taxonomic relationships \cite{SANCHEZ20127718}. 
\subsubsection{\textbf{Information Content-based methods:}} Information content (IC) of a concept is defined as the information derived from the concept when it appears in context \cite{SANCHEZ20131393}. A high IC value indicates that the word is more specific and clearly describes a concept with less ambiguity, while lower IC values indicate that the words are more abstract in meaning \cite{7572993}. The specificity of the word is determined using Inverse Document Frequency (IDF), which relies on the principle that the more specific a word is, the less it occurs in a document. Information content-based methods measure the similarity between terms using the IC value associated with them. Resnik and Philip \cite{resnik1995using} proposed a semantic similarity measure called $res$ which measures the similarity based on the idea that if two concepts share a common subsumer they share more information since the $IC$ value of the LCS is higher. Considering $IC$ represents the Information Content of the given term, $res$ is measured as,
\begin{equation}
    sim_{res}(t_1,t_2) = IC_{t_{lcs}}
\end{equation}
D. Lin \cite{lin1998information} proposed an extension of the $res$ measure taking into consideration the $IC$ value of both the terms that attribute to the individual information or description of the terms and the $IC$ value of their LCS that provides the shared commonality between the terms. $lin$ is measured as,
\begin{equation}
    sim_{lin}(t_1,t_2) = \frac{2IC_{t_{lcs}}}{IC_{t_1}+IC_{t_2}}
\end{equation}
Jiang and Conrath \cite{jiang1997semantic} calculate a distance measure based on the difference between the sum of the individual $IC$ values of the terms and the $IC$ value of their LCS using the below equation,
\begin{equation}
    dis_{jcn}(t_1,t_2) = IC_{t_1} + IC_{t_2} - 2IC{t_{lcs}}
\end{equation}
The distance measure replaces the shortest path length in equation (1), and the similarity is inversely proportional to the above distance. Hence $jcn$ is measured as, 
\begin{equation}
    sim_{jcn}(t_1,t_2) = \frac{1}{1+dis_{jcn}(t_1,t_2)}
\end{equation}

IC can be measured using an underlying corpora or from the intrinsic structure of the ontology itself \cite{sanchez2011ontology} based on the assumption that the ontologies are structured in a meaningful way. Some of the terms may not be included in one ontology, which provides a scope to use multiple ontologies to calculate their relationship \cite{rodriguez2003determining}. Based on whether the given terms are both present in a single ontology or not, IC-based methods can be classified as mono-ontological methods or multi-ontological methods. When multiple ontologies are involved the $IC$ of the Least Common Subsumer from both the ontologies are accessed to estimate the semantic similarity values. Jiang et al. \cite{JIANG2017248} proposed IC-based semantic similarity measures based on Wikipedia pages, concepts and neighbors. Wikipedia was both used as a structured taxonomy as well as a corpus to provide $IC$ values.
\subsubsection{\textbf{Combined knowledge-based methods:}}
Various similarity measures were proposed combining the various knowledge-based methods. Goa et al. \cite{GAO201580} proposed a semantic similarity method based on WordNet ontology where three different strategies are used to add weights to the edges and the shortest weighted path is used to measure the semantic similarity. According to the first strategy, the depths of all the terms in WordNet along the path between the two terms in consideration is added as a weight to the shortest path. In the second strategy, only the depth of the LCS of the terms was added as the weight, and in strategy three, the $IC$ value of the terms is added as weight. The shortest weighted path length is now calculated and then non-linearly transformed to produce semantic similarity measures. In comparison, it is shown that strategy three achieved a better correlation to the gold standards in comparison with traditional methods and the two other strategies proposed. Zhu and Iglesias \cite{7572993} proposed another weighted path measure called $wpath$ that adds the $IC$ value of the Least Common Subsumer as a weight to the shortest path length. $wpath$ is calculated as
\begin{equation}
    sim_{wpath}(t_1,t_2) = \frac{1}{1+ min\_len(t_1,t_2) * k^{IC_{t_{lcs}}}}
\end{equation}

This method was proposed to be used in various knowledge graphs (KG) like WordNet \cite{miller1995wordnet}, DBPedia \cite{bizer2009dbpedia},  YAGO \cite{hoffart2013yago2}, etc. and the parameter $k$ is a hyperparameter which has to be tuned for different KGs and different domains as different KGs have a different distribution of terms in each domain. Both corpus-based IC and intrinsic IC values were experimented and corpus IC-based $wpath$ measure achieved greater correlation in most of the gold standard datasets. 

Knowledge-based semantic similarity methods are computationally simple, and the underlying knowledge-base acts as a strong backbone for the models, and the most common problem of ambiguity like synonyms, idioms, and phrases are handled efficiently. Knowledge-based methods can easily be extended to calculate sentence to sentence similarity measure by defining rules for aggregation \cite{lee2011novel}. Lastra-D{\'\i}az et al. \cite{lastra2017hesml} developed a software Half-Edge Semantic Measures Library (HESML) to implement various ontology-based semantic similarity measures proposed and have shown an increase in performance time and scalability of the models.

However, knowledge-based systems are highly dependent on the underlying source resulting in the need to update them frequently which requires time and high computational resources. Although strong ontologies like WordNet, exist for the English language, similar resources are not available for other languages that results in the need for the building of strong and structured knowledge bases to implement knowledge-based methods in different languages and across different domains. Various research works were conducted on extending semantic similarity measures in the biomedical domain \cite{pedersen2007measures,biosses}. McInnes et al. \cite{mcinnes2013umls} built a domain-specific model called UMLS to measure the similarity between words in the biomedical domain. With nearly 6,500 world languages and numerous domains, this becomes a serious drawback for knowledge-based systems.

\section{Corpus-based Semantic-Similarity Methods}
Corpus-based semantic similarity methods measure semantic similarity between terms using the information retrieved from large corpora. The underlying principle called `distributional hypothesis' \cite{gorman2006scaling} exploits the idea that "similar words occur together, frequently"; however, the actual meaning of the words is not taken into consideration. While various techniques were used to construct the vector representation of the text data, several semantic distance measures based on the distributional hypothesis were proposed to estimate the similarity between the vectors. A comprehensive survey of various distributional semantic measures was carried out by Mohammad and Hurst \cite{mohammad2012distributional}, and the different measure and their respective formula are provided in Table \ref{tab:tab4} in Appendix \ref{appendix: A} . However, among all these measures, the cosine similarity gained significance and has been widely used among NLP researchers to date \cite{mohammad2012distributional}. In this section, we discuss in detail some of the widely used word-embeddings built using distributional hypothesis and some of the significant corpus-based semantic similarity methods.
\subsection{Word Embeddings}
Word embeddings provide vector representations of words wherein these vectors retain the underlying linguistic relationship between the words \cite{schnabel2015evaluation}. These vectors are computed using different approaches like neural networks \cite{mikolov2013efficient}, word co-occurrence matrix \cite{pennington2014glove}, or representations in terms of the context in which the word appears \cite{levy2014dependency}. Some of the most widely used pre-trained word embeddings include:
    \begin{itemize}
        \item \textbf{\textit{word2vec}} \cite{mikolov2013efficient}: Developed from Google News dataset, containing approximately 3 million vector representations of words and phrases, $word2vec$ is a neural network model used to produce distributed vector representation of words based on an underlying corpus. There are two different models of $word2vec$ proposed: the Continuous Bag of Words (CBOW) and the Skip-gram model. The architecture of the network is rather simple and contains an input layer, one hidden layer, and an output layer. The network is fed with a large text corpus as the input, and the output of the model is the vector representations of words. The CBOW model predicts the current word using the neighboring context words, while the Skip-gram model predicts the neighboring context words given a target word. $word2vec$ models are efficient in representing the words as vectors that retain the contextual similarity between words. The word vector calculations yielded good results in predicting the semantic similarity \cite{mikolov2013linguistic}. Many researchers extended the $word2vec$ model to propose context vectors \cite{context2vec}, dictionary vectors \cite{dict2vec}, sentence vectors \cite{pagliardini2018unsupervised} and paragraph vectors \cite{para2vec}.
        \item \textbf{\textit{GloVe}} \cite{pennington2014glove}: $GloVe$ developed by Stanford University relies on a global word co-occurrence matrix formed based on the underlying corpus. It estimates similarity based on the principle that words similar to each other occur together. The co-occurrence matrix is populated with occurrence values by doing a single pass over the underlying large corpora. $GloVe$ model was trained using five different corpora mostly Wikipedia dumps. While forming vectors, words are chosen within a specified context window owing to the fact that words far away have less relevance to the context word in consideration. The $GloVe$ loss function minimizes the least-square distance between the context window co-occurrence values and the global co-occurrence values \cite{LASTRADIAZ2019645}. $GloVe$ vectors were extended to form contextualized word vectors to differentiate words based on context \cite{mccann2017learned}. 
        \item \textbf{\textit{fastText}} \cite{bojanowski2017enriching}: Facebook AI researchers developed a word embedding model that builds word vectors based on Skip-gram models where each word is represented as a collection of character n-grams. $fastText$ learns word embeddings as the average of its character embeddings thus accounting for the morphological structure of the word which proves efficient in various languages like Finnish and Turkish. Even out-of-the-vocabulary words are assigned word vectors based on their characters or subunits.
        \item\textbf{\textit{Bidirectional Encoder Representations from Transformers(BERT)}} \cite{devlin2019bert}: Devlin et al. \cite{devlin2019bert} proposed a pretrained transformer-based word embeddings which can be fine-tuned by adding a final output layer to accommodate the embeddings to different NLP tasks. BERT uses the transformer architecture proposed by Vaswani et al. \cite{vaswani2017attention}, which produces attention-based word vectors using a bi-directional transformer encoder. The BERT framework involves two important processes namely ‘pre-training’ and ‘fine-tuning’. The model is pretrained using a corpus of nearly 3,300M words from both the Book corpus and English Wikipedia. Since the model is bidirectional in order to avoid the possibility of the model knowing the token itself when training from both directions the pretraining process is carried out in two different ways. In the first task, random words in the corpus are masked and the model is trained to predict these words. In the second task, the model is presented with sentence pairs from the corpus, in which 50 percent of the sentences are actually consecutive while the remaining are random pairs. The model is trained to predict if the given sentence pair are consecutive or not. In the ‘fine-tuning’ process, the model is trained for the specific down-stream NLP task at hand. The model is structured to take as input both single sentences and multiple sentences to accommodate a variety of NLP tasks. To train the model to perform a question answering task, the model is provided with various question-answer pairs and all the parameters are fine-tuned in accordance with the task. BERT embeddings provided state-of-the-art results in the STS-B data set with a Spearman's correlation of 86.5\% outperforming other BiLSTM models including ELMo \cite{peters2018deep}.
    \end{itemize}
Word embeddings are used to measure semantic similarity between texts of different languages by mapping the word embedding of one language over the vector space of another. On training with a limited yet sufficient number of translation pairs, the translation matrix can be computed to enable the overlap of embeddings across languages \cite{GLAVAS20181}. One of the major challenges faced when deploying word-embeddings to measure similarity is Meaning Conflation Deficiency. It denotes that word embeddings do not attribute to the different meanings of a word that pollutes the semantic space with noise by bringing irrelevant words closer to each other. For example, the words `finance' and `river' may appear in the same semantic space since the word `bank' has two different meanings \cite{camacho2018word}. It is critical to understand that word-embeddings exploit the distributional hypothesis for the construction of vectors and rely on large corpora, hence, they are classified under corpus-based semantic similarity methods. However, deep-neural network based-methods and most hybrid semantic similarity methods use word-embeddings to convert the text data to high dimensional vectors, and the efficiency of these embeddings plays a significant role in the performance of the semantic similarity methods \cite{mnih2013learning,levy2014neural}. 

\subsection{Types of corpus-based semantic similarity methods}
Based on the underlying methods using which the word-vectors are constructed there are a wide variety of corpus-based methods some of which are discussed in this section.
\subsubsection{\textbf{Latent Semantic Analysis (LSA) } \cite{landauer1997solution}:} LSA is one of the most popular and widely used corpus-based techniques used for measuring semantic similarity. A word co-occurrence matrix is formed where the rows represent the words and columns represent the paragraphs, and the cells are populated with word counts. This matrix is formed with a large underlying corpus, and dimensionality reduction is achieved by a mathematical technique called Singular Value Decomposition (SVD). SVD represents a given matrix as a product of three matrices, where two matrices represent the rows and columns as vectors derived from their eigenvalues and the third matrix is a diagonal matrix that has values that would reproduce the original matrix when multiplied with the other two matrices \cite{landauer1998introduction}. SVD reduces the number of columns while retaining the number of rows thereby preserving the similarity structure among the words. Then each word is represented as a vector using the values in its corresponding rows and semantic similarity is calculated as the cosine value between these vectors. LSA models are generalized by replacing words with texts and columns with different samples and are used to calculate the similarity between sentences, paragraphs, and documents.
\subsubsection{\textbf{Hyperspace Analogue to Language(HAL)} \cite{lund1996producing}:} HAL builds a word co-occurrence matrix that has both rows and columns representing the words in the vocabulary and the matrix elements are populated with association strength values. The association strength values are calculated by sliding a "window" the size of which can be varied, over the underlying corpus. The strength of association between the words in the window decreases with the increase in their distance from the focused word. For example, in the sentence "This is a survey of various semantic similarity measures", the words \textit{`survey'} and \textit{`variety'} have greater association value than the words \textit{`survey'} and \textit{`measures.'} Word vectors are formed by taking into consideration both the row and column of the given word. Dimensionality reduction is achieved by removing any columns with low entropy values. The semantic similarity is then calculated by measuring the Euclidean or Manhattan distance between the word vectors.
\subsubsection{\textbf{Explicit Semantic Analysis (ESA)} \cite{gabrilovich2007computing}:} ESA measures semantic similarity based on  Wiki-pedia concepts. The use of Wikipedia ensures that the proposed method can be used over various domains and languages. Since Wikipedia is constantly updated, the method is adaptable to the changes over time. First, each concept in Wikipedia is represented as an attribute vector of the words that occur in it, then an inverted index is formed, where each word is linked to all the concepts it is associated with. The association strength is weighted using the TF-IDF technique, and the concepts weakly associated with the words are removed. Thus the input text is represented by weighted vectors of concepts called the "interpretation vectors." Semantic similarity is measured by calculating the cosine similarity between these word vectors.
\subsubsection{\textbf{Word-Alignment models} \cite{sultan2015dls}:} Word-Alignment models calculate the semantic similarity of sentences based on their alignment over a large corpus \cite{sultan2014dls,kajiwara2016building,cer2017semeval}. The second, third, and fifth positions in SemEval tasks 2015 were secured by methods based on word alignment. The unsupervised method which was in the fifth place implemented the word alignment technique based on Paraphrase Database (PPDB)  \cite{ganitkevitch2013ppdb}. The system calculates the semantic similarity between two sentences as a proportion of the aligned context words in the sentences over the total words in both the sentences. The supervised methods which were at the second and third place used $word2vec$ to obtain the alignment of the words. In the first method, a sentence vector is formed by computing the "component-wise average" of the words in the sentence, and the cosine similarity between these sentence vectors is used as a measure of semantic similarity. The second supervised method takes into account only those words that have a contextual semantic similarity \cite{sultan2015dls}.
\subsubsection{\textbf{Latent Dirichlet Allocation (LDA)} \cite{SINOARA2019955}:} LDA is used to represent a topic or the general idea behind a document as a vector rather than every word in the document. This technique is widely used for topic modeling tasks and it has the advantage of reduced dimensionality considering that the topics are significantly less than the actual words in a document \cite{SINOARA2019955}. One of the novel approaches to determine document-to-document similarity is the use of vector representation of documents and calculate the cosine similarity between the vectors to ascertain the semantic similarity between documents \cite{BENEDETTI2019136}.
\subsubsection{\textbf{Normalised Google Distance} \cite{cilibrasi2007google}:} NGD measures the similarity between two terms based on the results obtained when the terms are queried using the Google search engine. It is based on the assumption that two words occur together more frequently in web-pages if they are more related. Give two terms $t_1$ and $t_2$ the following formula is used to calculate the NGD between the two terms.
\begin{equation}
    NGD(x,y) = \frac{max\;\{log\;f(t_1),log\;f(t_2)\}-log\;f(t_1,t_2)}{log\; G - min\;\{log\; f(t_1),log \;f(t_2)\}}
\end{equation}
where the functions $f(x)$ and $f(y)$ return the number of hits in Google search of the given terms, $f(x,y)$ returns the number of hits in Google search when the terms are searched together and $G$ represent the total number of pages in the overall google search. NGD is widely used to measure semantic relatedness rather than semantic similarity because related terms occur together more frequently in web pages though they may have opposite meaning.
\subsubsection{\textbf{Dependency-based models} \cite{agirre2009study}:} Dependency-based approaches ascertain the meaning of a given word or phrase using the neighbors of the word within a given window. The dependency-based models initially parse the corpus based on its distribution using Inductive Dependency Parsing \cite{nivre2006inductive}. For every given word a "syntactic context template" is built considering both the nodes preceding and succeeding the word in the built parse tree. For example, the phrase \textit{``thinks <term> delicious''} could have a context template as \textit{``pizza, burger, food''}. Vector representation of a word is formed by adding each window across the location that has the word in consideration, as it's root word, along with the frequency of the window of words appearing in the entire corpus. Once this vector is formed semantic similarity is calculated using cosine similarity between these vectors. Levy et al. \cite{levy2014dependency} proposed DEPS embedding as a word-embedding model based on dependency-based bag of words. This model was tested with the WS353 dataset where the task was to rank the similar words above the related words. On plotting a recall precision curve the DEPS curve showed greater affinity towards similarity rankings over BoW methods taken in comparison.
\subsubsection{\textbf{Kernel-based models} \cite{shawe2004kernel}:} Kernel-based methods were used to find patterns in text data thus enabling detecting similarity between text snippets. Two major types of kernels were used in text data namely the string or sequence kernel \cite{cancedda2003word} and the tree kernel \cite{moschitti2008tree}. Moschitti et al. \cite{moschitti2008tree} proposed tree kernels in 2007, that contains three different sub-structures in the tree kernel space namely a subtree - a tree whose root is not a leaf node along with its children nodes, a subset tree - a tree whose root is not a leaf node but not incorporating all its children nodes and does not break the grammatical rules, a partial tree - a tree structure closely similar to subset tree but it doesn’t always follow the grammatical rules. Tree kernels are widely used in identifying a structure in input sentences based on constituency or dependency, taking into consideration the grammatical rules of the language. Kernels are used by machine learning algorithms like Support Vector Machines(SVMs) to adapt to text data in various tasks like Semantic Role Labelling, Paraphrase Identification \cite{croce2017deep}, Answer Extraction \cite{moschitti2008kernels}, Question-Answer classification \cite{moschitti2007exploiting}, Relational text categorization \cite{moschitti2008kernel}, Answer Re-ranking in QA tasks \cite{severyn2012structural} and Relational text entailment \cite{moschitti2007fast}. Severyn et al. \cite{severyn2013learning} proposed a kernel-based semantic similarity method that represents the text directly as “structural objects” using Syntactic tree kernel \cite{collins2002new} and Partial tree kernels \cite{moschitti2006efficient}. The kernel function then combines the tree structures with semantic feature vectors from two of the best performing models in STS 2012 namely UKP \cite{bar2012ukp} and Takelab \cite{vsaric2012takelab} and some additional features including cosine similarity scores based on named entities, part of speech tags, and so on. The authors compare the performance of the model constructed using four different tree structures namely shallow tree, constituency tree, dependency tree, phrase-dependency tree, and the above-mentioned feature vectors. They establish that the tree kernel models perform better than all feature vectors combined. The model uses Support Vector Regression to obtain the final similarity score and it can be useful in various downstream NLP applications like question-answering, text-entailment extraction, etc. Amir et al. \cite{amir2017sentence} proposed another semantic similarity algorithm using kernel functions. They used constituency-based tree kernels where the sentence is broken down into subject, verb, and object based on the assumption most semantic properties of the sentence are attributed to these components. The input sentences are parsed using the Stanford Parser to extract various combinations of subject, verb, and object. The similarity between the various components of the given sentences is calculated using a knowledge base, and different averaging techniques are used to average the similarity values to estimate the overall similarity, and the best among them is chosen based on the root mean squared error value for a particular dataset. In recent research, deep learning methods have been used to replace the traditional machine learning models and efficiently use the structural integrity of kernels in the embedded feature extraction stage \cite{croce2017deep, collins2002convolution}. The model which achieved the best results in SemEval-2017 Task 1, proposed by Tian et al. \cite{tian2017ecnu} uses kernels to extract features from text data to calculate similarity. The model proposed an ensemble model that used both traditional NLP methods and deep learning methods. Two different features are namely the sentence pair matching features and single sentence features were used to predict the similarity values using regressors which added nonlinearity to the prediction. In single sentence feature extraction, dependency-based tree kernels are used to extract the dependency features in one given sentence, and in sentence pair matching features, constituency-based parse tree kernels are used to find the common sub-constructs among the three different characterizations of tree kernel spaces. The final similarity score is accessed by averaging the traditional NLP similarity value and the deep learning-based similarity value. The model achieved a Pearson's correlation of 73.16\% in the STS dataset.
\subsubsection{\textbf{Word-attention models} \cite{le2018acv}:}
 In most of the corpus-based methods all text components are considered to have equal significance; however, human interpretation of measuring similarity usually depends on keywords in a given context. Word attention models capture the importance of the words from underlying corpora \cite{LOPEZGAZPIO20191} before calculating the semantic similarity. Different techniques like word frequency, alignment, word association are used to capture the attention-weights of the text in consideration. Attention Constituency Vector Tree (ACV-Tree) proposed by Le et al. \cite{le2018acv} is similar to a parse tree where one word of a sentence is made the root and the remainder of the sentence is broken as a Noun Phrase (NP) and a Verb Phrase (VP). The nodes in the tree store three different attributes of the word into consideration: the word vector determined by an underlying corpus, the attention-weight, and the "modification-relations" of the word. The modification relations can be defined as the adjectives or adverbs that modify the meaning of another word. All three components are linked to form the representation of the word. A tree kernel function is used to determine the similarity between two words based on the equation below
 \begin{equation}
     TreeKernel(T_1,T_2) = \sum_{n_1\in N_{T_1}} \sum_{n_2\in N_{T_2}} \Delta (n_1,n_2)
 \end{equation}
 
\begin{equation}
    \Delta (n_1,n_2) = \Bigg\{ \begin{array}{l} 0, \mbox{ if } (n_1 \mbox{ and / or } n_2 \mbox{ are non-leaf-nodes) and }n_1 \ne n_2\\
    Aw \times SIM(vec_1,vec_2), \mbox{ if } n_1,n_2 \mbox{are leaf nodes}\\
    \mu( \lambda^2 + \sum_{p=1}^{l_m} \delta_p(c_{n_1},c_{n_2})), \mbox{ otherwise }
    \end{array}
\end{equation}

where $n_1,n_2$ represent the represents the nodes, $SIM(vec_1,vec_2)$ measures the cosine similarity between the vectors, $\delta_p(.)$ calculates the number of common subsequences of length $p$, $\lambda$, $\mu$ denote the decay factors for length of the child sequences and the height of the tree respectively, $c_{n_1},\mbox{ }c_{n_2}$ refer to the children nodes and $l_m = min (length(c_{n_1}),length( c_{n_2}) )$. The algorithm is tested using the STS benchmark datasets and has shown better performance in 12 out of 19 chosen STS Datasets \cite{le2018acv,8642425}.

Unlike knowledge-based systems, corpus-based systems are language and domain independent \cite{ALTINEL20181129}. Since they are dependent on statistical measures the methods can be easily adapted across various languages using an effective corpus. With the growth of the internet, building corpora of most languages or domains has become rather easy. Simple web crawling techniques can be used to build large corpora \cite{baroni2009wacky}. However, the corpus-based methods do not take into consideration the actual meaning of the words. The other challenge faced by corpus-based methods is the need to process the large corpora built, which is a rather time-consuming and resource-dependent task. Since the performance of the algorithms largely depends on the underlying corpus, building an efficient corpus is paramount. Though efforts are made by researchers to build a clean and efficient corpus like the C4 corpus built by web crawling and five steps to clean the corpus \cite{raffel2019exploring}, an "ideal corpus" is still not defined by researchers. 

\section{Deep Neural Network-based Methods}
Semantic similarity methods have exploited the recent developments in neural networks to enhance performance. The most widely used techniques include Convolutional Neural Networks (CNN), Long Short Term Memory (LSTM), Bidirectional Long Short Term Memory (Bi-LSTM), and Recursive Tree LSTM. Deep neural network models are built based on two fundamental operations: convolution and pooling. The convolution operation in text data may be defined as the sum of the element-wise product of a sentence vector and a weight matrix. Convolution operations are used for feature extraction. Pooling operations are used to eliminate features that have a negative impact, and only consider those feature values that have a considerable impact on the task at hand. There are different types of pooling operations and the most widely used is Max pooling, where only the maximum value in the given filter space is selected. This section describes some of the methods that deploy deep neural networks to estimate semantic similarity between text snippets. Although the methods described below exploit word embeddings built using large corpora, deep-neural networks are used to estimate the similarity between the word-embeddings, hence they are classified separately from corpus-based methods.
\subsection{Types of deep neural network-based semantic similarity methods:}
\begin{itemize}
    \item Wang et al. \cite{Wang2016} proposed a model to estimate semantic similarity between two sentences based on lexical decomposition and composition. The model uses $word2vec$ pretrained embeddings to form a vector representation of the sentences $s_1$ and $s_2$. A similarity matrix $M$ of dimension $i$ x $j$ is built where i and j are the number of words in sentence 1 ($S_1$) and sentence 2 ($S_2$) respectively. The cells of the matrix are populated with the cosine similarity between the words in the indices of the matrix. Three different functions are used to construct semantic matching vectors $\vec{s_1}$ and $\vec{s_2}$ , the global, local, and max function. The global function constructs the semantic matching vector of $S_1$ by taking the weighted sum of the vectors, of all the words in $S_2$, the local function, takes into consideration only word vectors within a given window size, and the max function takes only the vectors of the words, that have the maximum similarity. The second phase of the algorithm uses three different decomposition functions - rigid, linear, and orthogonal - to estimate the similarity component and the dissimilarity component between the sentence vectors and the semantic matching vectors. Both the similarity component and the dissimilarity component vectors are passed through a two-channel convolution layer followed by a single max-pooling layer. The similarity is then calculated using a sigmoid layer that estimates the similarity value within the range of 0 and 1. The model was tested using the QASent dataset \cite{Wang2007} and the WikiQA dataset \cite{Meek2018}. The two measures used to estimate the performance are mean average precision (MAP) and mean reciprocal rank (MRR). The model achieves the best MAP in the QASent dataset and the best MAP and MRR in the WikiQA dataset. Yang Shao \cite{shao2017hcti} proposed a semantic similarity algorithm that exploits, the recent development in neural networks using $GloVe$ word embeddings. Given two sentences, the model predicts a probability distribution over set semantic similarity values. The pre-processing steps involve the removal of punctuation, tokenization, and using $GloVe$ vectors to replace words with word embeddings. The length of the input is set to 30 words, which is achieved by removal or padding as deemed necessary. Some special hand-crafted features like flag values indicating if the words or numbers occurred in both the sentences and POS tagging one hot encoded values, were added to the $GloVe$ vectors. The vectors are then fed to a CNN with 300 filters and one max-pooling layer which is used to form the sentence vectors. ReLU activation function is used in the convolution layer. The semantic difference between the vectors is calculated by the element-wise absolute difference and the element-wise multiplication of the two, sentence-vectors generated. The vectors are further passed through two fully-connected layers, which predicts the probability distribution of the semantic similarity values. The model performance was evaluated using the SemEval datasets where the model was ranked 3rd in SemEval 2017 dataset track.
    \item The LSTM networks are a special kind of Recurrent Neural Networks (RNN). While processing text data, it is essential for the networks to remember previous words, to capture the context, and RNNs have the capacity to do so. However, not all the previous content has significance over the next word/phrase, hence RNNs suffer the drawback of long term dependency. LSTMs are designed to overcome this problem. LSTMs have gates which enable the network to choose the content it has to remember. For example, consider the text snippet, \textit{``Mary is from Finland. She is fluent in Finnish. She loves to travel.}'' While we reach the second sentence of the text snippet, it is essential to remember the words \textit{``Mary''} and \textit{``Finland.}'' However, on reaching the third sentence the network may forget the word \textit{``Finland.''} The architecture of LSTMs allows this. Many researchers use the LSTM architecture to measure semantic similarity between blocks of text. Tien et al. \cite{TIEN2019102090} uses a network combined with LSTM and CNN to form a sentence embedding from pretrained word embeddings followed by an LSTM architecture to predict their similarity. Tai et al. \cite{tai2015improved} proposed an LSTM architecture to estimate the semantic similarity between two given sentences. Initially, the sentences are converted to sentence representations using Tree-LSTM over the parse tree of the sentences. These sentence representations are then, fed to a neural network that calculates the absolute distance between the vectors and the angle between the vectors. The experiment was conducted using the SICK dataset, and the similarity measure varies with the range 1 to 5. The hidden layer consisted of 50 neurons and the final softmax layer classifies the sentences over the given range. The Tree-LSTM model achieved better Pearson's and Spearman's correlation in the gold standard datasets, than the other neural network models in comparison.
    \item He and Lin \cite{he-lin-2016-pairwise} proposed a hybrid architecture using Bi-LSTM and CNN to estimate the semantic similarity of the model. Bi-LSTMs have two LSTMs that run parallel, one from the beginning of the sentence and one from the end, thus capturing the entire context. In their model, He and Lin use Bi-LSTM for context modelling. A pairwise word interaction model is built that calculates a comparison unit between the vectors derived from the hidden states of the two LSTMs using the below formula
    \begin{equation}
        CoU (\vec{h_1},\vec{h_2}) =\{cos(\vec{h_1},\vec{h_2}), euc(\vec{h_1},\vec{h_2}),manh((\vec{h_1},\vec{h_2})\}
    \end{equation}
    where $\vec{h_1}$ and $\vec{h_2}$ represent the vectors from the hidden state of the LSTMs and the functions $cos()$, $euc()$, $manh()$ calculate the Cosine distance, Euclidean distance, and Manhattan distance, respectively. This model is similar to other recent neural network-based word attention models \cite{neuralnetwork,AlexanderM.Rush2015}. However, attention weights are not added, rather the distances are added as weights. The word interaction model is followed by a similarity focus layer where weights are added to the word interactions (calculated in the previous layers) based on their importance in determining the similarity. These re-weighted vectors are fed to the final convolution network. The network is composed of alternating spatial convolution layers and spatial max pooling layers, ReLU activation function is used and at the network ends with two fully connected layers followed by a LogSoftmax layer to obtain a non-linear solution. This model outperforms the previously mentioned Tree-LSTM model on the SICK dataset.
    \item Lopez-Gazpio et al. \cite{LOPEZGAZPIO20191} proposed an extension to the existing Decomposable Attention Model (DAM) proposed by Parikh et al. \cite{parikh2016decomposable} which was originally used for Natural Language Inference(NLI). NLI is used to categorize a given text block to a particular relation like entailment, neutral, or contradiction. The DAM model used feed-forward neural networks in three consecutive layers the attention layer, comparison layer, and aggregation layer. Given two sentences the attention layer produces two attention vectors for each sentence by finding the overlap between them. The comparison layer concatenates the attention vectors with the sentence vectors to form a single representative vector for each sentence. The final aggregation layer flattens the vectors and calculates the probability distribution over the given values. Lopez-Gazpio et al. \cite{LOPEZGAZPIO20191} used word n-grams to capture attention in the first layer instead of individual words. $n-grams$ maybe defined as a sequence of n words that are contiguous with the given word, n-grams are used to capture the context in various NLP tasks. In order to accommodate n-grams, a Recurrent Neural Network (RNN) is added to the attention layer. Variations were proposed by replacing RNN with Long-Term Short memory (LSTM) and Convolutional Neural Network (CNN). The model was used for semantic similarity calculations by replacing the final classes of entailment relationships with semantic similarity ranges from 0 to 5. The models achieved better performance in capturing the semantic similarity in the SICK dataset and the STS benchmark dataset when compared to DAM and other models like Sent2vec \cite{pagliardini2018unsupervised} and BiLSTM among others.
    \item\textbf{Transformer-based models:} Vaswani et al. \cite{vaswani2017attention} proposed a transformer model that relies on attention mechanisms to capture the semantic properties of words in the embeddings. The transformer has two parts ‘encoder’ and ‘decoder’. The encoder consists of layers of multi-head attention mechanisms followed by a fully connected feed-forward neural network. The decoder is similar to the encoder with one additional layer of multi-head attention which captures the attention weights in the output of the encoder. Although this model was proposed for the machine translation task, Devlin et al. \cite{devlin2019bert} used the transformer model to generate BERT word embeddings. Sun et al. \cite{sun2020ernie} proposed a multi-tasking framework using transformers called ERNIE 2.0. In this framework, the model is continuously pretrained i.e., when a new task is presented the model is fine-tuned to accommodate the new task while retaining the previously gained knowledge. The model outperformed BERT. XLNet proposed by Yang et al. \cite{yang2019xlnet} used an autoregression model as opposed to the autoencoder model and outperformed BERT and ERNIE 2.0. A number of variations of BERT models were proposed based on the corpus used to train the model and by optimizing the computational resources. Lan et al. \cite{lan2019albert} proposed ALBERT, with two techniques to reduce the computational complexity of BERT namely ‘factorized embedding parameterization’ and ‘cross-layer parameter sharing’. ALBERT outperformed all the above three models. Other variations of BERT models that use transformers include TinyBERT \cite{jiao2019tinybert}, RoBERTa \cite{liu2019roberta,sanh2019distilbert}, and a domain-specific variation trained on a scientific corpus with a focus on the BioMedical domain the SciBERT \cite{beltagy2019scibert}. Raffel et al. \cite{raffel2019exploring} proposed a transformer model with a well-defined corpus called ‘Colossal Clean Crawled Corpus’ or C4 to train the model named T5-11B. Unlike BERT they adopt a ‘text-to-text framework’ where the input sequence is attached with a token to identify the NLP task to be performed thus eliminating the two stages pre-training and fine-tuning. They propose five different versions of their model based on the number of trainable parameters each model has namely 1) T5-Small 2) T5-Base 3) T5-Large 4) T5-3B and 5)T511B and they have 60 million, 220 million, 770 million, 3 billion, and 11 billion parameters respectively. This model outperformed all other transformer-based models and achieved the state of the art results. As a result of their study, they confirm that the performance of the models increases with increased data and computational power and the performance can be further improved if larger models are built and it is important to note that in order to replicate their best model five GPUs are required among other resources. A compilation of the various transformer-based models and their Pearson's correlation on the STS-B dataset is provided below in Table \ref{tab:berttable}.
\end{itemize}
\begin{longtable}[c]{{|p{2cm}|p{6cm}|p{1cm}|p{2cm}|}}
\hline
Model Name & Title                                                                        & Year & Pearson's Correlation \\ \hline
\endfirsthead
\multicolumn{4}{c}%
{{\bfseries Table \thetable\ continued from previous page}} \\
\hline
Model Name & Title                                                                        & Year & Pearson's Correlation \\ \hline
\endhead
T5-11B     & Exploring the Limits of  Transfer Learning with a Unified Text-to-Text Transformer & 2019 & 0.925 \\ \hline
XLNet      & XLNet: Generalized Autoregressive Pretraining for Language Understanding      & 2019 & 0.925               \\ \hline
ALBERT     & ALBERT: A Lite BERT for Self-supervised Learning of Language Representations & 2019 & 0.925               \\ \hline
RoBERTa    & RoBERTa: A Robustly Optimized BERT Pretraining Approach                      & 2019 & 0.922               \\ \hline
ERNIE 2.0  & ERNIE 2.0: A Continual Pre-training Framework for Language Understanding     & 2019 & 0.912               \\ \hline
DistilBERT & DistilBERT, a distilled version of BERT: smaller, faster, cheaper and lighter      & 2019 & 0.907 \\ \hline
TinyBERT   & TinyBERT: Distilling BERT for Natural Language Understanding                 & 2019 & 0.799               \\ \hline
\caption{Pearson's Correlation of various transformer-based models on STS benchmark dataset.}
\label{tab:berttable}\\
\end{longtable}
Deep neural network-based methods outperform most of the traditional methods and the recent success of transformer-based models have served as a breakthrough in semantic similarity research. However, implementation of deep-learning models requires large computational resources, though variations of the models to minimize the computational resources are being proposed we see that the performance of the model takes a hit as well, for example, TinyBERT \cite{jiao2019tinybert}. And the performance of the models is largely increased by the use of a bigger corpus which again poses the challenge of building an ideal corpus. Most deep-learning models are "black-box" models and it is difficult to ascertain the features based on which the performance is achieved, hence it becomes difficult to be interpreted unlike in the case of corpus-based methods that have a strong mathematical foundation. Various fields like finance, insurance, etc., that deal with sensitive data may be reluctant to deploy deep neural network-based methods due to their lack of interpretability. 
\section{Hybrid Methods}
Based on all the previously discussed methods we see that each has its advantages and disadvantages. The knowledge-based methods exploit the underlying ontologies to disambiguate synonyms, while corpus-based methods are versatile as they can be used across languages. Deep neural network-based systems, though computationally expensive, provide better results. However, many researchers have found ways to exploit the best of each method and build hybrid models to measure semantic similarity. In this section, we describe the methodologies used in some of the widely used hybrid models.
\subsection{Types of hybrid semantic similarity methods:}
\begin{itemize}
    \item \textbf{Novel Approach to a Semantically-Aware Representation of Items (NASARI)} \cite{camacho2015nasari}: Camacho Collados et al. \cite{camacho2015nasari} proposed an approach the $NASARI$ where the knowledge source BabelNet is used to build a corpus based on which vector representation for concepts (words or group of words) are formed. Initially, the Wikipedia pages associated with a given concept, in this case, the \textit{synset} of BabelNet, and all the outgoing links from the given page are used to form a sub-corpus for the specific concept. The sub-corpus is further expanded with the Wikipedia pages of the hypernyms and hyponyms of the concept in the BabelNet network. The entire Wikipedia is considered as the reference corpus. Two different types of vector representation were proposed. In the first method, weighted vectors were formed using lexical specificity. Lexical specificity is a statistical method of identifying the most representative words for a given text, based on the hypergeometric distribution (sampling without replacement). Let "$T$ and $t$" denote the total content words in the reference corpus $RC$ and sub-corpus $SC$ respectively and "$F$ and $f$" denote the frequency of the given word in the reference corpus $RC$ and sub-corpus $SC$ respectively, then lexical specificity can be represented by the below equation
    \begin{equation}
    spec(T,t,F,f) = -log_{10} P(X\ge f)
    \end{equation}
    X represents a random variable that follows a hypergeometric relation with the parameters $T$, $t$ and $F$ and $P(X\ge f)$ is defined as,
    \begin{equation}
        P(X\ge f) = \sum_{i=f}^{F} P(X=i)
    \end{equation}
    $P(X=i)$ is the probability of a given term appearing exactly $i$ times in the given sub-corpus in hypergeometric distribution with $T$, $t$ and $F$. The second method forms a cluster of words in the sub-corpus that share a common hypernym in the WordNet taxonomy which is embedded in BabelNet. The specificity is then measured based on the frequency of the hypernym and all its hyponyms in the taxonomy, even those that did not occur in the given sub-corpus. This clustering technique forms a unified representation of the words that preserve the semantic properties. The specificity values are added as weights in both methods to rank the terms in a given text. The first method of vector representation was called $NASARI_{lexical}$ and the second method was called $NASARI_{unified}$. The similarity between these vectors is calculated using the measure called Weighted Overlap \cite{pilehvar2013align} as,
    \begin{equation}
        WO (v_1,v_2) = \sqrt{\frac{\sum_{d\in O}(rank(d,\vec{v_1}) + rank(d,\vec{v_2}))^{-1}}{\sum_{i=1}^{|O|(2i)^{-1}}}}
    \end{equation}
    where $O$ denotes the overlapping terms in each vector and $rank(d,\vec{v_i})$ represent the rank of the term $d$ in the vector $v_i$.
    
Camacho Collados et al. \cite{CAMACHOCOLLADOS201636} proposed an extension to their previous work and proposed a third vector representation by mapping the lexical vector to the semantic space of word embeddings produced by complex word embedding techniques like $word2vec$. This representation was called as $NASARI_{embedded}$. The similarity is measured as the cosine similarity between these vectors. All three methods were tested across the gold standard datasets M\&C, WS-Sim and SimLex-999. $NASARI_{lexical}$ achieved higher Pearson's and Spearman's correlation in average over the three datasets in comparison with other methods like ESA, $word2vec$, and $lin$.
    \item \textbf{Most Suitable Sense Annotation (MSSA)} \cite{RUAS2019288}: Ruas et al. proposed three different methodologies to form word-sense embeddings. Given a corpus, the word-sense disambiguation step is performed using one of the three proposed methods: Most Suitable Sense Annotation (MSSA), Most Suitable Sense Annotation N Refined (MSSA-NR), and Most Suitable Sense Annotation Dijkstra (MSSA-D). Given a corpus each word in the corpus is associated with a \textit{synset} in the WordNet ontology and \textit{"gloss-average-vector"} is calculated for each \textit{synset}. The gloss-average-vector is formed using the vector representation of the words in the gloss of each \textit{synset}.  MSSA calculates the gloss-average-vector using a small window of words and returns the \textit{synset} of the word which has the highest gloss-average-vector value. MSSA-D, however, considers the entire document from the first word to the last word and then determines the associated \textit{synset}. These two systems use Google News vectors\footnote{https://code.google.com/archive/p/word2vec/ .} to form the synset-embeddings. MSSA-NR is an iterative model, where the first pass produces the synset-embeddings, that are fed back in the second pass as a replacement to gloss-average-vectors to produce more refined synset-embeddings. These synset-embeddings are then fed to a $word2vec$ CBOW model to produce multi-sense word embeddings that are used to calculate the semantic similarity. This combination of MSSA variations and $word2vec$ produced solid results in gold standard datasets like R\&G, M\&C, WS353-Sim, and SimLex-999 \cite{RUAS2019288}.
    \item \textbf{Unsupervised Ensemble Semantic Textual Similarity Methods (UESTS)} \cite{hassan2019uests}: Hassan et al. proposed an ensemble semantic similarity method based on an underlying unsupervised word-aligner. The model calculates the semantic similarity as the weighted sum of four different semantic similarity measures between sentences $S_1$ and $S_2$ using the equation below 
    \begin{equation}
    \begin{aligned}
        sim_{USETS} (S_1,S_2) =  \alpha * sim_{WAL}(S_1,S_2) + \beta * sim_{SC}(S_1,S_2) \\ + \gamma * sim_{embed}(S_1,S_2) + \theta * sim_{ED}(S_1,S_2)
    \end{aligned}
    \end{equation}

    $sim_{WAL}(S_1,S_2)$ calculates similarity using a synset-based word aligner. The similarity between text is measured based on the number of shared neighbors each term has in the BableNet taxonomy. $sim_{SC}(S_1,S_2)$ measures similarity using soft cardinality measure between the terms in comparison. The soft cardinality function treats each word as a set and the similarity between them as an intersection between the sets. $sim_{embed}(S_1,S_2)$ forms word vector representations using the word embeddings proposed by Baroni et al. \cite{baroni2014don}. Then similarity is measured as the cosine value between the two vectors. $sim_{ED}(S_1,S_2)$ is a measure of dissimilarity between two given sentences. The edit distance is defined as the minimum number of edits it takes to convert one sentence to another. The edits may involve insertion, deletion, or substitution. $sim_{ED}(S_1,S_2)$ uses word-sense edit distance where word-senses are taken into consideration instead of actual words themselves. The hyperparameters $\alpha$, $\beta$, $\gamma$, and $\theta$ were tuned to values between 0 and 0.5 for different STS benchmark datasets. The ensemble model outperformed the STS benchmark unsupervised models in the 2017 SemEval series on various STS benchmark datasets.
\end{itemize}
Hybrid methods exploit both the structural efficiency offered by knowledge-based methods and the versatility of corpus-based methods. Many studies have been conducted to build multi-sense embeddings in order to incorporate the actual meaning of words into word vectors. Iacobacci et al. formed word embeddings called "Sensembed" by using BabelNet to form a sense annotated corpus and then using $word2vec$ to build word vectors thus having different vectors for different senses of the words. As we can see, hybrid models compensate for the shortcomings of one method by incorporating other methods. Hence the performance of hybrid methods is comparatively high. The first 5 places of SemEval 2017 semantic similarity tasks were awarded to ensemble models which clearly shows the shift in research towards hybrid models \cite{cer2017semeval}.

\section{Analysis of Survey} 
This section discusses the method used to build this survey article and provides an overview of the various research articles taken into consideration.
\subsection{Search Strategy:} 
\begin{figure}[h]
  \centering
  \begin{minipage}[b]{0.45\textwidth}
    \includegraphics[width=\textwidth]{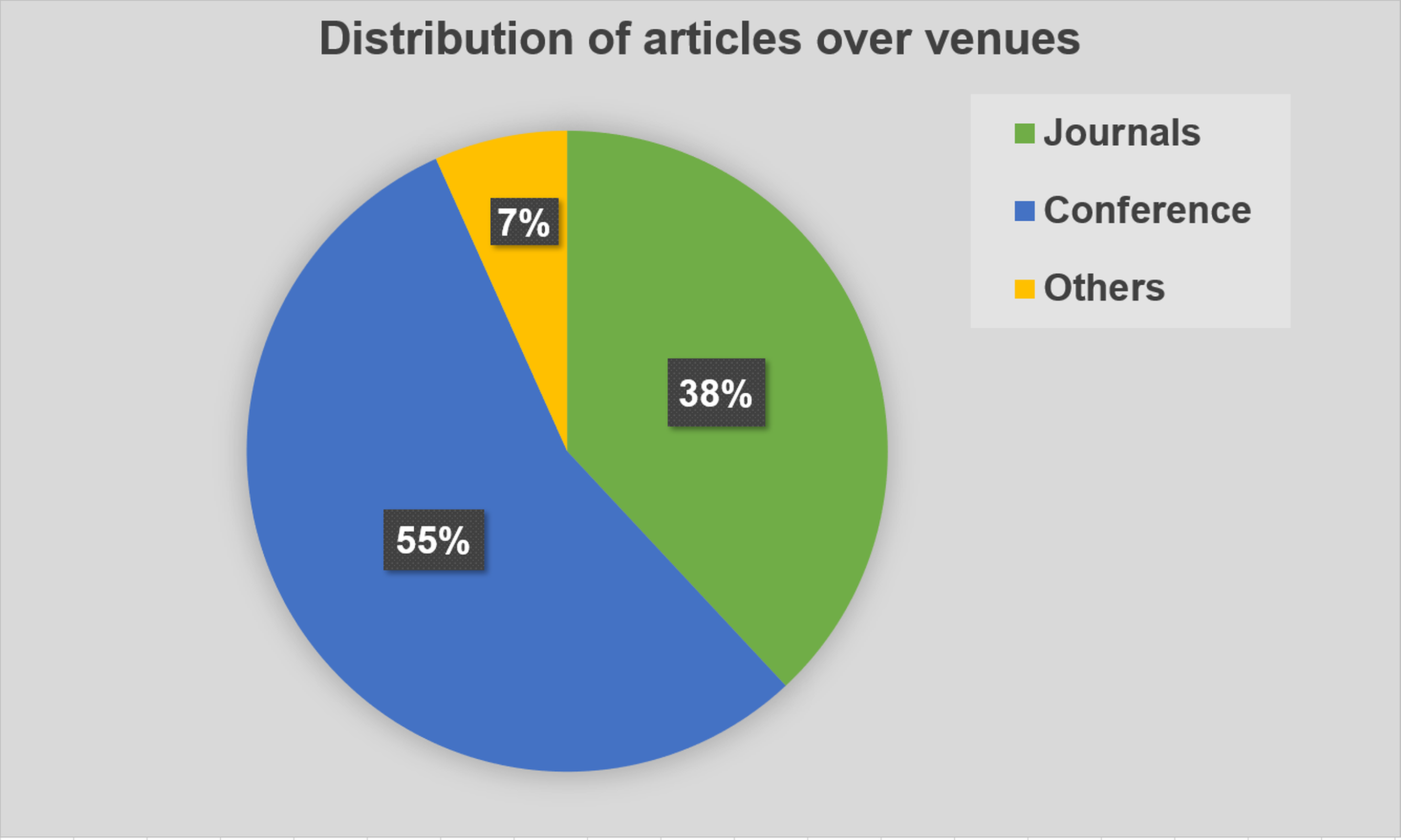}
    \caption{Distribution of articles over venues.}
    \label{fig:s1}
  \end{minipage}
  \hfill
  \begin{minipage}[b]{0.45\textwidth}
    \includegraphics[width=\textwidth]{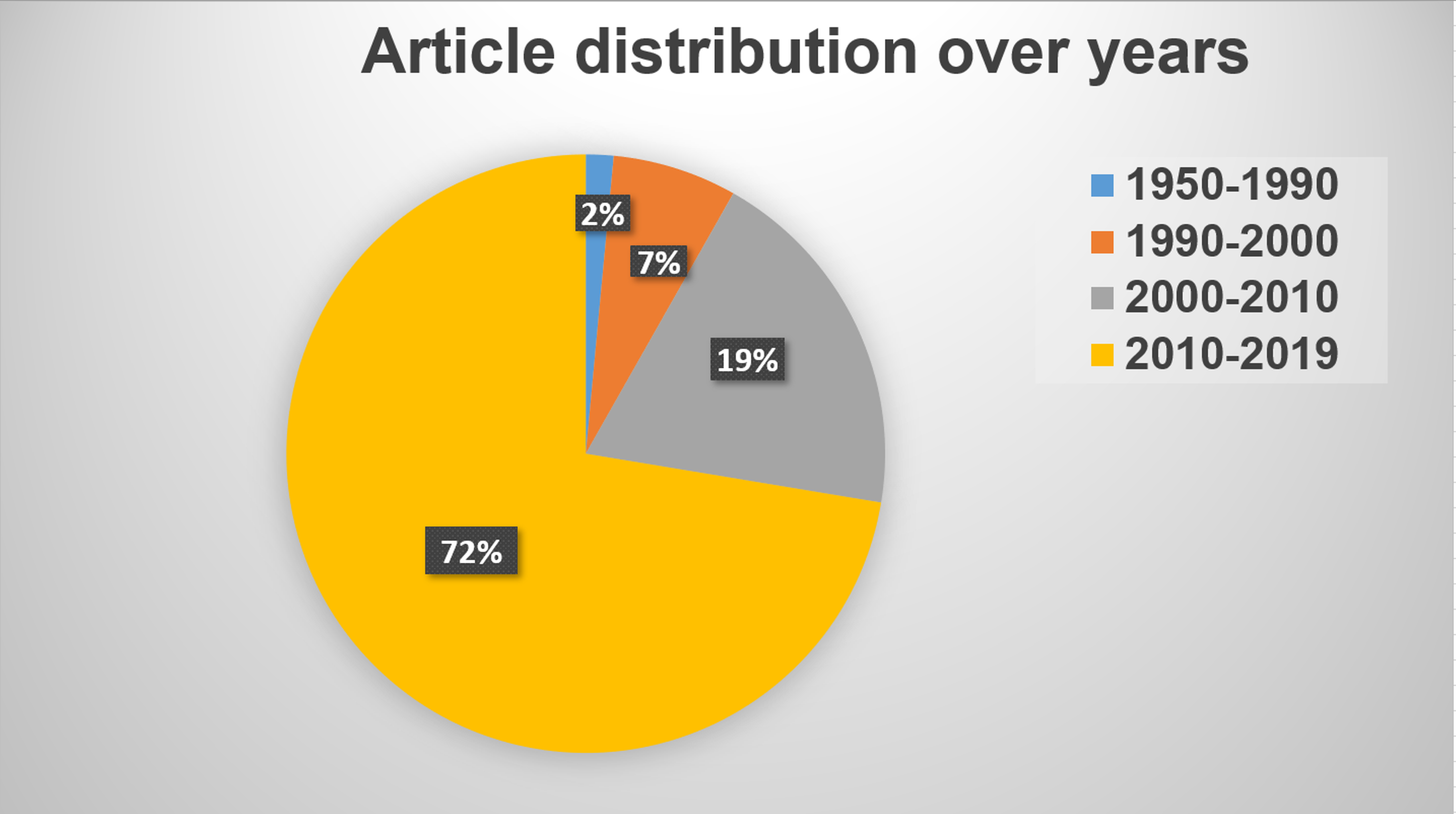}
    \caption{Distribution of articles over years.}
    \label{fig:s2}
  \end{minipage}
\end{figure}

\begin{figure}[h]
  \centering
  \begin{minipage}[b]{0.50\textwidth}
    \includegraphics[width=\textwidth]{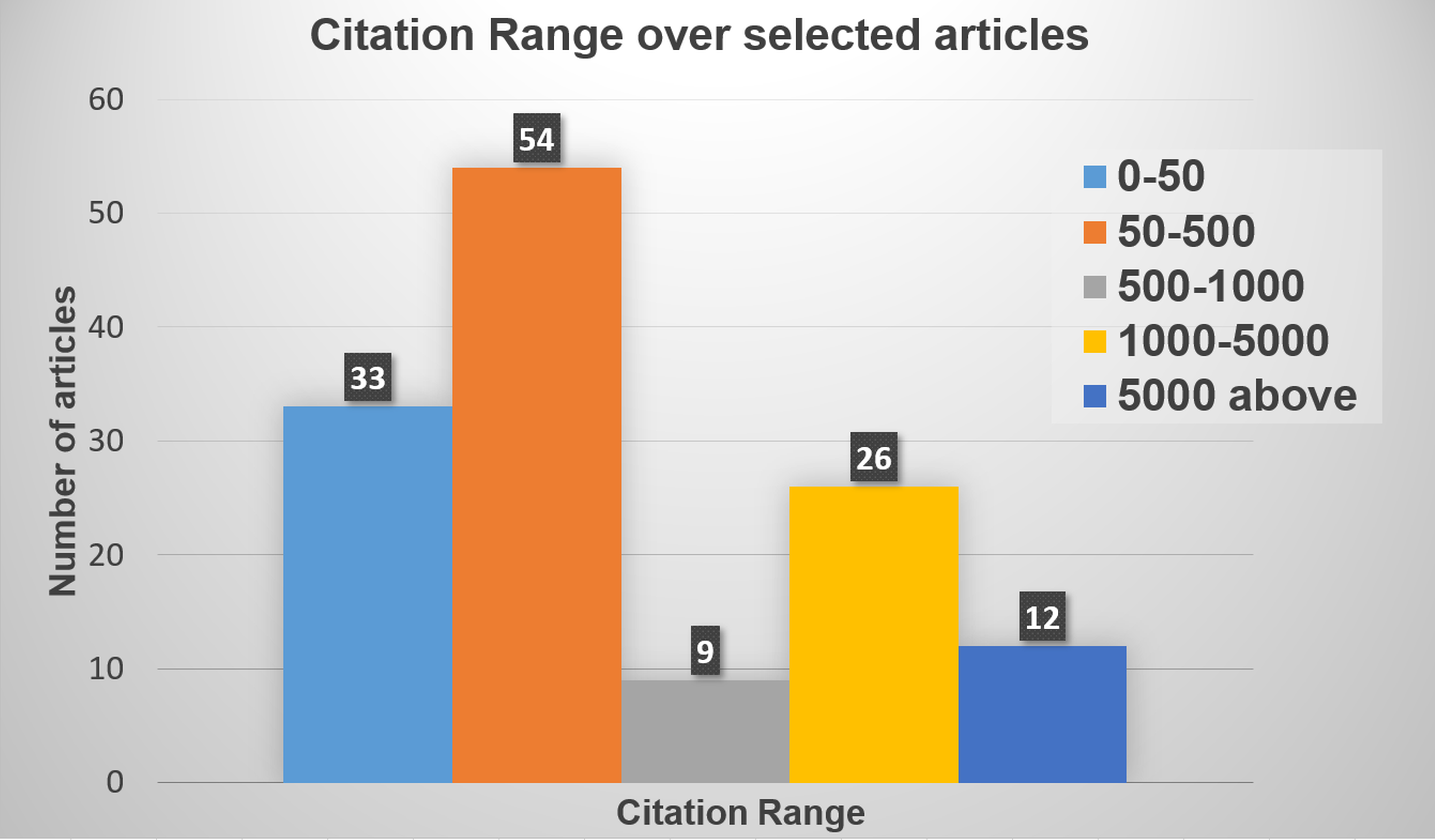}
    \caption{Distribution of citation range over the articles.\\}
    \label{fig:s3}
  \end{minipage}
  \hfill
  \begin{minipage}[b]{0.50\textwidth}
    \includegraphics[width=\textwidth]{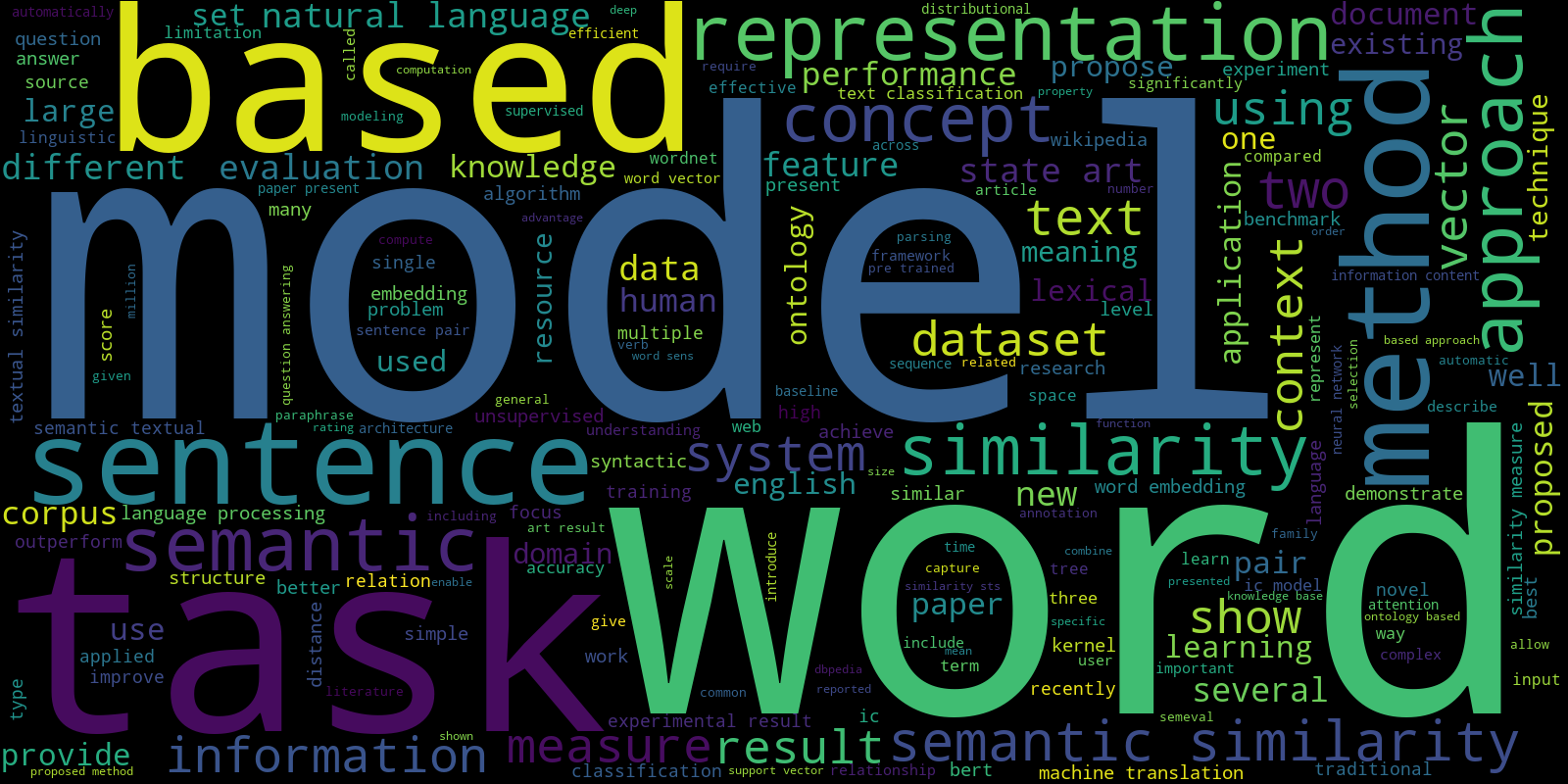}
    \caption{World cloud representing the collection of words from the abstracts of the papers used in the survey.}
    \label{fig:s4}
  \end{minipage}
\end{figure}

The articles considered for this survey were obtained using the Google Scholar search engine and the keywords used include \textit{``semantic similarity, word embedding, knowledge-based methods, corpus-based methods, deep neural network-based semantic similarity, LSTM, text processing, and semantic similarity datasets.''} The results of the search were fine-tuned using various parameters like the Journal Ranking, Google Scholar Index, number of citations, year of publication, etc. Only articles published in journals with Scimago Journal ranking of Quartile 1 and conferences that have a Google metrics H-index above 50 were considered. Exceptions were made for some articles that have a higher impact and relevance. The table of references sorted by the year of publication is included in Appendix \ref{appendix: B} as Table \ref{tab:my-table}. The table records 1) Title, 2) Year of Publication, 3) Author Names, 4) Venue, 5) SJR Quartile (for journals), 6) H-Index, and 7) Number of Citations (as of 02.04.2020). Some of the statistical results of the chosen articles are shown in the figures below. These figures highlight the quality of the articles chosen that in turn, highlights the quality of the survey.  Figure \ref{fig:s1} shows the distribution of the referenced articles over conferences, journals, and others. 55\% of the articles are from conferences and 38\% of the articles are from journals. The remaining 7\% of the articles are from arXiv and books. However, they have rather a high impact in relation to the topic of the survey. Figure \ref{fig:s2} highlights the distribution of the selected articles over the years. Nearly 72\% of the chosen articles are works carried out after 2010, the remaining 28\% of the articles represent the traditional methods adopted during the early stages of the evolution of semantic similarity. Figure \ref{fig:s3} represents the citation range of the articles. 54\% of the articles have 50 to 500 citations, 26\% have 1,000-5,000 citations, and 12\% of the article have more than 5000 citations. We see that 33\% of the articles have citations below 50 however, all these articles are published after the year 2017 which accounts for the fewer citations. 


\subsection{Word-cloud generation:} We implemented a simple python code to generate a word cloud using the abstracts from all the articles used in this survey. The abstracts from all the 118 articles were used to build a dataset that was then used in the python code. The extracted abstracts are initially pre-processed by converting the text to lower case, removing the punctuation, and removing the most commonly used English stop words available in the nltk\footnote{http://www.nltk.org/.} library. Then the word-cloud is built using the $wordcloud$ python library. The word cloud thus built is shown in Figure \ref{fig:word_cloud}. From the word cloud, we infer that though different keywords were used in our search for articles the general focus of the selected articles is semantic similarity. In a word cloud, the size of the words is proportional to the frequency of use of these words. The word \textit{``word''} is considerably bigger than the word \textit{``sentence''} showing that most of the research works focus on word-to-word similarity rather than sentence-to-sentence similarity. We could also infer that the words \textit{"vector"} and \textit{"representation"} have been used more frequently than the words \textit{"information"}, \textit{"context"}, and \textit{"concept"} indicating the influence of corpus-based methods over knowledge-based methods. With the given word cloud we showcase the focus of the survey graphically.

\section{Conclusion}
Measuring semantic similarity between two text snippets has been one of the most challenging tasks in the field of Natural Language Processing. Various methodologies have been proposed over the years to measure semantic similarity and this survey discusses the evolution, advantages, and disadvantages of these methods. Knowledge-based methods taken into consideration the actual meaning of text however, they are not adaptable across different domains and languages. Corpus-based methods have a statistical background and can be implemented across languages but they do not take into consideration the actual meaning of the text. Deep neural network-based methods show better performance, but they require high computational resources and lack interpretability. Hybrid methods are formed to take advantage of the benefits from different methods compensating for the shortcomings of each other. It is clear from the survey that each method has its advantages and disadvantages and it is difficult to choose one best model, however, most recent hybrid methods have shown promising results over other independent models. While the focus of recent research is shifted towards building more semantically aware word embeddings, and the transformer models have shown promising results, the need for determining a balance between computational efficiency and performance is still a work in progress. Research gaps can also be seen in areas such as building domain-specific word embeddings, addressing the need for an ideal corpus. This survey would serve as a good foundation for researchers who intend to find new methods to measure semantic similarity.

\begin{acks}
The authors would like to extend our gratitude to the research team in the DaTALab at Lakehead University for their support, in particular Abhijit Rao, Mohiuddin Qudar, Punardeep Sikka, and Andrew Heppner for their feedback and revisions on this publication. We would also like to thank Lakehead University, CASES, and the Ontario Council for Articulation and Transfer (ONCAT), without their support this research would not have been possible.
\end{acks}

\bibliographystyle{ACM-Reference-Format}
\bibliography{reference}


\begin{thebibliography}{134}


\ifx \showCODEN    \undefined \def \showCODEN     #1{\unskip}     \fi
\ifx \showDOI      \undefined \def \showDOI       #1{#1}\fi
\ifx \showISBNx    \undefined \def \showISBNx     #1{\unskip}     \fi
\ifx \showISBNxiii \undefined \def \showISBNxiii  #1{\unskip}     \fi
\ifx \showISSN     \undefined \def \showISSN      #1{\unskip}     \fi
\ifx \showLCCN     \undefined \def \showLCCN      #1{\unskip}     \fi
\ifx \shownote     \undefined \def \shownote      #1{#1}          \fi
\ifx \showarticletitle \undefined \def \showarticletitle #1{#1}   \fi
\ifx \showURL      \undefined \def \showURL       {\relax}        \fi
\providecommand\bibfield[2]{#2}
\providecommand\bibinfo[2]{#2}
\providecommand\natexlab[1]{#1}
\providecommand\showeprint[2][]{arXiv:#2}

\bibitem[\protect\citeauthoryear{Agirre, Alfonseca, Hall, Kravalova, Pasca, and
  Soroa}{Agirre et~al\mbox{.}}{2009}]%
        {agirre2009study}
\bibfield{author}{\bibinfo{person}{Eneko Agirre}, \bibinfo{person}{Enrique
  Alfonseca}, \bibinfo{person}{Keith Hall}, \bibinfo{person}{Jana Kravalova},
  \bibinfo{person}{Marius Pasca}, {and} \bibinfo{person}{Aitor Soroa}.}
  \bibinfo{year}{2009}\natexlab{}.
\newblock \showarticletitle{A Study on Similarity and Relatedness Using
  Distributional and WordNet-based Approaches}. In
  \bibinfo{booktitle}{\emph{Human Language Technologies: The 2009 Annual
  Conference of the North American Chapter of the Association for Computational
  Linguistics}}. Citeseer, \bibinfo{pages}{19}.
\newblock


\bibitem[\protect\citeauthoryear{Agirre, Banea, Cardie, Cer, Diab,
  Gonzalez-Agirre, Guo, Lopez-Gazpio, Maritxalar, Mihalcea,
  et~al\mbox{.}}{Agirre et~al\mbox{.}}{2015}]%
        {agirre2015semeval}
\bibfield{author}{\bibinfo{person}{Eneko Agirre}, \bibinfo{person}{Carmen
  Banea}, \bibinfo{person}{Claire Cardie}, \bibinfo{person}{Daniel Cer},
  \bibinfo{person}{Mona Diab}, \bibinfo{person}{Aitor Gonzalez-Agirre},
  \bibinfo{person}{Weiwei Guo}, \bibinfo{person}{Inigo Lopez-Gazpio},
  \bibinfo{person}{Montse Maritxalar}, \bibinfo{person}{Rada Mihalcea},
  {et~al\mbox{.}}} \bibinfo{year}{2015}\natexlab{}.
\newblock \showarticletitle{Semeval-2015 task 2: Semantic textual similarity,
  english, spanish and pilot on interpretability}. In
  \bibinfo{booktitle}{\emph{Proceedings of the 9th international workshop on
  semantic evaluation (SemEval 2015)}}. \bibinfo{pages}{252--263}.
\newblock


\bibitem[\protect\citeauthoryear{Agirre, Banea, Cardie, Cer, Diab,
  Gonzalez-Agirre, Guo, Mihalcea, Rigau, and Wiebe}{Agirre
  et~al\mbox{.}}{2014}]%
        {agirre2014semeval}
\bibfield{author}{\bibinfo{person}{Eneko Agirre}, \bibinfo{person}{Carmen
  Banea}, \bibinfo{person}{Claire Cardie}, \bibinfo{person}{Daniel Cer},
  \bibinfo{person}{Mona Diab}, \bibinfo{person}{Aitor Gonzalez-Agirre},
  \bibinfo{person}{Weiwei Guo}, \bibinfo{person}{Rada Mihalcea},
  \bibinfo{person}{German Rigau}, {and} \bibinfo{person}{Janyce Wiebe}.}
  \bibinfo{year}{2014}\natexlab{}.
\newblock \showarticletitle{Semeval-2014 task 10: Multilingual semantic textual
  similarity}. In \bibinfo{booktitle}{\emph{Proceedings of the 8th
  international workshop on semantic evaluation (SemEval 2014)}}.
  \bibinfo{pages}{81--91}.
\newblock


\bibitem[\protect\citeauthoryear{Agirre, Banea, Cer, Diab, Gonzalez~Agirre,
  Mihalcea, Rigau~Claramunt, and Wiebe}{Agirre et~al\mbox{.}}{2016}]%
        {agirre2016semeval}
\bibfield{author}{\bibinfo{person}{Eneko Agirre}, \bibinfo{person}{Carmen
  Banea}, \bibinfo{person}{Daniel Cer}, \bibinfo{person}{Mona Diab},
  \bibinfo{person}{Aitor Gonzalez~Agirre}, \bibinfo{person}{Rada Mihalcea},
  \bibinfo{person}{German Rigau~Claramunt}, {and} \bibinfo{person}{Janyce
  Wiebe}.} \bibinfo{year}{2016}\natexlab{}.
\newblock \showarticletitle{Semeval-2016 task 1: Semantic textual similarity,
  monolingual and cross-lingual evaluation}. In
  \bibinfo{booktitle}{\emph{SemEval-2016. 10th International Workshop on
  Semantic Evaluation; 2016 Jun 16-17; San Diego, CA. Stroudsburg (PA): ACL;
  2016. p. 497-511.}} ACL (Association for Computational Linguistics).
\newblock


\bibitem[\protect\citeauthoryear{Agirre, Cer, Diab, and Gonzalez-Agirre}{Agirre
  et~al\mbox{.}}{2012}]%
        {agirre2012semeval}
\bibfield{author}{\bibinfo{person}{Eneko Agirre}, \bibinfo{person}{Daniel Cer},
  \bibinfo{person}{Mona Diab}, {and} \bibinfo{person}{Aitor Gonzalez-Agirre}.}
  \bibinfo{year}{2012}\natexlab{}.
\newblock \showarticletitle{Semeval-2012 task 6: A pilot on semantic textual
  similarity}. In \bibinfo{booktitle}{\emph{* SEM 2012: The First Joint
  Conference on Lexical and Computational Semantics--Volume 1: Proceedings of
  the main conference and the shared task, and Volume 2: Proceedings of the
  Sixth International Workshop on Semantic Evaluation (SemEval 2012)}}.
  \bibinfo{pages}{385--393}.
\newblock


\bibitem[\protect\citeauthoryear{Agirre, Cer, Diab, Gonzalez-Agirre, and
  Guo}{Agirre et~al\mbox{.}}{2013}]%
        {agirre2013sem}
\bibfield{author}{\bibinfo{person}{Eneko Agirre}, \bibinfo{person}{Daniel Cer},
  \bibinfo{person}{Mona Diab}, \bibinfo{person}{Aitor Gonzalez-Agirre}, {and}
  \bibinfo{person}{Weiwei Guo}.} \bibinfo{year}{2013}\natexlab{}.
\newblock \showarticletitle{* SEM 2013 shared task: Semantic textual
  similarity}. In \bibinfo{booktitle}{\emph{Second Joint Conference on Lexical
  and Computational Semantics (* SEM), Volume 1: Proceedings of the Main
  Conference and the Shared Task: Semantic Textual Similarity}}.
  \bibinfo{pages}{32--43}.
\newblock


\bibitem[\protect\citeauthoryear{{Alexander M. Rush}, {Sumit Chopra}, and
  {Jason Weston}}{{Alexander M. Rush} et~al\mbox{.}}{2015}]%
        {AlexanderM.Rush2015}
\bibfield{author}{\bibinfo{person}{{Alexander M. Rush}},
  \bibinfo{person}{{Sumit Chopra}}, {and} \bibinfo{person}{{Jason Weston}}.}
  \bibinfo{year}{2015}\natexlab{}.
\newblock \showarticletitle{{A Neural Attention Model for Abstractive
  Sentence}}.
\newblock \bibinfo{journal}{\emph{Proceedings of the 2015 conference on
  empirical methods in natural language processing.}} \bibinfo{volume}{5},
  \bibinfo{number}{3} (\bibinfo{year}{2015}), \bibinfo{pages}{379--389}.
\newblock
\showISSN{2302-4496}


\bibitem[\protect\citeauthoryear{Altınel and Ganiz}{Altınel and
  Ganiz}{2018}]%
        {ALTINEL20181129}
\bibfield{author}{\bibinfo{person}{Berna Altınel} {and}
  \bibinfo{person}{Murat~Can Ganiz}.} \bibinfo{year}{2018}\natexlab{}.
\newblock \showarticletitle{Semantic text classification: A survey of past and
  recent advances}.
\newblock \bibinfo{journal}{\emph{Information Processing \& Management}}
  \bibinfo{volume}{54}, \bibinfo{number}{6} (\bibinfo{year}{2018}),
  \bibinfo{pages}{1129 -- 1153}.
\newblock
\showISSN{0306-4573}
\urldef\tempurl%
\url{https://doi.org/10.1016/j.ipm.2018.08.001}
\showDOI{\tempurl}


\bibitem[\protect\citeauthoryear{Amir, Tanasescu, and Zighed}{Amir
  et~al\mbox{.}}{2017}]%
        {amir2017sentence}
\bibfield{author}{\bibinfo{person}{Samir Amir}, \bibinfo{person}{Adrian
  Tanasescu}, {and} \bibinfo{person}{Djamel~A Zighed}.}
  \bibinfo{year}{2017}\natexlab{}.
\newblock \showarticletitle{Sentence similarity based on semantic kernels for
  intelligent text retrieval}.
\newblock \bibinfo{journal}{\emph{Journal of Intelligent Information Systems}}
  \bibinfo{volume}{48}, \bibinfo{number}{3} (\bibinfo{year}{2017}),
  \bibinfo{pages}{675--689}.
\newblock


\bibitem[\protect\citeauthoryear{Bahdanau, Cho, and Bengio}{Bahdanau
  et~al\mbox{.}}{2015}]%
        {neuralnetwork}
\bibfield{author}{\bibinfo{person}{Dzmitry Bahdanau},
  \bibinfo{person}{Kyunghyun Cho}, {and} \bibinfo{person}{Yoshua Bengio}.}
  \bibinfo{year}{2015}\natexlab{}.
\newblock \showarticletitle{Neural machine translation by jointly learning to
  align and translate}.
\newblock
\newblock
\shownote{3rd International Conference on Learning Representations, ICLR 2015 ;
  Conference date: 07-05-2015 Through 09-05-2015.}


\bibitem[\protect\citeauthoryear{Banerjee and Pedersen}{Banerjee and
  Pedersen}{2003}]%
        {banerjee2003extended}
\bibfield{author}{\bibinfo{person}{Satanjeev Banerjee} {and}
  \bibinfo{person}{Ted Pedersen}.} \bibinfo{year}{2003}\natexlab{}.
\newblock \showarticletitle{Extended gloss overlaps as a measure of semantic
  relatedness}. In \bibinfo{booktitle}{\emph{Ijcai}}, Vol.~\bibinfo{volume}{3}.
  \bibinfo{pages}{805--810}.
\newblock


\bibitem[\protect\citeauthoryear{B{\"a}r, Biemann, Gurevych, and Zesch}{B{\"a}r
  et~al\mbox{.}}{2012}]%
        {bar2012ukp}
\bibfield{author}{\bibinfo{person}{Daniel B{\"a}r}, \bibinfo{person}{Chris
  Biemann}, \bibinfo{person}{Iryna Gurevych}, {and} \bibinfo{person}{Torsten
  Zesch}.} \bibinfo{year}{2012}\natexlab{}.
\newblock \showarticletitle{Ukp: Computing semantic textual similarity by
  combining multiple content similarity measures}. In
  \bibinfo{booktitle}{\emph{* SEM 2012: The First Joint Conference on Lexical
  and Computational Semantics--Volume 1: Proceedings of the main conference and
  the shared task, and Volume 2: Proceedings of the Sixth International
  Workshop on Semantic Evaluation (SemEval 2012)}}. \bibinfo{pages}{435--440}.
\newblock


\bibitem[\protect\citeauthoryear{Baroni, Bernardini, Ferraresi, and
  Zanchetta}{Baroni et~al\mbox{.}}{2009}]%
        {baroni2009wacky}
\bibfield{author}{\bibinfo{person}{Marco Baroni}, \bibinfo{person}{Silvia
  Bernardini}, \bibinfo{person}{Adriano Ferraresi}, {and} \bibinfo{person}{Eros
  Zanchetta}.} \bibinfo{year}{2009}\natexlab{}.
\newblock \showarticletitle{The WaCky wide web: a collection of very large
  linguistically processed web-crawled corpora}.
\newblock \bibinfo{journal}{\emph{Language resources and evaluation}}
  \bibinfo{volume}{43}, \bibinfo{number}{3} (\bibinfo{year}{2009}),
  \bibinfo{pages}{209--226}.
\newblock


\bibitem[\protect\citeauthoryear{Baroni, Dinu, and Kruszewski}{Baroni
  et~al\mbox{.}}{2014}]%
        {baroni2014don}
\bibfield{author}{\bibinfo{person}{Marco Baroni}, \bibinfo{person}{Georgiana
  Dinu}, {and} \bibinfo{person}{Germ{\'a}n Kruszewski}.}
  \bibinfo{year}{2014}\natexlab{}.
\newblock \showarticletitle{Don’t count, predict! a systematic comparison of
  context-counting vs. context-predicting semantic vectors}. In
  \bibinfo{booktitle}{\emph{Proceedings of the 52nd Annual Meeting of the
  Association for Computational Linguistics (Volume 1: Long Papers)}}.
  \bibinfo{pages}{238--247}.
\newblock


\bibitem[\protect\citeauthoryear{Beltagy, Lo, and Cohan}{Beltagy
  et~al\mbox{.}}{2019}]%
        {beltagy2019scibert}
\bibfield{author}{\bibinfo{person}{Iz Beltagy}, \bibinfo{person}{Kyle Lo},
  {and} \bibinfo{person}{Arman Cohan}.} \bibinfo{year}{2019}\natexlab{}.
\newblock \showarticletitle{SciBERT: A Pretrained Language Model for Scientific
  Text}. In \bibinfo{booktitle}{\emph{Proceedings of the 2019 Conference on
  Empirical Methods in Natural Language Processing and the 9th International
  Joint Conference on Natural Language Processing (EMNLP-IJCNLP)}}.
  \bibinfo{pages}{3606--3611}.
\newblock


\bibitem[\protect\citeauthoryear{Benedetti, Beneventano, Bergamaschi, and
  Simonini}{Benedetti et~al\mbox{.}}{2019}]%
        {BENEDETTI2019136}
\bibfield{author}{\bibinfo{person}{Fabio Benedetti}, \bibinfo{person}{Domenico
  Beneventano}, \bibinfo{person}{Sonia Bergamaschi}, {and}
  \bibinfo{person}{Giovanni Simonini}.} \bibinfo{year}{2019}\natexlab{}.
\newblock \showarticletitle{Computing inter-document similarity with Context
  Semantic Analysis}.
\newblock \bibinfo{journal}{\emph{Information Systems}}  \bibinfo{volume}{80}
  (\bibinfo{year}{2019}), \bibinfo{pages}{136 -- 147}.
\newblock
\showISSN{0306-4379}
\urldef\tempurl%
\url{https://doi.org/10.1016/j.is.2018.02.009}
\showDOI{\tempurl}


\bibitem[\protect\citeauthoryear{Bizer, Lehmann, Kobilarov, Auer, Becker,
  Cyganiak, and Hellmann}{Bizer et~al\mbox{.}}{2009}]%
        {bizer2009dbpedia}
\bibfield{author}{\bibinfo{person}{Christian Bizer}, \bibinfo{person}{Jens
  Lehmann}, \bibinfo{person}{Georgi Kobilarov}, \bibinfo{person}{S{\"o}ren
  Auer}, \bibinfo{person}{Christian Becker}, \bibinfo{person}{Richard
  Cyganiak}, {and} \bibinfo{person}{Sebastian Hellmann}.}
  \bibinfo{year}{2009}\natexlab{}.
\newblock \showarticletitle{DBpedia-A crystallization point for the Web of
  Data}.
\newblock \bibinfo{journal}{\emph{Journal of web semantics}}
  \bibinfo{volume}{7}, \bibinfo{number}{3} (\bibinfo{year}{2009}),
  \bibinfo{pages}{154--165}.
\newblock


\bibitem[\protect\citeauthoryear{Bojanowski, Grave, Joulin, and
  Mikolov}{Bojanowski et~al\mbox{.}}{2017}]%
        {bojanowski2017enriching}
\bibfield{author}{\bibinfo{person}{Piotr Bojanowski}, \bibinfo{person}{Edouard
  Grave}, \bibinfo{person}{Armand Joulin}, {and} \bibinfo{person}{Tomas
  Mikolov}.} \bibinfo{year}{2017}\natexlab{}.
\newblock \showarticletitle{Enriching word vectors with subword information}.
\newblock \bibinfo{journal}{\emph{Transactions of the Association for
  Computational Linguistics}}  \bibinfo{volume}{5} (\bibinfo{year}{2017}),
  \bibinfo{pages}{135--146}.
\newblock


\bibitem[\protect\citeauthoryear{Bordes, Chopra, and Weston}{Bordes
  et~al\mbox{.}}{2014}]%
        {bordes2014question}
\bibfield{author}{\bibinfo{person}{Antoine Bordes}, \bibinfo{person}{Sumit
  Chopra}, {and} \bibinfo{person}{Jason Weston}.}
  \bibinfo{year}{2014}\natexlab{}.
\newblock \showarticletitle{Question Answering with Subgraph Embeddings}. In
  \bibinfo{booktitle}{\emph{Proceedings of the 2014 Conference on Empirical
  Methods in Natural Language Processing (EMNLP)}}. \bibinfo{pages}{615--620}.
\newblock


\bibitem[\protect\citeauthoryear{Camacho-Collados and
  Pilehvar}{Camacho-Collados and Pilehvar}{2018}]%
        {camacho2018word}
\bibfield{author}{\bibinfo{person}{Jose Camacho-Collados} {and}
  \bibinfo{person}{Mohammad~Taher Pilehvar}.} \bibinfo{year}{2018}\natexlab{}.
\newblock \showarticletitle{From word to sense embeddings: A survey on vector
  representations of meaning}.
\newblock \bibinfo{journal}{\emph{Journal of Artificial Intelligence Research}}
   \bibinfo{volume}{63} (\bibinfo{year}{2018}), \bibinfo{pages}{743--788}.
\newblock


\bibitem[\protect\citeauthoryear{Camacho-Collados, Pilehvar, and
  Navigli}{Camacho-Collados et~al\mbox{.}}{2015}]%
        {camacho2015nasari}
\bibfield{author}{\bibinfo{person}{Jos{\'e} Camacho-Collados},
  \bibinfo{person}{Mohammad~Taher Pilehvar}, {and} \bibinfo{person}{Roberto
  Navigli}.} \bibinfo{year}{2015}\natexlab{}.
\newblock \showarticletitle{Nasari: a novel approach to a semantically-aware
  representation of items}. In \bibinfo{booktitle}{\emph{Proceedings of the
  2015 Conference of the North American Chapter of the Association for
  Computational Linguistics: Human Language Technologies}}.
  \bibinfo{pages}{567--577}.
\newblock


\bibitem[\protect\citeauthoryear{Camacho-Collados, Pilehvar, and
  Navigli}{Camacho-Collados et~al\mbox{.}}{2016}]%
        {CAMACHOCOLLADOS201636}
\bibfield{author}{\bibinfo{person}{José Camacho-Collados},
  \bibinfo{person}{Mohammad~Taher Pilehvar}, {and} \bibinfo{person}{Roberto
  Navigli}.} \bibinfo{year}{2016}\natexlab{}.
\newblock \showarticletitle{Nasari: Integrating explicit knowledge and corpus
  statistics for a multilingual representation of concepts and entities}.
\newblock \bibinfo{journal}{\emph{Artificial Intelligence}}
  \bibinfo{volume}{240} (\bibinfo{year}{2016}), \bibinfo{pages}{36 -- 64}.
\newblock
\showISSN{0004-3702}
\urldef\tempurl%
\url{https://doi.org/10.1016/j.artint.2016.07.005}
\showDOI{\tempurl}


\bibitem[\protect\citeauthoryear{Cancedda, Gaussier, Goutte, and
  Renders}{Cancedda et~al\mbox{.}}{2003}]%
        {cancedda2003word}
\bibfield{author}{\bibinfo{person}{Nicola Cancedda}, \bibinfo{person}{Eric
  Gaussier}, \bibinfo{person}{Cyril Goutte}, {and} \bibinfo{person}{Jean-Michel
  Renders}.} \bibinfo{year}{2003}\natexlab{}.
\newblock \showarticletitle{Word-sequence kernels}.
\newblock \bibinfo{journal}{\emph{Journal of machine learning research}}
  \bibinfo{volume}{3}, \bibinfo{number}{Feb} (\bibinfo{year}{2003}),
  \bibinfo{pages}{1059--1082}.
\newblock


\bibitem[\protect\citeauthoryear{Cer, Diab, Agirre, Lopez-Gazpio, and
  Specia}{Cer et~al\mbox{.}}{2017}]%
        {cer2017semeval}
\bibfield{author}{\bibinfo{person}{Daniel Cer}, \bibinfo{person}{Mona Diab},
  \bibinfo{person}{Eneko Agirre}, \bibinfo{person}{I{\~n}igo Lopez-Gazpio},
  {and} \bibinfo{person}{Lucia Specia}.} \bibinfo{year}{2017}\natexlab{}.
\newblock \showarticletitle{SemEval-2017 Task 1: Semantic Textual Similarity
  Multilingual and Crosslingual Focused Evaluation}. In
  \bibinfo{booktitle}{\emph{Proceedings of the 11th International Workshop on
  Semantic Evaluation (SemEval-2017)}}. \bibinfo{pages}{1--14}.
\newblock


\bibitem[\protect\citeauthoryear{Cilibrasi and Vitanyi}{Cilibrasi and
  Vitanyi}{2007}]%
        {cilibrasi2007google}
\bibfield{author}{\bibinfo{person}{Rudi~L Cilibrasi} {and}
  \bibinfo{person}{Paul~MB Vitanyi}.} \bibinfo{year}{2007}\natexlab{}.
\newblock \showarticletitle{The google similarity distance}.
\newblock \bibinfo{journal}{\emph{IEEE Transactions on knowledge and data
  engineering}} \bibinfo{volume}{19}, \bibinfo{number}{3}
  (\bibinfo{year}{2007}), \bibinfo{pages}{370--383}.
\newblock


\bibitem[\protect\citeauthoryear{Collins and Duffy}{Collins and Duffy}{2002a}]%
        {collins2002convolution}
\bibfield{author}{\bibinfo{person}{Michael Collins} {and}
  \bibinfo{person}{Nigel Duffy}.} \bibinfo{year}{2002}\natexlab{a}.
\newblock \showarticletitle{Convolution kernels for natural language}. In
  \bibinfo{booktitle}{\emph{Advances in neural information processing
  systems}}. \bibinfo{pages}{625--632}.
\newblock


\bibitem[\protect\citeauthoryear{Collins and Duffy}{Collins and Duffy}{2002b}]%
        {collins2002new}
\bibfield{author}{\bibinfo{person}{Michael Collins} {and}
  \bibinfo{person}{Nigel Duffy}.} \bibinfo{year}{2002}\natexlab{b}.
\newblock \showarticletitle{New ranking algorithms for parsing and tagging:
  Kernels over discrete structures, and the voted perceptron}. In
  \bibinfo{booktitle}{\emph{Proceedings of the 40th Annual Meeting of the
  Association for Computational Linguistics}}. \bibinfo{pages}{263--270}.
\newblock


\bibitem[\protect\citeauthoryear{Croce, Filice, Castellucci, and Basili}{Croce
  et~al\mbox{.}}{2017}]%
        {croce2017deep}
\bibfield{author}{\bibinfo{person}{Danilo Croce}, \bibinfo{person}{Simone
  Filice}, \bibinfo{person}{Giuseppe Castellucci}, {and}
  \bibinfo{person}{Roberto Basili}.} \bibinfo{year}{2017}\natexlab{}.
\newblock \showarticletitle{Deep learning in semantic kernel spaces}. In
  \bibinfo{booktitle}{\emph{Proceedings of the 55th Annual Meeting of the
  Association for Computational Linguistics (Volume 1: Long Papers)}}.
  \bibinfo{pages}{345--354}.
\newblock


\bibitem[\protect\citeauthoryear{Devlin, Chang, Lee, and Toutanova}{Devlin
  et~al\mbox{.}}{2019}]%
        {devlin2019bert}
\bibfield{author}{\bibinfo{person}{Jacob Devlin}, \bibinfo{person}{Ming-Wei
  Chang}, \bibinfo{person}{Kenton Lee}, {and} \bibinfo{person}{Kristina
  Toutanova}.} \bibinfo{year}{2019}\natexlab{}.
\newblock \showarticletitle{BERT: Pre-training of Deep Bidirectional
  Transformers for Language Understanding}. In
  \bibinfo{booktitle}{\emph{Proceedings of the 2019 Conference of the North
  American Chapter of the Association for Computational Linguistics: Human
  Language Technologies, Volume 1 (Long and Short Papers)}}.
  \bibinfo{pages}{4171--4186}.
\newblock


\bibitem[\protect\citeauthoryear{Finkelstein, Gabrilovich, Matias, Rivlin,
  Solan, Wolfman, and Ruppin}{Finkelstein et~al\mbox{.}}{2001}]%
        {finkelstein2001placing}
\bibfield{author}{\bibinfo{person}{Lev Finkelstein}, \bibinfo{person}{Evgeniy
  Gabrilovich}, \bibinfo{person}{Yossi Matias}, \bibinfo{person}{Ehud Rivlin},
  \bibinfo{person}{Zach Solan}, \bibinfo{person}{Gadi Wolfman}, {and}
  \bibinfo{person}{Eytan Ruppin}.} \bibinfo{year}{2001}\natexlab{}.
\newblock \showarticletitle{Placing search in context: The concept revisited}.
  In \bibinfo{booktitle}{\emph{Proceedings of the 10th international conference
  on World Wide Web}}. \bibinfo{pages}{406--414}.
\newblock


\bibitem[\protect\citeauthoryear{Gabrilovich, Markovitch,
  et~al\mbox{.}}{Gabrilovich et~al\mbox{.}}{2007}]%
        {gabrilovich2007computing}
\bibfield{author}{\bibinfo{person}{Evgeniy Gabrilovich}, \bibinfo{person}{Shaul
  Markovitch}, {et~al\mbox{.}}} \bibinfo{year}{2007}\natexlab{}.
\newblock \showarticletitle{Computing semantic relatedness using
  wikipedia-based explicit semantic analysis.}. In
  \bibinfo{booktitle}{\emph{IJcAI}}, Vol.~\bibinfo{volume}{7}.
  \bibinfo{pages}{1606--1611}.
\newblock


\bibitem[\protect\citeauthoryear{Ganitkevitch, Van~Durme, and
  Callison-Burch}{Ganitkevitch et~al\mbox{.}}{2013}]%
        {ganitkevitch2013ppdb}
\bibfield{author}{\bibinfo{person}{Juri Ganitkevitch},
  \bibinfo{person}{Benjamin Van~Durme}, {and} \bibinfo{person}{Chris
  Callison-Burch}.} \bibinfo{year}{2013}\natexlab{}.
\newblock \showarticletitle{PPDB: The paraphrase database}. In
  \bibinfo{booktitle}{\emph{Proceedings of the 2013 Conference of the North
  American Chapter of the Association for Computational Linguistics: Human
  Language Technologies}}. \bibinfo{pages}{758--764}.
\newblock


\bibitem[\protect\citeauthoryear{Gao, Zhang, and Chen}{Gao
  et~al\mbox{.}}{2015}]%
        {GAO201580}
\bibfield{author}{\bibinfo{person}{Jian-Bo Gao}, \bibinfo{person}{Bao-Wen
  Zhang}, {and} \bibinfo{person}{Xiao-Hua Chen}.}
  \bibinfo{year}{2015}\natexlab{}.
\newblock \showarticletitle{A WordNet-based semantic similarity measurement
  combining edge-counting and information content theory}.
\newblock \bibinfo{journal}{\emph{Engineering Applications of Artificial
  Intelligence}}  \bibinfo{volume}{39} (\bibinfo{year}{2015}),
  \bibinfo{pages}{80 -- 88}.
\newblock
\showISSN{0952-1976}
\urldef\tempurl%
\url{https://doi.org/10.1016/j.engappai.2014.11.009}
\showDOI{\tempurl}


\bibitem[\protect\citeauthoryear{Gerz, Vuli{\'c}, Hill, Reichart, and
  Korhonen}{Gerz et~al\mbox{.}}{2016}]%
        {gerz2016simverb}
\bibfield{author}{\bibinfo{person}{Daniela Gerz}, \bibinfo{person}{Ivan
  Vuli{\'c}}, \bibinfo{person}{Felix Hill}, \bibinfo{person}{Roi Reichart},
  {and} \bibinfo{person}{Anna Korhonen}.} \bibinfo{year}{2016}\natexlab{}.
\newblock \showarticletitle{SimVerb-3500: A Large-Scale Evaluation Set of Verb
  Similarity}. In \bibinfo{booktitle}{\emph{Proceedings of the 2016 Conference
  on Empirical Methods in Natural Language Processing}}.
  \bibinfo{pages}{2173--2182}.
\newblock


\bibitem[\protect\citeauthoryear{Glavaš, Franco-Salvador, Ponzetto, and
  Rosso}{Glavaš et~al\mbox{.}}{2018}]%
        {GLAVAS20181}
\bibfield{author}{\bibinfo{person}{Goran Glavaš}, \bibinfo{person}{Marc
  Franco-Salvador}, \bibinfo{person}{Simone~P. Ponzetto}, {and}
  \bibinfo{person}{Paolo Rosso}.} \bibinfo{year}{2018}\natexlab{}.
\newblock \showarticletitle{A resource-light method for cross-lingual semantic
  textual similarity}.
\newblock \bibinfo{journal}{\emph{Knowledge-Based Systems}}
  \bibinfo{volume}{143} (\bibinfo{year}{2018}), \bibinfo{pages}{1 -- 9}.
\newblock
\showISSN{0950-7051}
\urldef\tempurl%
\url{https://doi.org/10.1016/j.knosys.2017.11.041}
\showDOI{\tempurl}


\bibitem[\protect\citeauthoryear{Gorman and Curran}{Gorman and Curran}{2006}]%
        {gorman2006scaling}
\bibfield{author}{\bibinfo{person}{James Gorman} {and} \bibinfo{person}{James~R
  Curran}.} \bibinfo{year}{2006}\natexlab{}.
\newblock \showarticletitle{Scaling distributional similarity to large
  corpora}. In \bibinfo{booktitle}{\emph{Proceedings of the 21st International
  Conference on Computational Linguistics and 44th Annual Meeting of the
  Association for Computational Linguistics}}. \bibinfo{pages}{361--368}.
\newblock


\bibitem[\protect\citeauthoryear{Hadj~Taieb, Zesch, and Ben~Aouicha}{Hadj~Taieb
  et~al\mbox{.}}{2019}]%
        {HadjTaieb2019}
\bibfield{author}{\bibinfo{person}{Mohamed~Ali Hadj~Taieb},
  \bibinfo{person}{Torsten Zesch}, {and} \bibinfo{person}{Mohamed
  Ben~Aouicha}.} \bibinfo{year}{2019}\natexlab{}.
\newblock \showarticletitle{A survey of semantic relatedness evaluation
  datasets and procedures}.
\newblock \bibinfo{journal}{\emph{Artificial Intelligence Review}}
  (\bibinfo{date}{23 Dec} \bibinfo{year}{2019}).
\newblock
\showISSN{1573-7462}
\urldef\tempurl%
\url{https://doi.org/10.1007/s10462-019-09796-3}
\showDOI{\tempurl}


\bibitem[\protect\citeauthoryear{Hassan, Abdelrahman, Bahgat, and Farag}{Hassan
  et~al\mbox{.}}{2019}]%
        {hassan2019uests}
\bibfield{author}{\bibinfo{person}{Basma Hassan}, \bibinfo{person}{Samir~E
  Abdelrahman}, \bibinfo{person}{Reem Bahgat}, {and} \bibinfo{person}{Ibrahim
  Farag}.} \bibinfo{year}{2019}\natexlab{}.
\newblock \showarticletitle{UESTS: An Unsupervised Ensemble Semantic Textual
  Similarity Method}.
\newblock \bibinfo{journal}{\emph{IEEE Access}}  \bibinfo{volume}{7}
  (\bibinfo{year}{2019}), \bibinfo{pages}{85462--85482}.
\newblock


\bibitem[\protect\citeauthoryear{He and Lin}{He and Lin}{2016}]%
        {he-lin-2016-pairwise}
\bibfield{author}{\bibinfo{person}{Hua He} {and} \bibinfo{person}{Jimmy Lin}.}
  \bibinfo{year}{2016}\natexlab{}.
\newblock \showarticletitle{Pairwise Word Interaction Modeling with Deep Neural
  Networks for Semantic Similarity Measurement}. In
  \bibinfo{booktitle}{\emph{Proceedings of the 2016 Conference of the North
  {A}merican Chapter of the Association for Computational Linguistics: Human
  Language Technologies}}. \bibinfo{publisher}{Association for Computational
  Linguistics}, \bibinfo{address}{San Diego, California},
  \bibinfo{pages}{937--948}.
\newblock
\urldef\tempurl%
\url{https://doi.org/10.18653/v1/N16-1108}
\showDOI{\tempurl}


\bibitem[\protect\citeauthoryear{Hill, Reichart, and Korhonen}{Hill
  et~al\mbox{.}}{2015}]%
        {hill2015simlex}
\bibfield{author}{\bibinfo{person}{Felix Hill}, \bibinfo{person}{Roi Reichart},
  {and} \bibinfo{person}{Anna Korhonen}.} \bibinfo{year}{2015}\natexlab{}.
\newblock \showarticletitle{Simlex-999: Evaluating semantic models with
  (genuine) similarity estimation}.
\newblock \bibinfo{journal}{\emph{Computational Linguistics}}
  \bibinfo{volume}{41}, \bibinfo{number}{4} (\bibinfo{year}{2015}),
  \bibinfo{pages}{665--695}.
\newblock


\bibitem[\protect\citeauthoryear{Hoffart, Suchanek, Berberich, and
  Weikum}{Hoffart et~al\mbox{.}}{2013}]%
        {hoffart2013yago2}
\bibfield{author}{\bibinfo{person}{Johannes Hoffart}, \bibinfo{person}{Fabian~M
  Suchanek}, \bibinfo{person}{Klaus Berberich}, {and} \bibinfo{person}{Gerhard
  Weikum}.} \bibinfo{year}{2013}\natexlab{}.
\newblock \showarticletitle{YAGO2: A spatially and temporally enhanced
  knowledge base from Wikipedia}.
\newblock \bibinfo{journal}{\emph{Artificial Intelligence}}
  \bibinfo{volume}{194} (\bibinfo{year}{2013}), \bibinfo{pages}{28--61}.
\newblock


\bibitem[\protect\citeauthoryear{Janda, Pawar, Du, and Mago}{Janda
  et~al\mbox{.}}{2019}]%
        {janda2019syntactic}
\bibfield{author}{\bibinfo{person}{Harneet~Kaur Janda}, \bibinfo{person}{Atish
  Pawar}, \bibinfo{person}{Shan Du}, {and} \bibinfo{person}{Vijay Mago}.}
  \bibinfo{year}{2019}\natexlab{}.
\newblock \showarticletitle{Syntactic, Semantic and Sentiment Analysis: The
  Joint Effect on Automated Essay Evaluation}.
\newblock \bibinfo{journal}{\emph{IEEE Access}}  \bibinfo{volume}{7}
  (\bibinfo{year}{2019}), \bibinfo{pages}{108486--108503}.
\newblock


\bibitem[\protect\citeauthoryear{Jiang and Conrath}{Jiang and Conrath}{1997}]%
        {jiang1997semantic}
\bibfield{author}{\bibinfo{person}{Jay~J Jiang} {and} \bibinfo{person}{David~W
  Conrath}.} \bibinfo{year}{1997}\natexlab{}.
\newblock \showarticletitle{Semantic Similarity Based on Corpus Statistics and
  Lexical Taxonomy}. In \bibinfo{booktitle}{\emph{Proceedings of the 10th
  Research on Computational Linguistics International Conference}}.
  \bibinfo{pages}{19--33}.
\newblock


\bibitem[\protect\citeauthoryear{Jiang, Bai, Zhang, and Hu}{Jiang
  et~al\mbox{.}}{2017}]%
        {JIANG2017248}
\bibfield{author}{\bibinfo{person}{Yuncheng Jiang}, \bibinfo{person}{Wen Bai},
  \bibinfo{person}{Xiaopei Zhang}, {and} \bibinfo{person}{Jiaojiao Hu}.}
  \bibinfo{year}{2017}\natexlab{}.
\newblock \showarticletitle{Wikipedia-based information content and semantic
  similarity computation}.
\newblock \bibinfo{journal}{\emph{Information Processing \& Management}}
  \bibinfo{volume}{53}, \bibinfo{number}{1} (\bibinfo{year}{2017}),
  \bibinfo{pages}{248 -- 265}.
\newblock
\showISSN{0306-4573}
\urldef\tempurl%
\url{https://doi.org/10.1016/j.ipm.2016.09.001}
\showDOI{\tempurl}


\bibitem[\protect\citeauthoryear{Jiang, Zhang, Tang, and Nie}{Jiang
  et~al\mbox{.}}{2015}]%
        {jiang2015feature}
\bibfield{author}{\bibinfo{person}{Yuncheng Jiang}, \bibinfo{person}{Xiaopei
  Zhang}, \bibinfo{person}{Yong Tang}, {and} \bibinfo{person}{Ruihua Nie}.}
  \bibinfo{year}{2015}\natexlab{}.
\newblock \showarticletitle{Feature-based approaches to semantic similarity
  assessment of concepts using Wikipedia}.
\newblock \bibinfo{journal}{\emph{Information Processing \& Management}}
  \bibinfo{volume}{51}, \bibinfo{number}{3} (\bibinfo{year}{2015}),
  \bibinfo{pages}{215--234}.
\newblock


\bibitem[\protect\citeauthoryear{Jiao, Yin, Shang, Jiang, Chen, Li, Wang, and
  Liu}{Jiao et~al\mbox{.}}{2019}]%
        {jiao2019tinybert}
\bibfield{author}{\bibinfo{person}{Xiaoqi Jiao}, \bibinfo{person}{Yichun Yin},
  \bibinfo{person}{Lifeng Shang}, \bibinfo{person}{Xin Jiang},
  \bibinfo{person}{Xiao Chen}, \bibinfo{person}{Linlin Li},
  \bibinfo{person}{Fang Wang}, {and} \bibinfo{person}{Qun Liu}.}
  \bibinfo{year}{2019}\natexlab{}.
\newblock \showarticletitle{Tinybert: Distilling bert for natural language
  understanding}.
\newblock \bibinfo{journal}{\emph{arXiv preprint arXiv:1909.10351}}
  (\bibinfo{year}{2019}).
\newblock


\bibitem[\protect\citeauthoryear{Kajiwara and Komachi}{Kajiwara and
  Komachi}{2016}]%
        {kajiwara2016building}
\bibfield{author}{\bibinfo{person}{Tomoyuki Kajiwara} {and}
  \bibinfo{person}{Mamoru Komachi}.} \bibinfo{year}{2016}\natexlab{}.
\newblock \showarticletitle{Building a monolingual parallel corpus for text
  simplification using sentence similarity based on alignment between word
  embeddings}. In \bibinfo{booktitle}{\emph{Proceedings of COLING 2016, the
  26th International Conference on Computational Linguistics: Technical
  Papers}}. \bibinfo{pages}{1147--1158}.
\newblock


\bibitem[\protect\citeauthoryear{Kim, Fiorini, Wilbur, and Lu}{Kim
  et~al\mbox{.}}{2017}]%
        {kim2017bridging}
\bibfield{author}{\bibinfo{person}{Sun Kim}, \bibinfo{person}{Nicolas Fiorini},
  \bibinfo{person}{W~John Wilbur}, {and} \bibinfo{person}{Zhiyong Lu}.}
  \bibinfo{year}{2017}\natexlab{}.
\newblock \showarticletitle{Bridging the gap: Incorporating a semantic
  similarity measure for effectively mapping PubMed queries to documents}.
\newblock \bibinfo{journal}{\emph{Journal of biomedical informatics}}
  \bibinfo{volume}{75} (\bibinfo{year}{2017}), \bibinfo{pages}{122--127}.
\newblock


\bibitem[\protect\citeauthoryear{Kim}{Kim}{2014}]%
        {kim2014convolutional}
\bibfield{author}{\bibinfo{person}{Yoon Kim}.} \bibinfo{year}{2014}\natexlab{}.
\newblock \showarticletitle{Convolutional Neural Networks for Sentence
  Classification}. In \bibinfo{booktitle}{\emph{Proceedings of the 2014
  Conference on Empirical Methods in Natural Language Processing (EMNLP)}}.
  \bibinfo{pages}{1746--1751}.
\newblock


\bibitem[\protect\citeauthoryear{Lan, Chen, Goodman, Gimpel, Sharma, and
  Soricut}{Lan et~al\mbox{.}}{2019}]%
        {lan2019albert}
\bibfield{author}{\bibinfo{person}{Zhenzhong Lan}, \bibinfo{person}{Mingda
  Chen}, \bibinfo{person}{Sebastian Goodman}, \bibinfo{person}{Kevin Gimpel},
  \bibinfo{person}{Piyush Sharma}, {and} \bibinfo{person}{Radu Soricut}.}
  \bibinfo{year}{2019}\natexlab{}.
\newblock \showarticletitle{ALBERT: A Lite BERT for Self-supervised Learning of
  Language Representations}. In \bibinfo{booktitle}{\emph{International
  Conference on Learning Representations}}.
\newblock


\bibitem[\protect\citeauthoryear{Landauer and Dumais}{Landauer and
  Dumais}{1997}]%
        {landauer1997solution}
\bibfield{author}{\bibinfo{person}{Thomas~K Landauer} {and}
  \bibinfo{person}{Susan~T Dumais}.} \bibinfo{year}{1997}\natexlab{}.
\newblock \showarticletitle{A solution to Plato's problem: The latent semantic
  analysis theory of acquisition, induction, and representation of knowledge.}
\newblock \bibinfo{journal}{\emph{Psychological review}} \bibinfo{volume}{104},
  \bibinfo{number}{2} (\bibinfo{year}{1997}), \bibinfo{pages}{211}.
\newblock


\bibitem[\protect\citeauthoryear{Landauer, Foltz, and Laham}{Landauer
  et~al\mbox{.}}{1998}]%
        {landauer1998introduction}
\bibfield{author}{\bibinfo{person}{Thomas~K Landauer}, \bibinfo{person}{Peter~W
  Foltz}, {and} \bibinfo{person}{Darrell Laham}.}
  \bibinfo{year}{1998}\natexlab{}.
\newblock \showarticletitle{An introduction to latent semantic analysis}.
\newblock \bibinfo{journal}{\emph{Discourse processes}} \bibinfo{volume}{25},
  \bibinfo{number}{2-3} (\bibinfo{year}{1998}), \bibinfo{pages}{259--284}.
\newblock


\bibitem[\protect\citeauthoryear{Lastra-D{\'\i}az and
  Garc{\'\i}a-Serrano}{Lastra-D{\'\i}az and Garc{\'\i}a-Serrano}{2015}]%
        {lastra2015new}
\bibfield{author}{\bibinfo{person}{Juan~J Lastra-D{\'\i}az} {and}
  \bibinfo{person}{Ana Garc{\'\i}a-Serrano}.} \bibinfo{year}{2015}\natexlab{}.
\newblock \showarticletitle{A new family of information content models with an
  experimental survey on WordNet}.
\newblock \bibinfo{journal}{\emph{Knowledge-Based Systems}}
  \bibinfo{volume}{89} (\bibinfo{year}{2015}), \bibinfo{pages}{509--526}.
\newblock


\bibitem[\protect\citeauthoryear{Lastra-D{\'\i}az, Garc{\'\i}a-Serrano, Batet,
  Fern{\'a}ndez, and Chirigati}{Lastra-D{\'\i}az et~al\mbox{.}}{2017}]%
        {lastra2017hesml}
\bibfield{author}{\bibinfo{person}{Juan~J Lastra-D{\'\i}az},
  \bibinfo{person}{Ana Garc{\'\i}a-Serrano}, \bibinfo{person}{Montserrat
  Batet}, \bibinfo{person}{Miriam Fern{\'a}ndez}, {and}
  \bibinfo{person}{Fernando Chirigati}.} \bibinfo{year}{2017}\natexlab{}.
\newblock \showarticletitle{HESML: A scalable ontology-based semantic
  similarity measures library with a set of reproducible experiments and a
  replication dataset}.
\newblock \bibinfo{journal}{\emph{Information Systems}}  \bibinfo{volume}{66}
  (\bibinfo{year}{2017}), \bibinfo{pages}{97--118}.
\newblock


\bibitem[\protect\citeauthoryear{Lastra-Díaz, Goikoetxea, Taieb,
  García-Serrano, Aouicha, and Agirre}{Lastra-Díaz et~al\mbox{.}}{2019}]%
        {LASTRADIAZ2019645}
\bibfield{author}{\bibinfo{person}{Juan~J. Lastra-Díaz}, \bibinfo{person}{Josu
  Goikoetxea}, \bibinfo{person}{Mohamed Ali~Hadj Taieb}, \bibinfo{person}{Ana
  García-Serrano}, \bibinfo{person}{Mohamed~Ben Aouicha}, {and}
  \bibinfo{person}{Eneko Agirre}.} \bibinfo{year}{2019}\natexlab{}.
\newblock \showarticletitle{A reproducible survey on word embeddings and
  ontology-based methods for word similarity: Linear combinations outperform
  the state of the art}.
\newblock \bibinfo{journal}{\emph{Engineering Applications of Artificial
  Intelligence}}  \bibinfo{volume}{85} (\bibinfo{year}{2019}),
  \bibinfo{pages}{645 -- 665}.
\newblock
\showISSN{0952-1976}
\urldef\tempurl%
\url{https://doi.org/10.1016/j.engappai.2019.07.010}
\showDOI{\tempurl}


\bibitem[\protect\citeauthoryear{Le and Mikolov}{Le and Mikolov}{2014}]%
        {para2vec}
\bibfield{author}{\bibinfo{person}{Quoc Le} {and} \bibinfo{person}{Tomas
  Mikolov}.} \bibinfo{year}{2014}\natexlab{}.
\newblock \showarticletitle{Distributed representations of sentences and
  documents}. In \bibinfo{booktitle}{\emph{International conference on machine
  learning}}. \bibinfo{pages}{1188--1196}.
\newblock


\bibitem[\protect\citeauthoryear{Le, Wang, Quan, He, and Yao}{Le
  et~al\mbox{.}}{2018}]%
        {le2018acv}
\bibfield{author}{\bibinfo{person}{Yuquan Le}, \bibinfo{person}{Zhi-Jie Wang},
  \bibinfo{person}{Zhe Quan}, \bibinfo{person}{Jiawei He}, {and}
  \bibinfo{person}{Bin Yao}.} \bibinfo{year}{2018}\natexlab{}.
\newblock \showarticletitle{ACV-tree: A New Method for Sentence Similarity
  Modeling.}. In \bibinfo{booktitle}{\emph{IJCAI}}.
  \bibinfo{pages}{4137--4143}.
\newblock


\bibitem[\protect\citeauthoryear{Lee}{Lee}{2011}]%
        {lee2011novel}
\bibfield{author}{\bibinfo{person}{Ming~Che Lee}.}
  \bibinfo{year}{2011}\natexlab{}.
\newblock \showarticletitle{A novel sentence similarity measure for
  semantic-based expert systems}.
\newblock \bibinfo{journal}{\emph{Expert Systems with Applications}}
  \bibinfo{volume}{38}, \bibinfo{number}{5} (\bibinfo{year}{2011}),
  \bibinfo{pages}{6392--6399}.
\newblock


\bibitem[\protect\citeauthoryear{Levy and Goldberg}{Levy and Goldberg}{2014a}]%
        {levy2014dependency}
\bibfield{author}{\bibinfo{person}{Omer Levy} {and} \bibinfo{person}{Yoav
  Goldberg}.} \bibinfo{year}{2014}\natexlab{a}.
\newblock \showarticletitle{Dependency-based word embeddings}. In
  \bibinfo{booktitle}{\emph{Proceedings of the 52nd Annual Meeting of the
  Association for Computational Linguistics (Volume 2: Short Papers)}}.
  \bibinfo{pages}{302--308}.
\newblock


\bibitem[\protect\citeauthoryear{Levy and Goldberg}{Levy and Goldberg}{2014b}]%
        {levy2014neural}
\bibfield{author}{\bibinfo{person}{Omer Levy} {and} \bibinfo{person}{Yoav
  Goldberg}.} \bibinfo{year}{2014}\natexlab{b}.
\newblock \showarticletitle{Neural word embedding as implicit matrix
  factorization}. In \bibinfo{booktitle}{\emph{Advances in neural information
  processing systems}}. \bibinfo{pages}{2177--2185}.
\newblock


\bibitem[\protect\citeauthoryear{Li, Wang, Zhu, Wang, and Wu}{Li
  et~al\mbox{.}}{2013}]%
        {li2013computing}
\bibfield{author}{\bibinfo{person}{Peipei Li}, \bibinfo{person}{Haixun Wang},
  \bibinfo{person}{Kenny~Q Zhu}, \bibinfo{person}{Zhongyuan Wang}, {and}
  \bibinfo{person}{Xindong Wu}.} \bibinfo{year}{2013}\natexlab{}.
\newblock \showarticletitle{Computing term similarity by large probabilistic
  isa knowledge}. In \bibinfo{booktitle}{\emph{Proceedings of the 22nd ACM
  international conference on Information \& Knowledge Management}}.
  \bibinfo{pages}{1401--1410}.
\newblock


\bibitem[\protect\citeauthoryear{Li, Bandar, and McLean}{Li
  et~al\mbox{.}}{2003}]%
        {li2003approach}
\bibfield{author}{\bibinfo{person}{Yuhua Li}, \bibinfo{person}{Zuhair~A
  Bandar}, {and} \bibinfo{person}{David McLean}.}
  \bibinfo{year}{2003}\natexlab{}.
\newblock \showarticletitle{An approach for measuring semantic similarity
  between words using multiple information sources}.
\newblock \bibinfo{journal}{\emph{IEEE Transactions on knowledge and data
  engineering}} \bibinfo{volume}{15}, \bibinfo{number}{4}
  (\bibinfo{year}{2003}), \bibinfo{pages}{871--882}.
\newblock


\bibitem[\protect\citeauthoryear{Li, McLean, Bandar, O'shea, and Crockett}{Li
  et~al\mbox{.}}{2006}]%
        {li2006sentence}
\bibfield{author}{\bibinfo{person}{Yuhua Li}, \bibinfo{person}{David McLean},
  \bibinfo{person}{Zuhair~A Bandar}, \bibinfo{person}{James~D O'shea}, {and}
  \bibinfo{person}{Keeley Crockett}.} \bibinfo{year}{2006}\natexlab{}.
\newblock \showarticletitle{Sentence similarity based on semantic nets and
  corpus statistics}.
\newblock \bibinfo{journal}{\emph{IEEE transactions on knowledge and data
  engineering}} \bibinfo{volume}{18}, \bibinfo{number}{8}
  (\bibinfo{year}{2006}), \bibinfo{pages}{1138--1150}.
\newblock


\bibitem[\protect\citeauthoryear{Lin et~al\mbox{.}}{Lin et~al\mbox{.}}{1998}]%
        {lin1998information}
\bibfield{author}{\bibinfo{person}{Dekang Lin} {et~al\mbox{.}}}
  \bibinfo{year}{1998}\natexlab{}.
\newblock \showarticletitle{An information-theoretic definition of
  similarity.}. In \bibinfo{booktitle}{\emph{Icml}}, Vol.~\bibinfo{volume}{98}.
  \bibinfo{pages}{296--304}.
\newblock


\bibitem[\protect\citeauthoryear{Liu, Ott, Goyal, Du, Joshi, Chen, Levy, Lewis,
  Zettlemoyer, and Stoyanov}{Liu et~al\mbox{.}}{2019}]%
        {liu2019roberta}
\bibfield{author}{\bibinfo{person}{Yinhan Liu}, \bibinfo{person}{Myle Ott},
  \bibinfo{person}{Naman Goyal}, \bibinfo{person}{Jingfei Du},
  \bibinfo{person}{Mandar Joshi}, \bibinfo{person}{Danqi Chen},
  \bibinfo{person}{Omer Levy}, \bibinfo{person}{Mike Lewis},
  \bibinfo{person}{Luke Zettlemoyer}, {and} \bibinfo{person}{Veselin
  Stoyanov}.} \bibinfo{year}{2019}\natexlab{}.
\newblock \showarticletitle{Roberta: A robustly optimized bert pretraining
  approach}.
\newblock \bibinfo{journal}{\emph{arXiv preprint arXiv:1907.11692}}
  (\bibinfo{year}{2019}).
\newblock


\bibitem[\protect\citeauthoryear{Lopez-Gazpio, Maritxalar, Gonzalez-Agirre,
  Rigau, Uria, and Agirre}{Lopez-Gazpio et~al\mbox{.}}{2017}]%
        {LOPEZGAZPIO2017186}
\bibfield{author}{\bibinfo{person}{I. Lopez-Gazpio}, \bibinfo{person}{M.
  Maritxalar}, \bibinfo{person}{A. Gonzalez-Agirre}, \bibinfo{person}{G.
  Rigau}, \bibinfo{person}{L. Uria}, {and} \bibinfo{person}{E. Agirre}.}
  \bibinfo{year}{2017}\natexlab{}.
\newblock \showarticletitle{Interpretable semantic textual similarity: Finding
  and explaining differences between sentences}.
\newblock \bibinfo{journal}{\emph{Knowledge-Based Systems}}
  \bibinfo{volume}{119} (\bibinfo{year}{2017}), \bibinfo{pages}{186 -- 199}.
\newblock
\showISSN{0950-7051}
\urldef\tempurl%
\url{https://doi.org/10.1016/j.knosys.2016.12.013}
\showDOI{\tempurl}


\bibitem[\protect\citeauthoryear{Lopez-Gazpio, Maritxalar, Lapata, and
  Agirre}{Lopez-Gazpio et~al\mbox{.}}{2019}]%
        {LOPEZGAZPIO20191}
\bibfield{author}{\bibinfo{person}{I. Lopez-Gazpio}, \bibinfo{person}{M.
  Maritxalar}, \bibinfo{person}{M. Lapata}, {and} \bibinfo{person}{E. Agirre}.}
  \bibinfo{year}{2019}\natexlab{}.
\newblock \showarticletitle{Word n-gram attention models for sentence
  similarity and inference}.
\newblock \bibinfo{journal}{\emph{Expert Systems with Applications}}
  \bibinfo{volume}{132} (\bibinfo{year}{2019}), \bibinfo{pages}{1 -- 11}.
\newblock
\showISSN{0957-4174}
\urldef\tempurl%
\url{https://doi.org/10.1016/j.eswa.2019.04.054}
\showDOI{\tempurl}


\bibitem[\protect\citeauthoryear{Lund and Burgess}{Lund and Burgess}{1996}]%
        {lund1996producing}
\bibfield{author}{\bibinfo{person}{Kevin Lund} {and} \bibinfo{person}{Curt
  Burgess}.} \bibinfo{year}{1996}\natexlab{}.
\newblock \showarticletitle{Producing high-dimensional semantic spaces from
  lexical co-occurrence}.
\newblock \bibinfo{journal}{\emph{Behavior research methods, instruments, \&
  computers}} \bibinfo{volume}{28}, \bibinfo{number}{2} (\bibinfo{year}{1996}),
  \bibinfo{pages}{203--208}.
\newblock


\bibitem[\protect\citeauthoryear{Marelli, Menini, Baroni, Bentivogli, Bernardi,
  and Zamparelli}{Marelli et~al\mbox{.}}{2014}]%
        {marellisick}
\bibfield{author}{\bibinfo{person}{Marco Marelli}, \bibinfo{person}{Stefano
  Menini}, \bibinfo{person}{Marco Baroni}, \bibinfo{person}{Luisa Bentivogli},
  \bibinfo{person}{Raffaella Bernardi}, {and} \bibinfo{person}{Roberto
  Zamparelli}.} \bibinfo{year}{2014}\natexlab{}.
\newblock \showarticletitle{A {SICK} cure for the evaluation of compositional
  distributional semantic models}. In \bibinfo{booktitle}{\emph{Proceedings of
  the Ninth International Conference on Language Resources and Evaluation
  ({LREC}'14)}}. \bibinfo{publisher}{European Language Resources Association
  (ELRA)}, \bibinfo{address}{Reykjavik, Iceland}, \bibinfo{pages}{216--223}.
\newblock
\urldef\tempurl%
\url{http://www.lrec-conf.org/proceedings/lrec2014/pdf/363_Paper.pdf}
\showURL{%
\tempurl}


\bibitem[\protect\citeauthoryear{McCann, Bradbury, Xiong, and Socher}{McCann
  et~al\mbox{.}}{2017}]%
        {mccann2017learned}
\bibfield{author}{\bibinfo{person}{Bryan McCann}, \bibinfo{person}{James
  Bradbury}, \bibinfo{person}{Caiming Xiong}, {and} \bibinfo{person}{Richard
  Socher}.} \bibinfo{year}{2017}\natexlab{}.
\newblock \showarticletitle{Learned in translation: contextualized word
  vectors}. In \bibinfo{booktitle}{\emph{Proceedings of the 31st International
  Conference on Neural Information Processing Systems}}. Curran Associates
  Inc., \bibinfo{pages}{6297--6308}.
\newblock


\bibitem[\protect\citeauthoryear{McInnes, Liu, Pedersen, Melton, and
  Pakhomov}{McInnes et~al\mbox{.}}{[n.d.]}]%
        {mcinnes2013umls}
\bibfield{author}{\bibinfo{person}{Bridget~T McInnes}, \bibinfo{person}{Ying
  Liu}, \bibinfo{person}{Ted Pedersen}, \bibinfo{person}{Genevieve~B Melton},
  {and} \bibinfo{person}{Serguei~V Pakhomov}.}
  \bibinfo{year}{[n.d.]}\natexlab{}.
\newblock \showarticletitle{UMLS:: Similarity: Measuring the Relatedness and
  Similarity of Biomedical Concepts}. In \bibinfo{booktitle}{\emph{Human
  Language Technologies: The 2013 Annual Conference of the North American
  Chapter of the Association for Computational Linguistics}}.
  \bibinfo{pages}{28}.
\newblock


\bibitem[\protect\citeauthoryear{Meek, Yi, and Wen-tau}{Meek
  et~al\mbox{.}}{2018}]%
        {Meek2018}
\bibfield{author}{\bibinfo{person}{Christopher Meek}, \bibinfo{person}{Yang
  Yi}, {and} \bibinfo{person}{Yih Wen-tau}.} \bibinfo{year}{2018}\natexlab{}.
\newblock \showarticletitle{WIKIQA : A Challenge Dataset for Open-Domain
  Question Answering}.
\newblock \bibinfo{journal}{\emph{Proceedings of the 2015 Conference on
  Empirical Methods in Natural Language Processing}} \bibinfo{number}{September
  2015} (\bibinfo{year}{2018}), \bibinfo{pages}{2013--2018}.
\newblock
\urldef\tempurl%
\url{https://doi.org/10.18653/v1/D15-1237}
\showDOI{\tempurl}


\bibitem[\protect\citeauthoryear{Melamud, Goldberger, and Dagan}{Melamud
  et~al\mbox{.}}{2016}]%
        {context2vec}
\bibfield{author}{\bibinfo{person}{Oren Melamud}, \bibinfo{person}{Jacob
  Goldberger}, {and} \bibinfo{person}{Ido Dagan}.}
  \bibinfo{year}{2016}\natexlab{}.
\newblock \showarticletitle{context2vec: Learning generic context embedding
  with bidirectional LSTM}. In \bibinfo{booktitle}{\emph{Proceedings of The
  20th SIGNLL Conference on Computational Natural Language Learning}}.
  \bibinfo{pages}{51--61}.
\newblock


\bibitem[\protect\citeauthoryear{Mihalcea and Csomai}{Mihalcea and
  Csomai}{2007}]%
        {mihalcea2007wikify}
\bibfield{author}{\bibinfo{person}{Rada Mihalcea} {and} \bibinfo{person}{Andras
  Csomai}.} \bibinfo{year}{2007}\natexlab{}.
\newblock \showarticletitle{Wikify! Linking documents to encyclopedic
  knowledge}. In \bibinfo{booktitle}{\emph{Proceedings of the sixteenth ACM
  conference on Conference on information and knowledge management}}.
  \bibinfo{pages}{233--242}.
\newblock


\bibitem[\protect\citeauthoryear{Mikolov, Chen, Corrado, and Dean}{Mikolov
  et~al\mbox{.}}{2013a}]%
        {mikolov2013efficient}
\bibfield{author}{\bibinfo{person}{Tomas Mikolov}, \bibinfo{person}{Kai Chen},
  \bibinfo{person}{Greg Corrado}, {and} \bibinfo{person}{Jeffrey Dean}.}
  \bibinfo{year}{2013}\natexlab{a}.
\newblock \showarticletitle{Efficient estimation of word representations in
  vector space}.
\newblock \bibinfo{journal}{\emph{arXiv preprint arXiv:1301.3781}}
  (\bibinfo{year}{2013}).
\newblock


\bibitem[\protect\citeauthoryear{Mikolov, Yih, and Zweig}{Mikolov
  et~al\mbox{.}}{2013b}]%
        {mikolov2013linguistic}
\bibfield{author}{\bibinfo{person}{Tom{\'a}{\v{s}} Mikolov},
  \bibinfo{person}{Wen-tau Yih}, {and} \bibinfo{person}{Geoffrey Zweig}.}
  \bibinfo{year}{2013}\natexlab{b}.
\newblock \showarticletitle{Linguistic regularities in continuous space word
  representations}. In \bibinfo{booktitle}{\emph{Proceedings of the 2013
  conference of the north american chapter of the association for computational
  linguistics: Human language technologies}}. \bibinfo{pages}{746--751}.
\newblock


\bibitem[\protect\citeauthoryear{Miller}{Miller}{1995}]%
        {miller1995wordnet}
\bibfield{author}{\bibinfo{person}{George~A Miller}.}
  \bibinfo{year}{1995}\natexlab{}.
\newblock \showarticletitle{WordNet: a lexical database for English}.
\newblock \bibinfo{journal}{\emph{Commun. ACM}} \bibinfo{volume}{38},
  \bibinfo{number}{11} (\bibinfo{year}{1995}), \bibinfo{pages}{39--41}.
\newblock


\bibitem[\protect\citeauthoryear{Miller and Charles}{Miller and
  Charles}{1991}]%
        {miller1991contextual}
\bibfield{author}{\bibinfo{person}{George~A Miller} {and}
  \bibinfo{person}{Walter~G Charles}.} \bibinfo{year}{1991}\natexlab{}.
\newblock \showarticletitle{Contextual correlates of semantic similarity}.
\newblock \bibinfo{journal}{\emph{Language and cognitive processes}}
  \bibinfo{volume}{6}, \bibinfo{number}{1} (\bibinfo{year}{1991}),
  \bibinfo{pages}{1--28}.
\newblock


\bibitem[\protect\citeauthoryear{Mnih and Kavukcuoglu}{Mnih and
  Kavukcuoglu}{2013}]%
        {mnih2013learning}
\bibfield{author}{\bibinfo{person}{Andriy Mnih} {and} \bibinfo{person}{Koray
  Kavukcuoglu}.} \bibinfo{year}{2013}\natexlab{}.
\newblock \showarticletitle{Learning word embeddings efficiently with
  noise-contrastive estimation}. In \bibinfo{booktitle}{\emph{Advances in
  neural information processing systems}}. \bibinfo{pages}{2265--2273}.
\newblock


\bibitem[\protect\citeauthoryear{Mohamed and Oussalah}{Mohamed and
  Oussalah}{2019}]%
        {mohamed2019srl}
\bibfield{author}{\bibinfo{person}{Muhidin Mohamed} {and}
  \bibinfo{person}{Mourad Oussalah}.} \bibinfo{year}{2019}\natexlab{}.
\newblock \showarticletitle{SRL-ESA-TextSum: A text summarization approach
  based on semantic role labeling and explicit semantic analysis}.
\newblock \bibinfo{journal}{\emph{Information Processing \& Management}}
  \bibinfo{volume}{56}, \bibinfo{number}{4} (\bibinfo{year}{2019}),
  \bibinfo{pages}{1356--1372}.
\newblock


\bibitem[\protect\citeauthoryear{Mohammad and Hirst}{Mohammad and
  Hirst}{2012}]%
        {mohammad2012distributional}
\bibfield{author}{\bibinfo{person}{Saif~M Mohammad} {and}
  \bibinfo{person}{Graeme Hirst}.} \bibinfo{year}{2012}\natexlab{}.
\newblock \showarticletitle{Distributional measures of semantic distance: A
  survey}.
\newblock \bibinfo{journal}{\emph{arXiv preprint arXiv:1203.1858}}
  (\bibinfo{year}{2012}).
\newblock


\bibitem[\protect\citeauthoryear{Moschitti}{Moschitti}{2006}]%
        {moschitti2006efficient}
\bibfield{author}{\bibinfo{person}{Alessandro Moschitti}.}
  \bibinfo{year}{2006}\natexlab{}.
\newblock \showarticletitle{Efficient convolution kernels for dependency and
  constituent syntactic trees}. In \bibinfo{booktitle}{\emph{European
  Conference on Machine Learning}}. Springer, \bibinfo{pages}{318--329}.
\newblock


\bibitem[\protect\citeauthoryear{Moschitti}{Moschitti}{2008}]%
        {moschitti2008kernel}
\bibfield{author}{\bibinfo{person}{Alessandro Moschitti}.}
  \bibinfo{year}{2008}\natexlab{}.
\newblock \showarticletitle{Kernel methods, syntax and semantics for relational
  text categorization}. In \bibinfo{booktitle}{\emph{Proceedings of the 17th
  ACM conference on Information and knowledge management}}.
  \bibinfo{pages}{253--262}.
\newblock


\bibitem[\protect\citeauthoryear{Moschitti, Pighin, and Basili}{Moschitti
  et~al\mbox{.}}{2008}]%
        {moschitti2008tree}
\bibfield{author}{\bibinfo{person}{Alessandro Moschitti},
  \bibinfo{person}{Daniele Pighin}, {and} \bibinfo{person}{Roberto Basili}.}
  \bibinfo{year}{2008}\natexlab{}.
\newblock \showarticletitle{Tree kernels for semantic role labeling}.
\newblock \bibinfo{journal}{\emph{Computational Linguistics}}
  \bibinfo{volume}{34}, \bibinfo{number}{2} (\bibinfo{year}{2008}),
  \bibinfo{pages}{193--224}.
\newblock


\bibitem[\protect\citeauthoryear{Moschitti and Quarteroni}{Moschitti and
  Quarteroni}{2008}]%
        {moschitti2008kernels}
\bibfield{author}{\bibinfo{person}{Alessandro Moschitti} {and}
  \bibinfo{person}{Silvia Quarteroni}.} \bibinfo{year}{2008}\natexlab{}.
\newblock \showarticletitle{Kernels on linguistic structures for answer
  extraction}. In \bibinfo{booktitle}{\emph{Proceedings of ACL-08: HLT, Short
  Papers}}. \bibinfo{pages}{113--116}.
\newblock


\bibitem[\protect\citeauthoryear{Moschitti, Quarteroni, Basili, and
  Manandhar}{Moschitti et~al\mbox{.}}{2007}]%
        {moschitti2007exploiting}
\bibfield{author}{\bibinfo{person}{Alessandro Moschitti},
  \bibinfo{person}{Silvia Quarteroni}, \bibinfo{person}{Roberto Basili}, {and}
  \bibinfo{person}{Suresh Manandhar}.} \bibinfo{year}{2007}\natexlab{}.
\newblock \showarticletitle{Exploiting syntactic and shallow semantic kernels
  for question answer classification}. In \bibinfo{booktitle}{\emph{Proceedings
  of the 45th annual meeting of the association of computational linguistics}}.
  \bibinfo{pages}{776--783}.
\newblock


\bibitem[\protect\citeauthoryear{Moschitti and Zanzotto}{Moschitti and
  Zanzotto}{2007}]%
        {moschitti2007fast}
\bibfield{author}{\bibinfo{person}{Alessandro Moschitti} {and}
  \bibinfo{person}{Fabio~Massimo Zanzotto}.} \bibinfo{year}{2007}\natexlab{}.
\newblock \showarticletitle{Fast and effective kernels for relational learning
  from texts}. In \bibinfo{booktitle}{\emph{Proceedings of the 24th
  international conference on Machine learning}}. \bibinfo{pages}{649--656}.
\newblock


\bibitem[\protect\citeauthoryear{Navigli and Ponzetto}{Navigli and
  Ponzetto}{2012}]%
        {navigli2012babelnet}
\bibfield{author}{\bibinfo{person}{Roberto Navigli} {and}
  \bibinfo{person}{Simone~Paolo Ponzetto}.} \bibinfo{year}{2012}\natexlab{}.
\newblock \showarticletitle{BabelNet: The automatic construction, evaluation
  and application of a wide-coverage multilingual semantic network}.
\newblock \bibinfo{journal}{\emph{Artificial Intelligence}}
  \bibinfo{volume}{193} (\bibinfo{year}{2012}).
\newblock


\bibitem[\protect\citeauthoryear{Nelson, McEvoy, and Schreiber}{Nelson
  et~al\mbox{.}}{2004}]%
        {nelson2004university}
\bibfield{author}{\bibinfo{person}{Douglas~L Nelson}, \bibinfo{person}{Cathy~L
  McEvoy}, {and} \bibinfo{person}{Thomas~A Schreiber}.}
  \bibinfo{year}{2004}\natexlab{}.
\newblock \showarticletitle{The University of South Florida free association,
  rhyme, and word fragment norms}.
\newblock \bibinfo{journal}{\emph{Behavior Research Methods, Instruments, \&
  Computers}} \bibinfo{volume}{36}, \bibinfo{number}{3} (\bibinfo{year}{2004}),
  \bibinfo{pages}{402--407}.
\newblock


\bibitem[\protect\citeauthoryear{Nivre}{Nivre}{2006}]%
        {nivre2006inductive}
\bibfield{author}{\bibinfo{person}{Joakim Nivre}.}
  \bibinfo{year}{2006}\natexlab{}.
\newblock \bibinfo{booktitle}{\emph{Inductive dependency parsing}}.
\newblock \bibinfo{publisher}{Springer}.
\newblock


\bibitem[\protect\citeauthoryear{Pagliardini, Gupta, and Jaggi}{Pagliardini
  et~al\mbox{.}}{2018}]%
        {pagliardini2018unsupervised}
\bibfield{author}{\bibinfo{person}{Matteo Pagliardini},
  \bibinfo{person}{Prakhar Gupta}, {and} \bibinfo{person}{Martin Jaggi}.}
  \bibinfo{year}{2018}\natexlab{}.
\newblock \showarticletitle{Unsupervised Learning of Sentence Embeddings Using
  Compositional n-Gram Features}. In \bibinfo{booktitle}{\emph{Proceedings of
  the 2018 Conference of the North American Chapter of the Association for
  Computational Linguistics: Human Language Technologies, Volume 1 (Long
  Papers)}}. \bibinfo{pages}{528--540}.
\newblock


\bibitem[\protect\citeauthoryear{Parikh, T{\"a}ckstr{\"o}m, Das, and
  Uszkoreit}{Parikh et~al\mbox{.}}{2016}]%
        {parikh2016decomposable}
\bibfield{author}{\bibinfo{person}{Ankur Parikh}, \bibinfo{person}{Oscar
  T{\"a}ckstr{\"o}m}, \bibinfo{person}{Dipanjan Das}, {and}
  \bibinfo{person}{Jakob Uszkoreit}.} \bibinfo{year}{2016}\natexlab{}.
\newblock \showarticletitle{A Decomposable Attention Model for Natural Language
  Inference}. In \bibinfo{booktitle}{\emph{Proceedings of the 2016 Conference
  on Empirical Methods in Natural Language Processing}}.
  \bibinfo{pages}{2249--2255}.
\newblock


\bibitem[\protect\citeauthoryear{{Pawar} and {Mago}}{{Pawar} and
  {Mago}}{2019}]%
        {8630924}
\bibfield{author}{\bibinfo{person}{A. {Pawar}} {and} \bibinfo{person}{V.
  {Mago}}.} \bibinfo{year}{2019}\natexlab{}.
\newblock \showarticletitle{Challenging the Boundaries of Unsupervised Learning
  for Semantic Similarity}.
\newblock \bibinfo{journal}{\emph{IEEE Access}}  \bibinfo{volume}{7}
  (\bibinfo{year}{2019}), \bibinfo{pages}{16291--16308}.
\newblock
\showISSN{2169-3536}


\bibitem[\protect\citeauthoryear{Pedersen, Pakhomov, Patwardhan, and
  Chute}{Pedersen et~al\mbox{.}}{2007}]%
        {pedersen2007measures}
\bibfield{author}{\bibinfo{person}{Ted Pedersen}, \bibinfo{person}{Serguei~VS
  Pakhomov}, \bibinfo{person}{Siddharth Patwardhan}, {and}
  \bibinfo{person}{Christopher~G Chute}.} \bibinfo{year}{2007}\natexlab{}.
\newblock \showarticletitle{Measures of semantic similarity and relatedness in
  the biomedical domain}.
\newblock \bibinfo{journal}{\emph{Journal of biomedical informatics}}
  \bibinfo{volume}{40}, \bibinfo{number}{3} (\bibinfo{year}{2007}),
  \bibinfo{pages}{288--299}.
\newblock


\bibitem[\protect\citeauthoryear{Pennington, Socher, and Manning}{Pennington
  et~al\mbox{.}}{2014}]%
        {pennington2014glove}
\bibfield{author}{\bibinfo{person}{Jeffrey Pennington},
  \bibinfo{person}{Richard Socher}, {and} \bibinfo{person}{Christopher~D
  Manning}.} \bibinfo{year}{2014}\natexlab{}.
\newblock \showarticletitle{Glove: Global vectors for word representation}. In
  \bibinfo{booktitle}{\emph{Proceedings of the 2014 conference on empirical
  methods in natural language processing (EMNLP)}}.
  \bibinfo{pages}{1532--1543}.
\newblock


\bibitem[\protect\citeauthoryear{Peters, Neumann, Iyyer, Gardner, Clark, Lee,
  and Zettlemoyer}{Peters et~al\mbox{.}}{2018}]%
        {peters2018deep}
\bibfield{author}{\bibinfo{person}{Matthew~E Peters}, \bibinfo{person}{Mark
  Neumann}, \bibinfo{person}{Mohit Iyyer}, \bibinfo{person}{Matt Gardner},
  \bibinfo{person}{Christopher Clark}, \bibinfo{person}{Kenton Lee}, {and}
  \bibinfo{person}{Luke Zettlemoyer}.} \bibinfo{year}{2018}\natexlab{}.
\newblock \showarticletitle{Deep contextualized word representations}. In
  \bibinfo{booktitle}{\emph{Proceedings of NAACL-HLT}}.
  \bibinfo{pages}{2227--2237}.
\newblock


\bibitem[\protect\citeauthoryear{Pilehvar and Camacho-Collados}{Pilehvar and
  Camacho-Collados}{2019}]%
        {pilehvar2019wic}
\bibfield{author}{\bibinfo{person}{Mohammad~Taher Pilehvar} {and}
  \bibinfo{person}{Jose Camacho-Collados}.} \bibinfo{year}{2019}\natexlab{}.
\newblock \showarticletitle{WiC: the Word-in-Context Dataset for Evaluating
  Context-Sensitive Meaning Representations}. In
  \bibinfo{booktitle}{\emph{Proceedings of the 2019 Conference of the North
  American Chapter of the Association for Computational Linguistics: Human
  Language Technologies, Volume 1 (Long and Short Papers)}}.
  \bibinfo{pages}{1267--1273}.
\newblock


\bibitem[\protect\citeauthoryear{Pilehvar, Jurgens, and Navigli}{Pilehvar
  et~al\mbox{.}}{2013}]%
        {pilehvar2013align}
\bibfield{author}{\bibinfo{person}{Mohammad~Taher Pilehvar},
  \bibinfo{person}{David Jurgens}, {and} \bibinfo{person}{Roberto Navigli}.}
  \bibinfo{year}{2013}\natexlab{}.
\newblock \showarticletitle{Align, disambiguate and walk: A unified approach
  for measuring semantic similarity}. In \bibinfo{booktitle}{\emph{Proceedings
  of the 51st Annual Meeting of the Association for Computational Linguistics
  (Volume 1: Long Papers)}}. \bibinfo{pages}{1341--1351}.
\newblock


\bibitem[\protect\citeauthoryear{Pilehvar and Navigli}{Pilehvar and
  Navigli}{2015}]%
        {PILEHVAR201595}
\bibfield{author}{\bibinfo{person}{Mohammad~Taher Pilehvar} {and}
  \bibinfo{person}{Roberto Navigli}.} \bibinfo{year}{2015}\natexlab{}.
\newblock \showarticletitle{From senses to texts: An all-in-one graph-based
  approach for measuring semantic similarity}.
\newblock \bibinfo{journal}{\emph{Artificial Intelligence}}
  \bibinfo{volume}{228} (\bibinfo{year}{2015}), \bibinfo{pages}{95 -- 128}.
\newblock
\showISSN{0004-3702}
\urldef\tempurl%
\url{https://doi.org/10.1016/j.artint.2015.07.005}
\showDOI{\tempurl}


\bibitem[\protect\citeauthoryear{Qu, Fang, Bai, and Jiang}{Qu
  et~al\mbox{.}}{2018}]%
        {QU20181002}
\bibfield{author}{\bibinfo{person}{Rong Qu}, \bibinfo{person}{Yongyi Fang},
  \bibinfo{person}{Wen Bai}, {and} \bibinfo{person}{Yuncheng Jiang}.}
  \bibinfo{year}{2018}\natexlab{}.
\newblock \showarticletitle{Computing semantic similarity based on novel models
  of semantic representation using Wikipedia}.
\newblock \bibinfo{journal}{\emph{Information Processing \& Management}}
  \bibinfo{volume}{54}, \bibinfo{number}{6} (\bibinfo{year}{2018}),
  \bibinfo{pages}{1002 -- 1021}.
\newblock
\showISSN{0306-4573}
\urldef\tempurl%
\url{https://doi.org/10.1016/j.ipm.2018.07.002}
\showDOI{\tempurl}


\bibitem[\protect\citeauthoryear{{Quan}, {Wang}, {Le}, {Yao}, {Li}, and
  {Yin}}{{Quan} et~al\mbox{.}}{2019}]%
        {8642425}
\bibfield{author}{\bibinfo{person}{Z. {Quan}}, \bibinfo{person}{Z. {Wang}},
  \bibinfo{person}{Y. {Le}}, \bibinfo{person}{B. {Yao}}, \bibinfo{person}{K.
  {Li}}, {and} \bibinfo{person}{J. {Yin}}.} \bibinfo{year}{2019}\natexlab{}.
\newblock \showarticletitle{An Efficient Framework for Sentence Similarity
  Modeling}.
\newblock \bibinfo{journal}{\emph{IEEE/ACM Transactions on Audio, Speech, and
  Language Processing}} \bibinfo{volume}{27}, \bibinfo{number}{4}
  (\bibinfo{date}{April} \bibinfo{year}{2019}), \bibinfo{pages}{853--865}.
\newblock
\showISSN{2329-9304}
\urldef\tempurl%
\url{https://doi.org/10.1109/TASLP.2019.2899494}
\showDOI{\tempurl}


\bibitem[\protect\citeauthoryear{Rada, Mili, Bicknell, and Blettner}{Rada
  et~al\mbox{.}}{1989}]%
        {rada1989development}
\bibfield{author}{\bibinfo{person}{Roy Rada}, \bibinfo{person}{Hafedh Mili},
  \bibinfo{person}{Ellen Bicknell}, {and} \bibinfo{person}{Maria Blettner}.}
  \bibinfo{year}{1989}\natexlab{}.
\newblock \showarticletitle{Development and application of a metric on semantic
  nets}.
\newblock \bibinfo{journal}{\emph{IEEE transactions on systems, man, and
  cybernetics}} \bibinfo{volume}{19}, \bibinfo{number}{1}
  (\bibinfo{year}{1989}), \bibinfo{pages}{17--30}.
\newblock


\bibitem[\protect\citeauthoryear{Raffel, Shazeer, Roberts, Lee, Narang, Matena,
  Zhou, Li, and Liu}{Raffel et~al\mbox{.}}{2019}]%
        {raffel2019exploring}
\bibfield{author}{\bibinfo{person}{Colin Raffel}, \bibinfo{person}{Noam
  Shazeer}, \bibinfo{person}{Adam Roberts}, \bibinfo{person}{Katherine Lee},
  \bibinfo{person}{Sharan Narang}, \bibinfo{person}{Michael Matena},
  \bibinfo{person}{Yanqi Zhou}, \bibinfo{person}{Wei Li}, {and}
  \bibinfo{person}{Peter~J Liu}.} \bibinfo{year}{2019}\natexlab{}.
\newblock \showarticletitle{Exploring the limits of transfer learning with a
  unified text-to-text transformer}.
\newblock \bibinfo{journal}{\emph{arXiv preprint arXiv:1910.10683}}
  (\bibinfo{year}{2019}).
\newblock


\bibitem[\protect\citeauthoryear{Resnik}{Resnik}{1995}]%
        {resnik1995using}
\bibfield{author}{\bibinfo{person}{Philip Resnik}.}
  \bibinfo{year}{1995}\natexlab{}.
\newblock \showarticletitle{Using information content to evaluate semantic
  similarity in a taxonomy}. In \bibinfo{booktitle}{\emph{Proceedings of the
  14th international joint conference on Artificial intelligence-Volume 1}}.
  \bibinfo{pages}{448--453}.
\newblock


\bibitem[\protect\citeauthoryear{Rodr{\'\i}guez and Egenhofer}{Rodr{\'\i}guez
  and Egenhofer}{2003}]%
        {rodriguez2003determining}
\bibfield{author}{\bibinfo{person}{M~Andrea Rodr{\'\i}guez} {and}
  \bibinfo{person}{Max~J. Egenhofer}.} \bibinfo{year}{2003}\natexlab{}.
\newblock \showarticletitle{Determining semantic similarity among entity
  classes from different ontologies}.
\newblock \bibinfo{journal}{\emph{IEEE transactions on knowledge and data
  engineering}} \bibinfo{volume}{15}, \bibinfo{number}{2}
  (\bibinfo{year}{2003}), \bibinfo{pages}{442--456}.
\newblock


\bibitem[\protect\citeauthoryear{Ruas, Grosky, and Aizawa}{Ruas
  et~al\mbox{.}}{2019}]%
        {RUAS2019288}
\bibfield{author}{\bibinfo{person}{Terry Ruas}, \bibinfo{person}{William
  Grosky}, {and} \bibinfo{person}{Akiko Aizawa}.}
  \bibinfo{year}{2019}\natexlab{}.
\newblock \showarticletitle{Multi-sense embeddings through a word sense
  disambiguation process}.
\newblock \bibinfo{journal}{\emph{Expert Systems with Applications}}
  \bibinfo{volume}{136} (\bibinfo{year}{2019}), \bibinfo{pages}{288 -- 303}.
\newblock
\showISSN{0957-4174}
\urldef\tempurl%
\url{https://doi.org/10.1016/j.eswa.2019.06.026}
\showDOI{\tempurl}


\bibitem[\protect\citeauthoryear{Rubenstein and Goodenough}{Rubenstein and
  Goodenough}{1965}]%
        {rubenstein1965contextual}
\bibfield{author}{\bibinfo{person}{Herbert Rubenstein} {and}
  \bibinfo{person}{John~B Goodenough}.} \bibinfo{year}{1965}\natexlab{}.
\newblock \showarticletitle{Contextual correlates of synonymy}.
\newblock \bibinfo{journal}{\emph{Commun. ACM}} \bibinfo{volume}{8},
  \bibinfo{number}{10} (\bibinfo{year}{1965}), \bibinfo{pages}{627--633}.
\newblock


\bibitem[\protect\citeauthoryear{S{\'a}nchez, Batet, and Isern}{S{\'a}nchez
  et~al\mbox{.}}{2011}]%
        {sanchez2011ontology}
\bibfield{author}{\bibinfo{person}{David S{\'a}nchez},
  \bibinfo{person}{Montserrat Batet}, {and} \bibinfo{person}{David Isern}.}
  \bibinfo{year}{2011}\natexlab{}.
\newblock \showarticletitle{Ontology-based information content computation}.
\newblock \bibinfo{journal}{\emph{Knowledge-based systems}}
  \bibinfo{volume}{24}, \bibinfo{number}{2} (\bibinfo{year}{2011}),
  \bibinfo{pages}{297--303}.
\newblock


\bibitem[\protect\citeauthoryear{Sanh, Debut, Chaumond, and Wolf}{Sanh
  et~al\mbox{.}}{2019}]%
        {sanh2019distilbert}
\bibfield{author}{\bibinfo{person}{Victor Sanh}, \bibinfo{person}{Lysandre
  Debut}, \bibinfo{person}{Julien Chaumond}, {and} \bibinfo{person}{Thomas
  Wolf}.} \bibinfo{year}{2019}\natexlab{}.
\newblock \showarticletitle{DistilBERT, a distilled version of BERT: smaller,
  faster, cheaper and lighter}.
\newblock \bibinfo{journal}{\emph{arXiv preprint arXiv:1910.01108}}
  (\bibinfo{year}{2019}).
\newblock


\bibitem[\protect\citeauthoryear{{\v{S}}ari{\'c}, Glava{\v{s}}, Karan,
  {\v{S}}najder, and Ba{\v{s}}i{\'c}}{{\v{S}}ari{\'c} et~al\mbox{.}}{2012}]%
        {vsaric2012takelab}
\bibfield{author}{\bibinfo{person}{Frane {\v{S}}ari{\'c}},
  \bibinfo{person}{Goran Glava{\v{s}}}, \bibinfo{person}{Mladen Karan},
  \bibinfo{person}{Jan {\v{S}}najder}, {and} \bibinfo{person}{Bojana~Dalbelo
  Ba{\v{s}}i{\'c}}.} \bibinfo{year}{2012}\natexlab{}.
\newblock \showarticletitle{Takelab: Systems for measuring semantic text
  similarity}. In \bibinfo{booktitle}{\emph{* SEM 2012: The First Joint
  Conference on Lexical and Computational Semantics--Volume 1: Proceedings of
  the main conference and the shared task, and Volume 2: Proceedings of the
  Sixth International Workshop on Semantic Evaluation (SemEval 2012)}}.
  \bibinfo{pages}{441--448}.
\newblock


\bibitem[\protect\citeauthoryear{Schnabel, Labutov, Mimno, and
  Joachims}{Schnabel et~al\mbox{.}}{2015}]%
        {schnabel2015evaluation}
\bibfield{author}{\bibinfo{person}{Tobias Schnabel}, \bibinfo{person}{Igor
  Labutov}, \bibinfo{person}{David Mimno}, {and} \bibinfo{person}{Thorsten
  Joachims}.} \bibinfo{year}{2015}\natexlab{}.
\newblock \showarticletitle{Evaluation methods for unsupervised word
  embeddings}. In \bibinfo{booktitle}{\emph{Proceedings of the 2015 conference
  on empirical methods in natural language processing}}.
  \bibinfo{pages}{298--307}.
\newblock


\bibitem[\protect\citeauthoryear{Severyn and Moschitti}{Severyn and
  Moschitti}{2012}]%
        {severyn2012structural}
\bibfield{author}{\bibinfo{person}{Aliaksei Severyn} {and}
  \bibinfo{person}{Alessandro Moschitti}.} \bibinfo{year}{2012}\natexlab{}.
\newblock \showarticletitle{Structural relationships for large-scale learning
  of answer re-ranking}. In \bibinfo{booktitle}{\emph{Proceedings of the 35th
  international ACM SIGIR conference on Research and development in information
  retrieval}}. \bibinfo{pages}{741--750}.
\newblock


\bibitem[\protect\citeauthoryear{Severyn, Nicosia, and Moschitti}{Severyn
  et~al\mbox{.}}{2013}]%
        {severyn2013learning}
\bibfield{author}{\bibinfo{person}{Aliaksei Severyn}, \bibinfo{person}{Massimo
  Nicosia}, {and} \bibinfo{person}{Alessandro Moschitti}.}
  \bibinfo{year}{2013}\natexlab{}.
\newblock \showarticletitle{Learning semantic textual similarity with
  structural representations}. In \bibinfo{booktitle}{\emph{Proceedings of the
  51st Annual Meeting of the Association for Computational Linguistics (Volume
  2: Short Papers)}}. \bibinfo{pages}{714--718}.
\newblock


\bibitem[\protect\citeauthoryear{Shao}{Shao}{2017}]%
        {shao2017hcti}
\bibfield{author}{\bibinfo{person}{Yang Shao}.}
  \bibinfo{year}{2017}\natexlab{}.
\newblock \showarticletitle{HCTI at SemEval-2017 Task 1: Use Convolutional
  Neural Network to evaluate semantic textual similarity}. In
  \bibinfo{booktitle}{\emph{Proceedings of the 11th International Workshop on
  Semantic Evaluation (SemEval-2017)}}. \bibinfo{pages}{130--133}.
\newblock


\bibitem[\protect\citeauthoryear{Shawe-Taylor, Cristianini,
  et~al\mbox{.}}{Shawe-Taylor et~al\mbox{.}}{2004}]%
        {shawe2004kernel}
\bibfield{author}{\bibinfo{person}{John Shawe-Taylor}, \bibinfo{person}{Nello
  Cristianini}, {et~al\mbox{.}}} \bibinfo{year}{2004}\natexlab{}.
\newblock \bibinfo{booktitle}{\emph{Kernel methods for pattern analysis}}.
\newblock \bibinfo{publisher}{Cambridge university press}.
\newblock


\bibitem[\protect\citeauthoryear{Silberer and Lapata}{Silberer and
  Lapata}{2014}]%
        {silberer2014learning}
\bibfield{author}{\bibinfo{person}{Carina Silberer} {and}
  \bibinfo{person}{Mirella Lapata}.} \bibinfo{year}{2014}\natexlab{}.
\newblock \showarticletitle{Learning grounded meaning representations with
  autoencoders}. In \bibinfo{booktitle}{\emph{Proceedings of the 52nd Annual
  Meeting of the Association for Computational Linguistics (Volume 1: Long
  Papers)}}. \bibinfo{pages}{721--732}.
\newblock


\bibitem[\protect\citeauthoryear{Sinoara, Camacho-Collados, Rossi, Navigli, and
  Rezende}{Sinoara et~al\mbox{.}}{2019}]%
        {SINOARA2019955}
\bibfield{author}{\bibinfo{person}{Roberta~A. Sinoara}, \bibinfo{person}{Jose
  Camacho-Collados}, \bibinfo{person}{Rafael~G. Rossi},
  \bibinfo{person}{Roberto Navigli}, {and} \bibinfo{person}{Solange~O.
  Rezende}.} \bibinfo{year}{2019}\natexlab{}.
\newblock \showarticletitle{Knowledge-enhanced document embeddings for text
  classification}.
\newblock \bibinfo{journal}{\emph{Knowledge-Based Systems}}
  \bibinfo{volume}{163} (\bibinfo{year}{2019}), \bibinfo{pages}{955 -- 971}.
\newblock
\showISSN{0950-7051}
\urldef\tempurl%
\url{https://doi.org/10.1016/j.knosys.2018.10.026}
\showDOI{\tempurl}


\bibitem[\protect\citeauthoryear{Soğancıoğlu, Öztürk, and
  Özgür}{Soğancıoğlu et~al\mbox{.}}{2017}]%
        {biosses}
\bibfield{author}{\bibinfo{person}{Gizem Soğancıoğlu},
  \bibinfo{person}{Hakime Öztürk}, {and} \bibinfo{person}{Arzucan Özgür}.}
  \bibinfo{year}{2017}\natexlab{}.
\newblock \showarticletitle{{BIOSSES: a semantic sentence similarity estimation
  system for the biomedical domain}}.
\newblock \bibinfo{journal}{\emph{Bioinformatics}} \bibinfo{volume}{33},
  \bibinfo{number}{14} (\bibinfo{date}{07} \bibinfo{year}{2017}),
  \bibinfo{pages}{i49--i58}.
\newblock
\showISSN{1367-4803}
\urldef\tempurl%
\url{https://doi.org/10.1093/bioinformatics/btx238}
\showDOI{\tempurl}
\showeprint{https://academic.oup.com/bioinformatics/article-pdf/33/14/i49/25157316/btx238.pdf}


\bibitem[\protect\citeauthoryear{Sultan, Bethard, and Sumner}{Sultan
  et~al\mbox{.}}{2014}]%
        {sultan2014dls}
\bibfield{author}{\bibinfo{person}{Md~Arafat Sultan}, \bibinfo{person}{Steven
  Bethard}, {and} \bibinfo{person}{Tamara Sumner}.}
  \bibinfo{year}{2014}\natexlab{}.
\newblock \showarticletitle{DLS@ CU: Sentence Similarity from Word Alignment}.
  In \bibinfo{booktitle}{\emph{Proceedings of the 8th International Workshop on
  Semantic Evaluation (SemEval 2014)}}. \bibinfo{pages}{241--246}.
\newblock


\bibitem[\protect\citeauthoryear{Sultan, Bethard, and Sumner}{Sultan
  et~al\mbox{.}}{2015}]%
        {sultan2015dls}
\bibfield{author}{\bibinfo{person}{Md~Arafat Sultan}, \bibinfo{person}{Steven
  Bethard}, {and} \bibinfo{person}{Tamara Sumner}.}
  \bibinfo{year}{2015}\natexlab{}.
\newblock \showarticletitle{Dls@ cu: Sentence similarity from word alignment
  and semantic vector composition}. In \bibinfo{booktitle}{\emph{Proceedings of
  the 9th International Workshop on Semantic Evaluation (SemEval 2015)}}.
  \bibinfo{pages}{148--153}.
\newblock


\bibitem[\protect\citeauthoryear{Sun, Wang, Li, Feng, Tian, Wu, and Wang}{Sun
  et~al\mbox{.}}{2020}]%
        {sun2020ernie}
\bibfield{author}{\bibinfo{person}{Yu Sun}, \bibinfo{person}{Shuohuan Wang},
  \bibinfo{person}{Yu-Kun Li}, \bibinfo{person}{Shikun Feng},
  \bibinfo{person}{Hao Tian}, \bibinfo{person}{Hua Wu}, {and}
  \bibinfo{person}{Haifeng Wang}.} \bibinfo{year}{2020}\natexlab{}.
\newblock \showarticletitle{ERNIE 2.0: A Continual Pre-Training Framework for
  Language Understanding.}. In \bibinfo{booktitle}{\emph{AAAI}}.
  \bibinfo{pages}{8968--8975}.
\newblock


\bibitem[\protect\citeauthoryear{Sánchez and Batet}{Sánchez and
  Batet}{2013}]%
        {SANCHEZ20131393}
\bibfield{author}{\bibinfo{person}{David Sánchez} {and}
  \bibinfo{person}{Montserrat Batet}.} \bibinfo{year}{2013}\natexlab{}.
\newblock \showarticletitle{A semantic similarity method based on information
  content exploiting multiple ontologies}.
\newblock \bibinfo{journal}{\emph{Expert Systems with Applications}}
  \bibinfo{volume}{40}, \bibinfo{number}{4} (\bibinfo{year}{2013}),
  \bibinfo{pages}{1393 -- 1399}.
\newblock
\showISSN{0957-4174}
\urldef\tempurl%
\url{https://doi.org/10.1016/j.eswa.2012.08.049}
\showDOI{\tempurl}


\bibitem[\protect\citeauthoryear{Sánchez, Batet, Isern, and Valls}{Sánchez
  et~al\mbox{.}}{2012}]%
        {SANCHEZ20127718}
\bibfield{author}{\bibinfo{person}{David Sánchez}, \bibinfo{person}{Montserrat
  Batet}, \bibinfo{person}{David Isern}, {and} \bibinfo{person}{Aida Valls}.}
  \bibinfo{year}{2012}\natexlab{}.
\newblock \showarticletitle{Ontology-based semantic similarity: A new
  feature-based approach}.
\newblock \bibinfo{journal}{\emph{Expert Systems with Applications}}
  \bibinfo{volume}{39}, \bibinfo{number}{9} (\bibinfo{year}{2012}),
  \bibinfo{pages}{7718 -- 7728}.
\newblock
\showISSN{0957-4174}
\urldef\tempurl%
\url{https://doi.org/10.1016/j.eswa.2012.01.082}
\showDOI{\tempurl}


\bibitem[\protect\citeauthoryear{Tai, Socher, and Manning}{Tai
  et~al\mbox{.}}{2015}]%
        {tai2015improved}
\bibfield{author}{\bibinfo{person}{Kai~Sheng Tai}, \bibinfo{person}{Richard
  Socher}, {and} \bibinfo{person}{Christopher~D Manning}.}
  \bibinfo{year}{2015}\natexlab{}.
\newblock \showarticletitle{Improved Semantic Representations From
  Tree-Structured Long Short-Term Memory Networks}. In
  \bibinfo{booktitle}{\emph{Proceedings of the 53rd Annual Meeting of the
  Association for Computational Linguistics and the 7th International Joint
  Conference on Natural Language Processing (Volume 1: Long Papers)}}.
  \bibinfo{pages}{1556--1566}.
\newblock


\bibitem[\protect\citeauthoryear{Tian, Zhou, Lan, and Wu}{Tian
  et~al\mbox{.}}{2017}]%
        {tian2017ecnu}
\bibfield{author}{\bibinfo{person}{Junfeng Tian}, \bibinfo{person}{Zhiheng
  Zhou}, \bibinfo{person}{Man Lan}, {and} \bibinfo{person}{Yuanbin Wu}.}
  \bibinfo{year}{2017}\natexlab{}.
\newblock \showarticletitle{Ecnu at semeval-2017 task 1: Leverage kernel-based
  traditional nlp features and neural networks to build a universal model for
  multilingual and cross-lingual semantic textual similarity}. In
  \bibinfo{booktitle}{\emph{Proceedings of the 11th international workshop on
  semantic evaluation (SemEval-2017)}}. \bibinfo{pages}{191--197}.
\newblock


\bibitem[\protect\citeauthoryear{Tien, Le, Tomohiro, and Tatsuya}{Tien
  et~al\mbox{.}}{2019}]%
        {TIEN2019102090}
\bibfield{author}{\bibinfo{person}{Nguyen~Huy Tien},
  \bibinfo{person}{Nguyen~Minh Le}, \bibinfo{person}{Yamasaki Tomohiro}, {and}
  \bibinfo{person}{Izuha Tatsuya}.} \bibinfo{year}{2019}\natexlab{}.
\newblock \showarticletitle{Sentence modeling via multiple word embeddings and
  multi-level comparison for semantic textual similarity}.
\newblock \bibinfo{journal}{\emph{Information Processing \& Management}}
  \bibinfo{volume}{56}, \bibinfo{number}{6} (\bibinfo{year}{2019}),
  \bibinfo{pages}{102090}.
\newblock
\showISSN{0306-4573}
\urldef\tempurl%
\url{https://doi.org/10.1016/j.ipm.2019.102090}
\showDOI{\tempurl}


\bibitem[\protect\citeauthoryear{Tissier, Gravier, and Habrard}{Tissier
  et~al\mbox{.}}{2017}]%
        {dict2vec}
\bibfield{author}{\bibinfo{person}{Julien Tissier}, \bibinfo{person}{Christophe
  Gravier}, {and} \bibinfo{person}{Amaury Habrard}.}
  \bibinfo{year}{2017}\natexlab{}.
\newblock \showarticletitle{Dict2vec: Learning Word Embeddings using Lexical
  Dictionaries}. In \bibinfo{booktitle}{\emph{Conference on Empirical Methods
  in Natural Language Processing (EMNLP 2017)}}. \bibinfo{pages}{254--263}.
\newblock


\bibitem[\protect\citeauthoryear{Vaswani, Shazeer, Parmar, Uszkoreit, Jones,
  Gomez, Kaiser, and Polosukhin}{Vaswani et~al\mbox{.}}{2017}]%
        {vaswani2017attention}
\bibfield{author}{\bibinfo{person}{Ashish Vaswani}, \bibinfo{person}{Noam
  Shazeer}, \bibinfo{person}{Niki Parmar}, \bibinfo{person}{Jakob Uszkoreit},
  \bibinfo{person}{Llion Jones}, \bibinfo{person}{Aidan~N Gomez},
  \bibinfo{person}{Lukasz Kaiser}, {and} \bibinfo{person}{Illia Polosukhin}.}
  \bibinfo{year}{2017}\natexlab{}.
\newblock \showarticletitle{Attention is All you Need}. In
  \bibinfo{booktitle}{\emph{NIPS}}.
\newblock


\bibitem[\protect\citeauthoryear{Wang, Smith, and Mitamura}{Wang
  et~al\mbox{.}}{2007}]%
        {Wang2007}
\bibfield{author}{\bibinfo{person}{Mengqiu Wang}, \bibinfo{person}{Noah~A.
  Smith}, {and} \bibinfo{person}{Teruko Mitamura}.}
  \bibinfo{year}{2007}\natexlab{}.
\newblock \showarticletitle{{What is the Jeopardy model? A quasi-synchronous
  grammar for QA}}.
\newblock \bibinfo{journal}{\emph{EMNLP-CoNLL 2007 - Proceedings of the 2007
  Joint Conference on Empirical Methods in Natural Language Processing and
  Computational Natural Language Learning}} \bibinfo{number}{June}
  (\bibinfo{year}{2007}), \bibinfo{pages}{22--32}.
\newblock


\bibitem[\protect\citeauthoryear{Wang, Mi, and Ittycheriah}{Wang
  et~al\mbox{.}}{2016}]%
        {Wang2016}
\bibfield{author}{\bibinfo{person}{Zhiguo Wang}, \bibinfo{person}{Haitao Mi},
  {and} \bibinfo{person}{Abraham Ittycheriah}.}
  \bibinfo{year}{2016}\natexlab{}.
\newblock \showarticletitle{{Sentence similarity learning by lexical
  decomposition and composition}}.
\newblock \bibinfo{journal}{\emph{COLING 2016 - 26th International Conference
  on Computational Linguistics, Proceedings of COLING 2016: Technical Papers}}
  \bibinfo{number}{challenge 2} (\bibinfo{year}{2016}),
  \bibinfo{pages}{1340--1349}.
\newblock
\showISBNx{9784879747020}
\showeprint{1602.07019}


\bibitem[\protect\citeauthoryear{Wu and Palmer}{Wu and Palmer}{1994}]%
        {wu1994verbs}
\bibfield{author}{\bibinfo{person}{Zhibiao Wu} {and} \bibinfo{person}{Martha
  Palmer}.} \bibinfo{year}{1994}\natexlab{}.
\newblock \showarticletitle{Verbs semantics and lexical selection}. In
  \bibinfo{booktitle}{\emph{Proceedings of the 32nd annual meeting on
  Association for Computational Linguistics}}. Association for Computational
  Linguistics, \bibinfo{pages}{133--138}.
\newblock


\bibitem[\protect\citeauthoryear{Yang, Dai, Yang, Carbonell, Salakhutdinov, and
  Le}{Yang et~al\mbox{.}}{2019}]%
        {yang2019xlnet}
\bibfield{author}{\bibinfo{person}{Zhilin Yang}, \bibinfo{person}{Zihang Dai},
  \bibinfo{person}{Yiming Yang}, \bibinfo{person}{Jaime Carbonell},
  \bibinfo{person}{Russ~R Salakhutdinov}, {and} \bibinfo{person}{Quoc~V Le}.}
  \bibinfo{year}{2019}\natexlab{}.
\newblock \showarticletitle{Xlnet: Generalized autoregressive pretraining for
  language understanding}. In \bibinfo{booktitle}{\emph{Advances in neural
  information processing systems}}. \bibinfo{pages}{5753--5763}.
\newblock


\bibitem[\protect\citeauthoryear{{Zhu} and {Iglesias}}{{Zhu} and
  {Iglesias}}{2017}]%
        {7572993}
\bibfield{author}{\bibinfo{person}{G. {Zhu}} {and} \bibinfo{person}{C.~A.
  {Iglesias}}.} \bibinfo{year}{2017}\natexlab{}.
\newblock \showarticletitle{Computing Semantic Similarity of Concepts in
  Knowledge Graphs}.
\newblock \bibinfo{journal}{\emph{IEEE Transactions on Knowledge and Data
  Engineering}} \bibinfo{volume}{29}, \bibinfo{number}{1} (\bibinfo{date}{Jan}
  \bibinfo{year}{2017}), \bibinfo{pages}{72--85}.
\newblock
\showISSN{2326-3865}
\urldef\tempurl%
\url{https://doi.org/10.1109/TKDE.2016.2610428}
\showDOI{\tempurl}


\bibitem[\protect\citeauthoryear{Zou, Socher, Cer, and Manning}{Zou
  et~al\mbox{.}}{2013}]%
        {zou2013bilingual}
\bibfield{author}{\bibinfo{person}{Will~Y Zou}, \bibinfo{person}{Richard
  Socher}, \bibinfo{person}{Daniel Cer}, {and} \bibinfo{person}{Christopher~D
  Manning}.} \bibinfo{year}{2013}\natexlab{}.
\newblock \showarticletitle{Bilingual word embeddings for phrase-based machine
  translation}. In \bibinfo{booktitle}{\emph{Proceedings of the 2013 Conference
  on Empirical Methods in Natural Language Processing}}.
  \bibinfo{pages}{1393--1398}.
\newblock


\end{thebibliography}
\begin{appendices}
\section{Semantic distance measures and their formulae}\label{appendix: A}
\begingroup

\setlength{\tabcolsep}{7pt} 
\renewcommand{\arraystretch}{2.5}
\begin{longtable}[c]{{|p{0.4cm}|p{4cm}|p{8cm}|}}
\hline
\textbf{SNo} & \textbf{Semantic distance measure} & \textbf{Formula}  \\ \hline
\endfirsthead
\multicolumn{3}{c}%
{{\bfseries Table \thetable\ continued from previous page}} \\
\hline
\textbf{SNo} & \textbf{Semantic distance measure} & \textbf{Formula}  \\ \hline
\endhead
1   & \fontsize{8}{7.2}\selectfont $ \alpha$ - skew divergence (ASD)  &  $ {\displaystyle \sum_{w\in C(w_1)\cup C(w_2)} P(w|w_1 ) log \frac{P(w|w_1)}{\alpha P(w|w_2) + (1-\alpha)P(w|w_1)}} $   \\ \hline
2   & \fontsize{8}{7.2}\selectfont Cosine similarity         & 
$ {\displaystyle \frac{\sum_{w\in C(w_1)\cup C(w_2)}P(w \mid w_1) \times P(w\mid w_2)}{\sqrt{\sum_{w\in C(w_1)} P(w\mid w_1)^2} \times \sqrt{\sum_{w\in C(w_2)} P(w | w_2)^2}} }$      \\ \hline
3   & \fontsize{8}{7.2}\selectfont Co-occurence Retrieval Models (CRM) & $ {\displaystyle \gamma \bigg[ \frac{2\times P \times R}{P + R}\bigg] + (1 - \gamma) \bigg[ \beta [P] + (1 - \beta ) [R] \bigg]} $ \\ \hline
4   & \fontsize{8}{7.2}\selectfont Dice coefficient & $ {\displaystyle \frac{2 \times \sum_{w\in C(w_1)\cup C(w_2)} \mbox{min} (P (w|w_1), P(w|w_2))} {\sum_{w\in C(w_1)} P (w|w_1) + \sum_{w\in C(w_2)} P(w|w_2) }} $ \\ \hline
5   & \fontsize{8}{7.2}\selectfont Manhattan Distance or L1 norm & $ {\displaystyle \sum_{w\in C(w_1)\cup C(w_2)} \big| P(w|w_1) - P(w|w_2)\big| } $ \\ \hline
6   & \fontsize{8}{7.2}\selectfont Division measure & $ {\displaystyle \sum_{w\in C(w_1)\cup C(w_2)} \bigg| \mbox{log} \frac{P(w|w_1)}{P(w|w_2)} \bigg| } $ \\ \hline
7   & \fontsize{8}{7.2}\selectfont Hindle  & $ {\sum_{w \in C(w)} \begin{dcases} \mbox{min} ( I (w, w_1) , I (w,w_2)), \\ \mbox{\hspace{1cm} if both} I(w,w_1) \mbox{and} I(w, w_2) > 0\\ |\mbox{max} ( I (w, w_1) , I (w,w_2))|, \\ \mbox{\hspace{1cm}if both} I(w,w_1) \mbox{and} I(w, w_2) < 0\\ 0 , \mbox{otherwise} \end{dcases}}$ \\\hline
8   & \fontsize{8}{7.2}\selectfont Jaccard & $ {\displaystyle \frac{\sum_{w\in C(w_1)\cup C(w_2)} \mbox{min}(P(w|w_1), P(w|w_2))}{\sum_{w\in C(w_1)\cup C(w_2)} \mbox{max}(P(w|w_1), P(w|w_2))}} $ \\ \hline
9   & \fontsize{8}{7.2}\selectfont Jensen-Shannon divergence (JSD) &  ${ \begin{array}{c}
    \sum_{w\in C(w_1)\cup C(w_2)} \bigg( P(w|w_1)  \text{log} \frac{P(w|w_1)}{\frac{1}{2} (P(w|w_1) + P (w|w_2)) } \\+ P (w|w_2) \text{log} \frac{P(w|w_2)}{\frac{1}{2} (P(w|w_1) + P(w|w_2))}\bigg)  \end{array}}$  \\ \hline
10  & \fontsize{8}{7.2}\selectfont Kullback-Leibler divergence - common occurance & $ {\displaystyle \sum_{w\in C(w_1)\cup C(w_2)} P(w|w_1) \mbox{log} \frac{P(w|w_1)}{P(w|w_2)} }  $ \\ \hline
11  & \fontsize{8}{7.2}\selectfont Kullback-Leibler divergence - absolute & $ {\displaystyle \sum_{w\in C(w_1)\cup C(w_2)} P(w|w_1) \bigg| \mbox{log} \frac{P(w|w_1)}{P(w|w_2)} \bigg| }  $ \\ \hline
12  & \fontsize{8}{7.2}\selectfont Kullback-Leibler divergence - average & $ {\displaystyle \frac{1}{2}\sum_{w\in C(w_1)\cup C(w_2)} (P(w|w_1) - P(w|w_2)) \mbox{log} \frac{P(w|w_1)}{P(w|w_2)}  }  $ \\ \hline
13  & \fontsize{8}{7.2}\selectfont Kullback-Leibler divergence - maximum & $ {\displaystyle \mbox{max} (KLD (w_1, w_2), KLD (w_2,w_1)) }  $ \\ \hline
14  & \fontsize{8}{7.2}\selectfont Euclidean Distance or L2 norm & $ {\displaystyle \sqrt{\sum_{w\in C(w_1)\cup C(w_2)} (P(w|w_1) - P(w|w_2))^2} } $ \\ \hline
15  & \fontsize{8}{7.2}\selectfont Lin & $ { \displaystyle \frac{\sum_{(r,w) \in T(w_1)\cap T(w_2)} (I (w_1,r,w) + I(w_2,r,w))}{\sum(r,w')\in T(w_1) I(w_1,r,w') + \sum_{(r,w'')\in T(w_2)} I(w_2,r,w'')} } $ \\ \hline
16  & \fontsize{8}{7.2}\selectfont Product measure & $ { \displaystyle \sum_{w \in C(w_1) \cup C(w_2)} \frac{P(w|w_1) \times P(w|w_2)}{(\frac{1}{2} (P(w|w_1) + P(w|w_2)))^2} } $ \\ \hline
\caption{Table of semantic measures and their formulae - adapted from Mohammad and Hurst\cite{mohammad2012distributional}}
\label{tab:tab4}\\
\end{longtable}
\endgroup

\section{Table of References}\label{appendix: B}

\begin{tiny}
\begin{longtable}[c]{{|p{0.3cm}|p{3cm}|p{0.3cm}|p{3cm}|p{3cm}|p{0.3cm}|p{0.3cm}|p{1cm}|}}
\hline
\textbf{Cita-tion} &
  \textbf{Title} &
  \textbf{Year} &
  \textbf{Authors} &
  \textbf{Venue} &
  \textbf{SJR Quartile} &
  \textbf{H- Index} &
  \textbf{Citations as on 02.04.2020} \\ \hline
\endfirsthead
\multicolumn{8}{c}%
{{\bfseries Table \thetable\ continued from previous page}} \\
\hline
\textbf{Cita-tion} &
  \textbf{Title} &
  \textbf{Year} &
  \textbf{Authors} &
  \textbf{Venue} &
  \textbf{SJR Quartile} &
  \textbf{H- Index} &
  \textbf{Citations as on 02.04.2020} \\ \hline
\endhead
 \cite{agirre2009study} &
  A study on similarity and relatedness using distributional and wordnet-based approaches &
  2009 &
  Agirre, Eneko and Alfonseca, Enrique and Hall, Keith and Kravalova, Jana and Pasca, Marius and Soroa, Aitor &
  Human Language Technologies: The 2009 Annual Conference of the North American Chapter of the Association for Computational Linguistics &
  &
  61 &
  809 \\ \hline
 \cite{agirre2015semeval} &
  Semeval-2015 task 2: Semantic textual similarity, english, spanish and pilot on interpretability &
  2015 &
  Agirre, Eneko and Banea, Carmen and Cardie, Claire and Cer, Daniel and Diab, Mona and Gonzalez-Agirre, Aitor and Guo, Weiwei and Lopez-Gazpio, Inigo and Maritxalar, Montse and Mihalcea, Rada and others &
  Proceedings of the 9th international workshop on semantic evaluation (SemEval 2015) &
  &
  49 &
  242 \\ \hline
 \cite{agirre2014semeval} &
  Semeval-2014 task 10: Multilingual semantic textual similarity &
  2014 &
  Agirre, Eneko and Banea, Carmen and Cardie, Claire and Cer, Daniel and Diab, Mona and Gonzalez-Agirre, Aitor and Guo, Weiwei and Mihalcea, Rada and Rigau, German and Wiebe, Janyce &
  Proceedings of the 8th international workshop on semantic evaluation (SemEval 2014) &
  &
  49 &
  220 \\ \hline
\cite{agirre2016semeval} &
  Semeval-2016 task 1: Semantic textual similarity, monolingual and cross-lingual evaluation &
  2016 &
  Agirre, Eneko and Banea, Carmen and Cer, Daniel and Diab, Mona and Gonzalez Agirre, Aitor and Mihalcea, Rada and Rigau Claramunt, German and Wiebe, Janyce &
  SemEval-2016. 10th International Workshop on Semantic Evaluation; &
  &
  49 &
  200 \\ \hline
 \cite{agirre2012semeval} &
  Semeval-2012 task 6: A pilot on semantic textual similarity &
  2012 &
  Agirre, Eneko and Cer, Daniel and Diab, Mona and Gonzalez-Agirre, Aitor &
  Proceedings of the Sixth International Workshop on Semantic Evaluation (SemEval 2012) &
  &
  49 &
  498 \\ \hline
 \cite{agirre2013sem} &
  SEM 2013 shared task: Semantic textual similarity &
  2013 &
  Agirre, Eneko and Cer, Daniel and Diab, Mona and Gonzalez-Agirre, Aitor and Guo, Weiwei &
  Second Joint Conference on Lexical and Computational Semantics (* SEM), Volume 1: Proceedings of the Main Conference and the Shared Task: Semantic Textual Similarity &
  &
  49 &
  268 \\ \hline
\cite{AlexanderM.Rush2015} &
  A Neural Attention Model for Abstractive Sentence &
  2015 &
  Alexander M. Rush, Sumit Chopra and Jason Weston &
  EMNLP &
  &
  88 &
  1350 \\ \hline
\cite{ALTINEL20181129} &
  Semantic text classification: A survey of past and recent advances &
  2018 &
  Altinel, Berna and Ganiz, Murat Can &
  Information Processing \& Management &
  Q1 &
  88 &
  29 \\ \hline
\cite{amir2017sentence} &
  Sentence similarity based on semantic kernels for intelligent text retrieval &
  2017 &
  Amir, Samir and Tanasescu, Adrian and Zighed, Djamel A &
  Journal of Intelligent Information Systems &
  Q2 &
  52 &
  8
   \\ \hline
\cite{neuralnetwork} &
  Neural machine translation by jointly learning to align and translate &
  2015 &
  Dzmitry Bahdanau and Kyunghyun Cho and Yoshua Bengio &
  International Conference on Learning Representations &
  &
  150 &
  10967 \\ \hline
\cite{banerjee2003extended} &
  Extended gloss overlaps as a measure of semantic relatedness &
  2003 &
  Banerjee, Satanjeev and Pedersen, Ted &
  IJCAI &
  &
  109 &
  953 \\ \hline
\cite{bar2012ukp} &
  Ukp: Computing semantic textual similarity by combining multiple content similarity measures &
  2012 &
  B{\"a}r, Daniel and Biemann, Chris and Gurevych, Iryna and Zesch, Torsten &
  Proceedings of the Sixth International Workshop on Semantic Evaluation (SemEval 2012) &
  &
  50 &
  227 \\ \hline
\cite{baroni2009wacky} &
  The WaCky wide web: a collection of very large linguistically processed web-crawled corpora &
  2009 &
  Baroni, Marco and Bernardini, Silvia and Ferraresi, Adriano and Zanchetta, Eros &
  Language resources and evaluation &
  &
  40 &
  1130 \\ \hline
\cite{baroni2014don} &
  Don’t count, predict! a systematic comparison of context-counting vs. context-predicting semantic vectors &
  2014 &
  Baroni, Marco and Dinu, Georgiana and Kruszewski, Germán &
  Proceedings of the 52nd Annual Meeting of the Association for Computational Linguistics (Volume 1: Long Papers) &
  &
  106 &
  1166 \\ \hline
\cite{beltagy2019scibert} &
  SciBERT: A Pretrained Language Model for Scientific Text &
  2019 &
  Beltagy, Iz and Lo, Kyle and Cohan, Arman &
  EMNLP &
  &
  88 &
  74 \\ \hline
\cite{BENEDETTI2019136} &
  Computing inter-document similarity with Context Semantic Analysis &
  2019 &
  Fabio Benedetti and Domenico Beneventano and Sonia Bergamaschi and Giovanni Simonini &
  Information Systems &
  Q1 &
  76 &
  24 \\ \hline
\cite{bizer2009dbpedia} &
  DBpedia-A crystallization point for the Web of Data &
  2009 &
  Bizer, Christian and Lehmann, Jens and Kobilarov, Georgi and Auer, S{\"o}ren and Becker, Christian and Cyganiak, Richard and Hellmann, Sebastian &
  Journal of web semantics &
  &
  28 &
  2331 \\ \hline
\cite{bojanowski2017enriching} &
  Enriching word vectors with subword information &
  2017 &
  Bojanowski, Piotr and Grave, Edouard and Joulin, Armand and Mikolov, Tomas &
  Transactions of the Association for Computational Linguistics &
  &
  47 &
  2935 \\ \hline
\cite{bordes2014question} &
  Question Answering with Subgraph Embeddings &
  2014 &
  Bordes, Antoine and Chopra, Sumit and Weston, Jason &
  EMNLP &
  &
  88 &
  433 \\ \hline
\cite{camacho2018word} &
  From Word to Sense Embeddings: A Survey on Vector Representations of Meaning &
  2018 &
  Camacho-Collados, Jose and Pilehvar, Mohammad Taher &
  Journal of Artificial Intelligence Research &
  Q1 &
  103 &
  69 \\ \hline
\cite{camacho2015nasari} &
  Nasari: a novel approach to a semantically-aware representation of items &
  2015 &
  Camacho-Collados, Jos\{\textbackslash{}'e\} and Pilehvar, Mohammad Taher and Navigli, Roberto &
  Proceedings of the 2015 Conference of the North American Chapter of the Association for Computational Linguistics: Human Language Technologies &
  &
  61 &
  74 \\ \hline
\cite{CAMACHOCOLLADOS201636} &
  Nasari: Integrating explicit knowledge and corpus statistics for a multilingual representation of concepts and entities &
  2016 &
  José Camacho-Collados and Mohammad Taher Pilehvar and Roberto Navigli &
  Artificial Intelligence &
  Q1 &
  135 &
  117 \\ \hline
\cite{cancedda2003word} &
  Word-sequence kernels &
  2003 &
  Cancedda, Nicola and Gaussier, Eric and Goutte, Cyril and Renders, Jean-Michel &
  Journal of machine learning research &
  Q1&
  188 &
  291 \\ \hline
\cite{cer2017semeval} &
  Semeval-2017 task 1: Semantic textual similarity-multilingual and cross-lingual focused evaluation &
  2017 &
  Cer, Daniel and Diab, Mona and Agirre, Eneko and Lopez-Gazpio, Inigo and Specia, Lucia &
  Proceedings of the 11th International Workshop on Semantic Evaluation (SemEval-2017) &
  &
  49 &
  227 \\ \hline
\cite{cilibrasi2007google} &
  The google similarity distance &
  2007 &
  Cilibrasi, Rudi L and Vitanyi, Paul MB &
  IEEE Transactions on knowledge and data engineering &
  Q1 &
  148 &
  2042 \\ \hline
\cite{collins2002convolution} &
  Convolution kernels for natural language &
  2002 &
  Collins, Michael and Duffy, Nigel &
  Advances in neural information processing systems &
  Q1 &
  169 &
  1118 \\ \hline
 \cite{collins2002new} &
  New ranking algorithms for parsing and tagging: Kernels over discrete structures, and the voted perceptron &
  2002 &
  Collins, Michael and Duffy, Nigel &
  Proceedings of the 40th Annual Meeting of the Association for Computational Linguistics &
  &
  135 &
  671
   \\ \hline
\cite{croce2017deep} &
  Deep learning in semantic kernel spaces &
  2017 &
  Croce, Danilo and Filice, Simone and Castellucci, Giuseppe and Basili, Roberto &
  Proceedings of the 55th Annual Meeting of the Association for Computational Linguistics (Volume 1: Long Papers) &
  &
  106 &
  15 \\ \hline
\cite{devlin2019bert} &
  BERT: Pre-training of Deep Bidirectional Transformers for Language Understanding &
  2019 &
  Devlin, Jacob and Chang, Ming-Wei and Lee, Kenton and Toutanova, Kristina &
  Proceedings of the 2019 Conference of the North American Chapter of the Association for Computational Linguistics: Human Language Technologies, Volume 1 (Long and Short Papers) &
  &
  61 &
  7390 \\ \hline
\cite{finkelstein2001placing} &
  Placing search in context: The concept revisited &
  2001 &
  Finkelstein, Lev and Gabrilovich, Evgeniy and Matias, Yossi and Rivlin, Ehud and Solan, Zach and Wolfman, Gadi and Ruppin, Eytan &
  Proceedings of the 10th international conference on World Wide Web &
  &
  70 &
  1768 \\ \hline
\cite{gabrilovich2007computing} &
  Computing semantic relatedness using wikipedia-based explicit semantic analysis. &
  2007 &
  Gabrilovich, Evgeniy and Markovitch, Shaul and others &
  IJCAI &
  &
  109 &
  2514 \\ \hline
\cite{ganitkevitch2013ppdb} &
  PPDB: The paraphrase database &
  2013 &
  Ganitkevitch, Juri and Van Durme, Benjamin and Callison-Burch, Chris &
  Proceedings of the 2013 Conference of the North American Chapter of the Association for Computational Linguistics: Human Language Technologies &
  &
  61 &
  493 \\ \hline
\cite{GAO201580} &
  A WordNet-based semantic similarity measurement combining edge-counting and information content theory &
  2015 &
  Jian-Bo Gao and Bao-Wen Zhang and Xiao-Hua Chen &
  Engineering Applications of Aritifical Intelligence &
  Q1 &
  86 &
  74 \\ \hline
\cite{gerz2016simverb} &
  SimVerb-3500: A Large-Scale Evaluation Set of Verb Similarity &
  2016 &
  Gerz, Daniela and Vuli\{\textbackslash{}'c\}, Ivan and Hill, Felix and Reichart, Roi and Korhonen, Anna &
  EMNLP &
  &
  88 &
  113 \\ \hline
\cite{GLAVAS20181} &
  A resource-light method for cross-lingual semantic textual similarity &
  2018 &
  Goran Glavaš and Marc Franco-Salvador and Simone P. Ponzetto and Paolo Rosso &
  Knowledge-based Systems &
  Q1 &
  94 &
  13 \\ \hline
\cite{gorman2006scaling}&
  Scaling distributional similarity to large corpora &
  2006 &
  Gorman, James and Curran, James R &
  44th Annual Meeting of the Association for Computational Linguistics &
   &
  135 &
  54 \\ \hline
\cite{HadjTaieb2019} &
  A survey of semantic relatedness evaluation datasets and procedures &
  2019 &
  Hadj Taieb, Mohamed Ali and Zesch, Torsten and Ben Aouicha, Mohamed &
  Artificial Intelligence Review &
  Q1 &
  63 &
  -- \\ \hline
\cite{hassan2019uests} &
  UESTS: An Unsupervised Ensemble Semantic Textual Similarity Method &
  2019 &
  Hassan, Basma and Abdelrahman, Samir E and Bahgat, Reem and Farag, Ibrahim &
  IEEE Access &
  Q1 &
  56 &
  1 \\ \hline
\cite{he-lin-2016-pairwise} &
  Pairwise Word Interaction Modeling with Deep Neural Networks for Semantic Similarity Measurement &
  2016 &
  He, Hua  and   Lin, Jimmy &
  Proceedings of the 2016 Conference of the North American Chapter of the Association for Computational Linguistics: Human Language Technologies &
  &
  61 &
  140 \\ \hline
\cite{hill2015simlex} &
  Simlex-999: Evaluating semantic models with (genuine) similarity estimation &
  2015 &
  Hill, Felix and Reichart, Roi and Korhonen, Anna &
  Computational Linguistics &
  Q2 &
  85 &
  728 \\ \hline
\cite{hoffart2013yago2} &
  YAGO2: A spatially and temporally enhanced knowledge base from Wikipedia &
  2013 &
  Hoffart, Johannes and Suchanek, Fabian M and Berberich, Klaus and Weikum, Gerhard &
  Artificial Intelligence &
  Q1 &
  135 &
  1064 \\ \hline
\cite{janda2019syntactic} &
  Syntactic, Semantic and Sentiment Analysis: The Joint Effect on Automated Essay Evaluation &
  2019 &
  Janda, Harneet Kaur and Pawar, Atish and Du, Shan and Mago, Vijay &
  IEEE Access &
  Q1 &
  56 &
  -- \\ \hline
\cite{jiang1997semantic} &
  Semantic similarity based on corpus statistics and lexical taxonomy &
  1997 &
  Jiang, Jay J and Conrath, David W &
  COLING &
  &
  41 &
  3682 \\ \hline
\cite{JIANG2017248} &
  Wikipedia-based information content and semantic similarity computation &
  2017 &
  Yuncheng Jiang and Wen Bai and Xiaopei Zhang and Jiaojiao Hu &
  Information Processing \& Management &
  Q1 &
  88 &
  43 \\ \hline
\cite{jiang2015feature} &
  Feature-based approaches to semantic similarity assessment of concepts using Wikipedia &
  2015 &
  Jiang, Yuncheng and Zhang, Xiaopei and Tang, Yong and Nie, Ruihua &
  Information Processing \& Management &
  Q1 &
  88 &
  55 \\ \hline
\cite{jiao2019tinybert} &
  Tinybert: Distilling bert for natural language understanding &
  2019 &
  Jiao, Xiaoqi and Yin, Yichun and Shang, Lifeng and Jiang, Xin and Chen, Xiao and Li, Linlin and Wang, Fang and Liu, Qun &
  arXiv &
  &
  &
  56 \\ \hline
\cite{kajiwara2016building} &
  Building a monolingual parallel corpus for text simplification using sentence similarity based on alignment between word embeddings &
  2016 &
  Kajiwara, Tomoyuki and Komachi, Mamoru &
  COLING  &
  &
  49 &
  39
  \\ \hline
\cite{kim2017bridging} &
  Bridging the gap: Incorporating a semantic similarity measure for effectively mapping PubMed queries to documents &
  2017 &
  Kim, Sun and Fiorini, Nicolas and Wilbur, W John and Lu, Zhiyong &
  Journal of biomedical informatics &
  Q1 &
  83 &
  14 \\ \hline
\cite{kim2014convolutional} &
  Convolutional Neural Networks for Sentence Classification &
  2014 &
  Kim, Yoon &
  EMNLP &
  &
  88 &
  6790 \\ \hline
\cite{lan2019albert} &
  ALBERT: A Lite BERT for Self-supervised Learning of Language Representations &
  2019 &
  Lan, Zhenzhong and Chen, Mingda and Goodman, Sebastian and Gimpel, Kevin and Sharma, Piyush and Soricut, Radu &
  International Conference on Learning Representations &
  &
  150 &
  270 \\ \hline
\cite{landauer1997solution} &
  A solution to Plato's problem: The latent semantic analysis theory of acquisition, induction, and representation of knowledge. &
  1997 &
  Landauer, Thomas K and Dumais, Susan T &
  Psychological review &
  Q1 &
  192 &
  6963 \\ \hline
\cite{landauer1998introduction} &
  An introduction to latent semantic analysis &
  1998 &
  Landauer, Thomas K and Foltz, Peter W and Laham, Darrell &
  Discourse Processes &
  Q1 &
  50 &
  5752 \\ \hline
\cite{lastra2015new} &
  A new family of information content models with an experimental survey on WordNet &
  2015 &
  Lastra-D{\'\i}az, Juan J and Garc{\'\i}a-Serrano, Ana &
  Knowledge-Based Systems &
  Q1 &
  94 &
  12 \\ \hline
\cite{lastra2017hesml} &
  HESML: A scalable ontology-based semantic similarity measures library with a set of reproducible experiments and a replication dataset &
  2017 &
 Lastra-D{\'\i}az, Juan J and Garc{\'\i}a-Serrano, Ana and Batet, Montserrat and Fern{\'a}ndez, Miriam and Chirigati, Fernando &
  Information Systems &
  Q1 &
  76 &
  27 \\ \hline
\cite{LASTRADIAZ2019645} &
  A reproducible survey on word embeddings and ontology-based methods forword similarity: Linear combinations outperform the state of the art &
  2019 &
  Juan J. Lastra-Díaz and Josu Goikoetxea and Mohamed Ali Hadj Taieb and Ana García-Serrano and Mohamed Ben Aouicha and Eneko Agirre &
  Engineering Applications of Aritificial Intelligence &
  Q1 &
  86 &
  7 \\ \hline
\cite{para2vec} &
  Distributed representations of sentences and documents &
  2014 &
  Le, Quoc and Mikolov, Tomas &
  International conference on machine learning &
  &
  135 &
  5406
  \\ \hline
\cite{le2018acv} &
  ACV-tree: A New Method for Sentence Similarity Modeling. &
  2018 &
  Le, Yuquan and Wang, Zhi-Jie and Quan, Zhe and He, Jiawei and Yao, Bin &
  IJCAI &
  &
  109 &
  4 \\ \hline
\cite{lee2011novel} &
  A novel sentence similarity measure for semantic-based expert systems. &
  2011 &
  Lee, Ming Che &
  Expert Systems with Applications &
  Q1 &
  162 &
  47 \\ \hline
\cite{levy2014dependency} &
  Dependency-based word embeddings &
  2014 &
  Levy, Omer and Goldberg, Yoav &
  Proceedings of the 52nd Annual Meeting of the Association for Computational Linguistics &
  &
  106 &
  860 \\ \hline
 \cite{levy2014neural} &
  Neural word embedding as implicit matrix factorization &
  2014 &
  Levy, Omer and Goldberg, Yoav &
  Book &
  &
  &
  1480 \\ \hline
\cite{li2013computing} &
  Computing term similarity by large probabilistic isa knowledge &
  2013 &
  Li, Peipei and Wang, Haixun and Zhu, Kenny Q and Wang, Zhongyuan and Wu, Xindong &
  Proceedings of the 22nd ACM international conference on Information \& Knowledge Management &
  &
  48 &
  56 \\ \hline
\cite{li2003approach} &
  An approach for measuring semantic similarity between words using multiple information sources &
  2003 &
  Li, Yuhua and Bandar, Zuhair A and McLean, David &
  IEEE Transactions on knowledge and data engineering &
  Q1 &
  148 &
  1315 \\ \hline
\cite{li2006sentence} &
  Sentence similarity based on semantic nets and corpus statistics &
  2006 &
  Li, Yuhua and McLean, David and Bandar, Zuhair A and O'shea, James D and Crockett, Keeley &
  IEEE transactions on knowledge and data engineering &
  Q1 &
  148 &
  849 \\ \hline
\cite{lin1998information} &
  An information-theoretic definition of similarity. &
  1998 &
  Lin &
  ICML &
  &
  135 &
  5263 \\ \hline
\cite{liu2019roberta} &
  Roberta: A robustly optimized bert pretraining approach &
  2019 &
  Liu, Yinhan and Ott, Myle and Goyal, Naman and Du, Jingfei and Joshi, Mandar and Chen, Danqi and Levy, Omer and Lewis, Mike and Zettlemoyer, Luke and Stoyanov, Veselin &
  arXiv &
  &
  &
  229 \\ \hline
\cite{LOPEZGAZPIO2017186} &
  Interpretable semantic textual similarity: Finding and explaining differences between sentences &
  2017 &
  I. Lopez-Gazpio, M. Maritxalar, A. Gonzalez-Agirre, G. Rigau, L. Uria, E. Agirre, &
  Knowledge-based Systems &
  Q1 &
  94 &
  16 \\ \hline
\cite{LOPEZGAZPIO20191} &
  Word n-gram attention models for sentence similarity and inference &
  2019 &
  I. Lopez-Gazpio and M. Maritxalar and M. Lapata and E. Agirre &
  Expert Systems with Applications &
  Q1 &
  162 &
  2 \\ \hline
\cite{lund1996producing} &
  Producing high-dimensional semantic spaces from lexical co-occurrence &
  1996 &
  Lund, Kevin and Burgess, Curt &
  Behavior research methods &
  Q1 &
  114 &
  1869 \\ \hline
\cite{marellisick} &
  A SICK cure for the evaluation of compositional distributional semantic models &
  2014 &
  Marelli, M and Menini, S and Baroni, M and Bentivogli, L and Bernardi, R and Zamparelli, R &
  International Conference on Language Resources and Evaluation (LREC) &
  &
  45 &
  464 \\ \hline
\cite{mccann2017learned} &
  Learned in translation: Contextualized word vectors &
  2017 &
 McCann, Bryan and Bradbury, James and Xiong, Caiming and Socher, Richard &
  NIPS &
  &
  169 &
  376 \\ \hline
\cite{mcinnes2013umls} &
  UMLS:: Similarity: Measuring the Relatedness and Similarity of Biomedical Concept &
  2013 &
  McInnes, Bridget T and Liu, Ying and Pedersen, Ted and Melton, Genevieve B and Pakhomov, Serguei V &
  Human Language Technologies: The 2013 Annual Conference of the North American Chapter of the Association for Computational Linguistics &
  &
  61 &
  14 \\ \hline
\cite{Meek2018} &
  WIKIQA : A Challenge Dataset for Open-Domain Question Answering &
  2018 &
  Meek, Wen-tau Yih Christopher &
  EMNLP &
  &
  88 &
  351 \\ \hline
\cite{context2vec} &
  context2vec: Learning generic context embedding with bidirectional LSTM &
  2016 &
  Melamud, Oren and Goldberger, Jacob and Dagan, Ido &
  Proceedings of The 20th SIGNLL Conference on Computational Natural Language Learning &
  &
  34 &
  198 \\ \hline
\cite{mihalcea2007wikify} &
  Wikify! Linking documents to encyclopedic knowledge &
  2007 &
  Mihalcea, Rada and Csomai, Andras &
  Proceedings of the sixteenth ACM conference on Conference on information and knowledge management &
  &
  48 &
  1120 \\ \hline
\cite{mikolov2013efficient} &
  Efficient estimation of word representations in vector space &
  2013 &
  Mikolov, Tomas and Chen, Kai and Corrado, Greg and Dean, Jeffrey &
  arXiv &
  &
  &
  14807 \\ \hline
\cite{mikolov2013linguistic} &
  Linguistic regularities in continuous space word representations &
  2013 &
  Mikolov, Tomas and Yih, Wen-tau and Zweig, Geoffrey &
  Proceedings of the 2013 conference of the north american chapter of the association for computational linguistics: Human language technologies &
  &
  61 &
  2663 \\ \hline
\cite{miller1995wordnet} &
  WordNet: a lexical database for English &
  1995 &
  Miller, George A &
  Communications of the ACM &
  Q1 &
  189 &
  13223 \\ \hline
\cite{miller1991contextual} &
  Contextual correlates of semantic similarity &
  1991 &
  Miller, George A and Charles, Walter G &
  Language and cognitive processes &
  &
  &
  1727 \\ \hline
\cite{mnih2013learning} &
  Learning word embeddings efficiently with noise-contrastive estimation &
  2013 &
  Mnih, Andriy and Kavukcuoglu, Koray &
  Advances in neural information processing systems &
  Q1 &
  169 &
  495 \\ \hline
\cite{mohamed2019srl} &
  SRL-ESA-TextSum: A text summarization approach based on semantic role labeling and explicit semantic analysis &
  2019 &
  Mohamed, Muhidin and Oussalah, Mourad &
  Information Processing \& Management &
  Q1 &
  88 &
  2 \\ \hline
\cite{mohammad2012distributional} &
  Distributional measures of semantic distance: A survey &
  2012 &
  Mohammad, Saif M and Hirst, Graeme &
  arXiv &
   &
   &
  51 \\ \hline
\cite{moschitti2006efficient} &
  Efficient convolution kernels for dependency and constituent syntactic trees &
  2006 &
  Moschitti, Alessandro &
  European Conference on Machine Learning &
  &
  31 &
  493\\ \hline
\cite{moschitti2008kernel} &
 Kernel methods, syntax and semantics for relational text categorization &
  2008 &
  Moschitti, Alessandro &
  Proceedings of the 17th ACM conference on Information and knowledge management &
  &
  54 &
  105 \\ \hline
\cite{moschitti2008tree} &
  Tree kernels for semantic role labeling &
  2008 &
  Moschitti, Alessandro and Pighin, Daniele and Basili, Roberto &
  Computational Linguistics &
  Q1 &
  92 &
  180\\ \hline
\cite{moschitti2008kernels} &
  Kernels on linguistic structures for answer extraction &
  2008 &
  Moschitti, Alessandro and Quarteroni, Silvia &
  Proceedings of ACL-08: HLT, Short Papers &
   &
  90 &
  34\\ \hline
\cite{moschitti2007exploiting} &
  Exploiting syntactic and shallow semantic kernels for question answer classification &
  2007 &
  Moschitti, Alessandro and Quarteroni, Silvia and Basili, Roberto and Manandhar, Suresh &
  Annual meeting of the association of computational linguistics &
   &
  135 &
  229\\ \hline
\cite{moschitti2007fast} &
    Fast and effective kernels for relational learning from texts &
   2008 &
   Moschitti, Alessandro and Zanzotto, Fabio Massimo &
   International conference on machine learning &
   &
  135 &
  56\\ \hline
\cite{navigli2012babelnet} &
  BabelNet: The automatic construction, evaluation and application of a wide-coverage multilingual semantic network &
  2012 &
  Navigli, Roberto and Ponzetto, Simone Paolo &
  Artificial Intelligence &
  Q1 &
  135 &
  1110 \\ \hline
\cite{nelson2004university} &
  The University of South Florida free association, rhyme, and word fragment norms &
  2004 &
  Nelson, Douglas L and McEvoy, Cathy L and Schreiber, Thomas A &
  Behavior Research Methods, Instruments, \& Computers &
  Q1 &
  114 &
  2162 \\ \hline
\cite{nivre2006inductive} &
  Inductive Dependency Parsing &
  2006 &
  J Nivre &
  Book &
  &
  &
  313 \\ \hline
\cite{pagliardini2018unsupervised} &
  Unsupervised Learning of Sentence Embeddings using Compositional n-Gram Features &
  2018 &
  Pagliardini, Matteo and Gupta, Prakhar and Jaggi, Martin &
  North American Chapter of the Association for Computational Linguistics: Human Language Technologies &
  &
  61 &
  233 \\ \hline
\cite{parikh2016decomposable} &
  A Decomposable Attention Model for Natural Language Inference &
  2016 &
  Parikh, Ankur and Tackstrom, Oscar and Das, Dipanjan and Uszkoreit, Jakob &
  EMNLP &
  &
  88 &
  550 \\ \hline
\cite{8630924} &
  Challenging the Boundaries of Unsupervised Learning for Semantic Similarity &
  2019 &
  A. Pawar and V. Mago, &
  IEEE Access &
  Q1 &
  56 &
  11 \\ \hline
\cite{pedersen2007measures} &
  Measures of semantic similarity and relatedness in the biomedical domain &
  2007 &
  Pedersen, Ted and Pakhomov, Serguei VS and Patwardhan, Siddharth and Chute, Christopher G &
  Journal of biomedical informatics &
  Q1 &
  83 &
  555 \\ \hline
\cite{pennington2014glove} &
  Glove: Global vectors for word representation &
  2014 &
  Pennington, Jeffrey and Socher, Richard and Manning, Christopher D &
  EMNLP &
  &
  88 &
  12376 \\ \hline
 \cite{peters2018deep} &
  Deep contextualized word representations &
  2018 &
  Peters, Matthew E and Neumann, Mark and Iyyer, Mohit and Gardner, Matt and Clark, Christopher and Lee, Kenton and Zettlemoyer, Luke &
  Proceedings of NAACL-HLT &
  &
  61 &
  3842 \\ \hline
\cite{pilehvar2019wic} &
  WiC: the Word-in-Context Dataset for Evaluating Context-Sensitive Meaning Representations &
  2019 &
  Pilehvar, Mohammad Taher and Camacho-Collados, Jose &
  Proceedings of the 2019 Conference of the North American Chapter of the Association for Computational Linguistics: Human Language Technologies, Volume 1 (Long and Short Papers) &
  &
  61 &
  11 \\ \hline
\cite{pilehvar2013align} &
  Align, disambiguate and walk: A unified approach for measuring semantic similarity &
  2013 &
  Pilehvar, Mohammad Taher and Jurgens, David and Navigli, Roberto &
  Proceedings of the 51st Annual Meeting of the Association for Computational Linguistics (Volume 1: Long Papers) &
  &
  106 &
  184 \\ \hline
\cite{PILEHVAR201595} &
  From senses to texts: An all-in-one graph-based approach for measuring semantic similarity &
  2015 &
  Mohammad Taher Pilehvar and Roberto Navigli &
  Artificial Intelligence &
  Q1 &
  135 &
  66 \\ \hline
\cite{QU20181002} &
  Computing semantic similarity based on novel models of semantic representation using Wikipedia &
  2018 &
  Rong Qu and Yongyi Fang and Wen Bai and Yuncheng Jiang &
  Information Processing \& Management &
  Q1 &
  88 &
  11 \\ \hline
\cite{8642425} &
  An Efficient Framework for Sentence Similarity Modeling &
  2019 &
  Z. Quan and Z. Wang and Y. Le and B. Yao and K. Li and J. Yin &
  IEEE/ACM Transactions on Audio, Speech and Language Processing &
  Q1 &
  55 &
  4 \\ \hline
\cite{rada1989development} &
  Development and application of a metric on semantic nets &
  1989 &
  Rada, Roy and Mili, Hafedh and Bicknell, Ellen and Blettner, Maria &
  IEEE transactions on systems, man, and cybernetics &
  Q1 &
  111 &
  2347 \\ \hline
\cite{raffel2019exploring} &
  Exploring the limits of transfer learning with a unified text-to-text transformer &
  2020 &
  Raffel, Colin and Shazeer, Noam and Roberts, Adam and Lee, Katherine and Narang, Sharan and Matena, Michael and Zhou, Yanqi and Li, Wei and Liu, Peter J &
  arXiv &
  &
  &
  192 \\ \hline
\cite{resnik1995using} &
  Using information content to evaluate semantic similarity in a taxonomy &
  1995 &
  Resnik, Philip &
  IJCAI &
  &
  109 &
  4300 \\ \hline
\cite{rodriguez2003determining} &
  Determining semantic similarity among entity classes from different ontologies &
  2003 &
  Rodr{\'\i}guez, M Andrea and Egenhofer, Max J. &
  EMNLP &
  &
  88 &
  1183 \\ \hline
\cite{RUAS2019288} &
  Multi-sense embeddings through a word sense disambiguation process &
  2019 &
  Terry Ruas and William Grosky and Akiko Aizawa &
  Expert Systems with Applications &
  Q1 &
  162 &
  4 \\ \hline
\cite{rubenstein1965contextual} &
  Contextual correlates of synonymy &
  1965 &
  Rubenstein, Herbert and Goodenough, John &
  Communications of the ACM &
  Q1 &
  189 &
  1336 \\ \hline
\cite{sanchez2011ontology} &
  Ontology-based information content computation&
  2011 &
  S{\'a}nchez, David and Batet, Montserrat and Isern, David &
  Knowledge-based systems&
  Q1 &
  94 &
  251 \\ \hline
\cite{sanh2019distilbert} &
  DistilBERT, a distilled version of BERT: smaller, faster, cheaper and lighter &
  2019 &
  Sanh, Victor and Debut, Lysandre and Chaumond, Julien and Wolf, Thomas &
  arXiv &
  &
  &
  112 \\ \hline
\cite{vsaric2012takelab} &
    Takelab: Systems for measuring semantic text similarity &
  2012 &
  {\v{S}}ari{\'c}, Frane and Glava{\v{s}}, Goran and Karan, Mladen and {\v{S}}najder, Jan and Ba{\v{s}}i{\'c}, Bojana Dalbelo &
  Proceedings of the Sixth International Workshop on Semantic Evaluation (SemEval 2012) &
  &
  49 &
  224 \\ \hline
\cite{schnabel2015evaluation} &
  Evaluation methods for unsupervised word embeddings &
  2015 &
  Schnabel, Tobias and Labutov, Igor and Mimno, David and Joachims, Thorsten &
  EMNLP &
  &
  88 &
  334 \\ \hline
\cite{severyn2012structural} &
  Structural relationships for large-scale learning of answer re-ranking  &
  2012 &
  Severyn, Aliaksei and Moschitti, Alessandro &
  ACM SIGIR Conference on Research and Development in Information Retrieval &
  &
  57 &
  85 \\ \hline
\cite{severyn2013learning} &
  Learning semantic textual similarity with structural representations &
  2013 &
  Severyn, Aliaksei and Nicosia, Massimo and Moschitti, Alessandro &
  Proceedings of the 51st Annual Meeting of the Association for Computational Linguistics (Volume 2: Short Papers) &
  &
  135 &
  40 \\ \hline
\cite{shao2017hcti} &
  HCTI at SemEval-2017 Task 1: Use Convolutional Neural Network to evaluate semantic textual similarity &
  2017 &
  Shao, Yang &
  Proceedings of the 11th International Workshop on Semantic Evaluation (SemEval-2017) &
  &
  49 &
  32 \\ \hline
\cite{shawe2004kernel} &
  Kernel methods for pattern analysis &
  2004 &
  Shawe-Taylor, John and Cristianini, Nello and others &
  Book &
  &
  &
  7721 \\ \hline
\cite{silberer2014learning} &
  Learning grounded meaning representations with autoencoders &
  2014 &
  Silberer, Carina and Lapata, Mirella &
  Proceedings of the 52nd Annual Meeting of the Association for Computational Linguistics (Volume 1: Long Papers) &
  &
  61 &
  127 \\ \hline
\cite{SINOARA2019955} &
  Knowledge-enhanced document embeddings for text classification &
  2018 &
  Roberta A. Sinoara and Jose Camacho-Collados and Rafael G. Rossi and Roberto Navigli and Solange O. Rezende &
  Knowledge-based Systems &
  Q1 &
  94 &
  25 \\ \hline
\cite{biosses} &
  BIOSSES: a semantic sentence similarity estimation system for the biomedical domain &
  2017 &
  Soğancıoğlu, Gizem and Öztürk, Hakime and Özgür, Arzucan &
  Bioinformatics &
  Q1 &
  335 &
  34 \\ \hline
\cite{sultan2014dls} &
  DLS@ CU: Sentence Similarity from Word Alignment &
  2014 &
  Sultan, Md Arafat and Bethard, Steven and Sumner, Tamara &
  Proceedings of the 8th International Workshop on Semantic Evaluation (SemEval 2014) &
  &
  49 &
  112 \\ \hline
\cite{sultan2015dls} &
  Dls@ cu: Sentence similarity from word alignment and semantic vector composition &
  2015 &
  Sultan, Md Arafat and Bethard, Steven and Sumner, Tamara &
  Proceedings of the 9th International Workshop on Semantic Evaluation SemEval 2015 &
  &
  49 &
  105 \\ \hline
\cite{sun2020ernie} &
  ERNIE 2.0: A Continual Pre-Training Framework for Language Understanding. &
  2020 &
  Sun, Yu and Wang, Shuohuan and Li, Yu-Kun and Feng, Shikun and Tian, Hao and Wu, Hua and Wang, Haifeng &
  AAAI &
  &
  95 &
  53 \\ \hline
\cite{SANCHEZ20131393} &
  A semantic similarity method based on information content exploiting multiple ontologies &
  2013 &
  David Sánchez and Montserrat Batet &
  Expert Systems with Applications &
  Q1 &
  162 &
  82 \\ \hline
\cite{SANCHEZ20127718} &
  Ontology-based semantic similarity: A new feature-based approach &
  2012 &
  David Sánchez and Montserrat Batet and David Isern and Aida Valls &
  Expert Systems with Applications &
  Q1 &
  162 &
  361 \\ \hline
\cite{tai2015improved} &
  Improved semantic representations from tree-structured long short-term memory networks &
  2015 &
  Tai, Kai Sheng and Socher, Richard and Manning, Christopher D &
  Proceedings of the 53rd Annual Meeting of the Association for Computational Linguistics and the 7th International Joint Conference on Natural Language Processing (Volume 1: Long Papers) &
  &
  106 &
  1676 \\ \hline
\cite{tian2017ecnu} &
  Ecnu at semeval-2017 task 1: Leverage kernel-based traditional nlp features and neural networks to build a universal model for multilingual and cross-lingual semantic textual similarity &
  2017 &
  Tian, Junfeng and Zhou, Zhiheng and Lan, Man and Wu, Yuanbin &
  Proceedings of the 11th international workshop on semantic evaluation (SemEval-2017) &
  &
  49 &
  34 \\ \hline
\cite{TIEN2019102090} &
  Sentence modeling via multiple word embeddings and multi-level comparison for semantic textual similarity &
  2019 &
  Nguyen Huy Tien and Nguyen Minh Le and Yamasaki Tomohiro and Izuha Tatsuya &
  Information Processing \& Management &
  Q1 &
  88 &
  7 \\ \hline
\cite{dict2vec} &
  Dict2vec: Learning Word Embeddings using Lexical Dictionaries &
  2017 &
  Tissier, Julien and Gravier, Christophe and Habrard, Amaury &
  EMNLP &
  &
  112 &
  51
  \\ \hline
\cite{vaswani2017attention} &
  Attention is All you Need &
  2017 &
  Vaswani, Ashish and Shazeer, Noam and Parmar, Niki and Uszkoreit, Jakob and Jones, Llion and Gomez, Aidan N and Kaiser, Lukasz and Polosukhin, Illia &
  NIPS &
  &
  169 &
  9994 \\ \hline
\cite{Wang2007} &
  What is the Jeopardy model? A quasi-synchronous grammar for QA &
  2007 &
  Wang, Mengqiu and Smith, Noah A. and Mitamura, Teruko &
  EMNLP &
  &
  88 &
  337 \\ \hline
\cite{Wang2016} &
  Sentence similarity learning by lexical decomposition and composition &
  2016 &
  Wang, Zhiguo and Mi, Haitao and Ittycheriah, Abraham &
  COLING &
  &
  41 &
  119 \\ \hline
\cite{wu1994verbs} &
  Verbs semantics and lexical selection &
  1994 &
  Wu, Zhibiao and Palmer, Martha &
  Proceedings of the 32nd annual meeting on Association for Computational Linguistics &
  &
  106 &
  3895 \\ \hline
\cite{yang2019xlnet} &
  Xlnet: Generalized autoregressive pretraining for language understanding &
  2019 &
  Yang, Zhilin and Dai, Zihang and Yang, Yiming and Carbonell, Jaime and Salakhutdinov, Russ R and Le, Quoc V &
  Advances in neural information processing systems &
  Q1 &
  169 &
  865
   \\ \hline
\cite{7572993} &
  Computing Semantic Similarity of Concepts in Knowledge Graphs &
  2017 &
  G. Zhu and C. A. Iglesias &
  IEEE Transactions on Knowledge and Data Engineering &
  Q1 &
  148 &
  88 \\ \hline
\cite{zou2013bilingual} &
  Bilingual word embeddings for phrase-based machine translation &
  2013 &
  Zou, Will Y and Socher, Richard and Cer, Daniel and Manning, Christopher D &
  EMNLP &
  &
  88 &
  468 \label{tab:my-table}\\ \hline 
\\
\caption{Table of references used in the analysis of the survey.}
\end{longtable}
\end{tiny}
\end{appendices}
\end{document}